\documentclass[11pt]{article}
\usepackage[left=1in, right=1in, top=1in, bottom=1in]{geometry}
\pdfoutput=1

\usepackage{mylatexstyle,natbib,hyperref}
\def \la {\langle}
\def \ra {\rangle}

\usepackage{caption}
\usepackage{mathrsfs}
\usepackage{fullpage}
\usepackage[protrusion=false,expansion=true]{microtype}
\usepackage{wrapfig}
\usepackage{amsthm}
\usepackage{thmtools}
\usepackage{array}

\ifdefined\final
\usepackage[disable]{todonotes}
\else
\usepackage[textsize=tiny]{todonotes}
\fi
\allowdisplaybreaks

\title{Initialization Matters: On the Benign Overfitting of Two-Layer ReLU CNN with Fully Trainable Layers}
\author{
    Shuning Shang\thanks{Department of Computer Science and Technology, Zhejiang University. Email: {\tt shuningshang@gmail.com}}\and
    Xuran Meng\thanks{Department of Biostatistics, University of Michigan. Email: {\tt xuranm@umich.edu}}\and
    Yuan Cao\thanks{Department of Statistics and Actuarial Science, University of Hong Kong. Email: {\texttt{yuancao@hku.hk}}}\and
    Difan Zou\thanks{Department of Computer Science and Insitute of Data Science, The University of Hong Kong. Email: {\texttt{dzou@cs.hku.hk}}}
}

\date{}

\begin{document}
\maketitle
\allowdisplaybreaks
\begin{abstract}
Benign overfitting refers to how over-parameterized neural networks can fit training data perfectly and generalize well to unseen data. While this has been widely investigated theoretically, existing works are limited to two-layer networks with fixed output layers, where only the hidden weights are trained. We extend the analysis to two-layer ReLU convolutional neural networks (CNNs) with fully trainable layers, which is closer to the practice. Our results show that the initialization scaling of the output layer is crucial to the training dynamics: large scales make the model training behave similarly to that with the fixed output, the hidden layer grows rapidly while the output layer remains largely unchanged; in contrast, small scales result in more complex layer interactions, the hidden layer initially grows to a specific ratio relative to the output layer, after which both layers jointly grow and maintain that ratio throughout training.
Furthermore, in both settings, we provide nearly matching upper and lower bounds on the test errors, identifying the sharp conditions on the initialization scaling and signal-to-noise ratio (SNR) in which the benign overfitting can be achieved or not. Numerical experiments back up the theoretical results.

\end{abstract}


\section{Introduction}

Over the past several years, the phenomenon of \emph{benign overfitting} has significantly reshaped our understanding of over-parameterized deep neural networks~\citep{neyshabur2019role,bartlett2020benign}. Traditional statistical wisdom expects that models with a large number of parameters  are prone to overfitting, resulting in poor generalization on unseen data. Contrary to this expectation, modern deep neural networks often generalize well despite being highly over-parameterized, fitting the training data perfectly while maintaining low test error. This surprising behavior has sparked considerable interest in uncovering the underlying mechanisms that allow such models to avoid the harmful effects of overfitting.

In deep learning, the generalization performance of neural networks is highly influenced by the training process, which encompasses the optimization algorithms used, the loss functions, the initialization of model parameters, etc. Understanding how these factors contribute to the generalization abilities of over-parameterized models is crucial and has attracted a vast body of research. Recently, a series of theoretical works have been devoted to studying the training dynamics of a simplified neural network model, i.e., two-layer neural network models, on datasets comprising both signal and noise components (or on Gaussian mixture data) \citep{chatterji2021finite,frei2022benign,cao2022benign,kou2023benign,meng2023per,lu2024benign}. For instance, \citet{frei2022benign} analyzed risk bounds for networks employing smoothed leaky ReLU activations when learning from log-concave mixture data with label-flipping noise. By introducing a ``signal-noise decomposition'' method, \citet{cao2022benign} identified conditions that lead to sharp phase transitions between benign and harmful overfitting in two-layer convolution neural networks (CNN) with activation functions $\text{ReLU}^q$ for $q>2$. Building upon this, \citet{kou2023benign} extended the analysis to standard ReLU networks and showed that benign overfitting can be achieved under a milder assumption on the signal-to-noise ratio (SNR).  Furthermore, \citet{meng2024benign} considered a more challenging non-linear XOR dataset and proved the benign overfitting results for two-layer ReLU CNNs, which reveal the ability of over-parameterized networks to handle more complicated data structures.

Despite their contributions and valuable insights into training dynamics of neural networks, most of these works focus on a simplified setting where \textit{only the hidden layer} is trained while the output layer remains fixed. In this case, the training dynamics of the neural network weights exhibit a clean and fixed pattern in the early to middle stages, which is almost independent of the initialization scaling of the model weights (as long as they are feasibly small). Technically, it suffices to characterize the projection of the hidden weights onto the signal and noise vectors, which usually form a sequence with relatively stable dynamics. For instance, it is a linearly increasing sequence \citep{kou2023benign} when using ReLU activation, an exponentially increasing sequence \citep{zou2023benefits} when using quadratic activation, and one with a growth rate beyond exponential \citep{li2021towards,cao2022benign} when using high-order polynomial activation. However, such a simplified setting cannot fully explain the behavior of neural network training in the practical setting, where the hidden and output layers are trained jointly. In such a more realistic scenario, we need to characterize the dynamics of both the hidden and output weights simultaneously, which poses considerably more challenges compared to the case of training only the hidden layer.  Specifically, the output and hidden weights interact with each other, leading to intertwined dynamics that cannot be explained by the sequence with a fixed pattern. Given these technical difficulties, it remains unclear about the role of the learnable output layer during the training phase and how it impacts the benign overfitting results until convergence.



\subsection{Our Main Results}

In this paper, we aim to advance the research direction of benign overfitting towards more practical scenarios by investigating the training dynamics of two-layer convolutional neural networks (CNNs) where both layers are trainable. We use the signal-noise data model with signal ($\bmu$) and noise ($\bxi \sim \mathcal{N}(0, \sigma_p^2d)$) components to study model behavior under these training dynamics. In particular, the major contributions of our paper are stated as the following informal theorem under certain conditions on dimension $d$, 
sample size $n$, neural network width $m$, learning rate $\eta$, initialization scales for the weights in hidden layer $\sigma_0$ and output layer $v_0$.

\begin{theorem}[\textbf{Informal}]
    There exists a threshold $\tilde v = \mathrm{poly}(d, n, m, \sigma_p)$ for the initialization scale of output layer $v_0$ and $T^* = \mathrm{poly}(d, n, m, \sigma_p, \eta, v_0, \epsilon)$
    that the training loss converges to $\epsilon$. Meanwhile,
    \begin{itemize}
        \item When $v_0 \ge \tilde v$, if $ \|\bmu\|_2^4 / \sigma_p^4  = \tilde \Omega(d/n) $, the trained CNN will generalize with small classification error within time $T^*$. Otherwise the test error remains above $\Theta(1)$.
        \item When $v_0 \le \tilde v$, if $\|\bmu\|_2^2  / \sigma_p = \tilde \Omega(1/(mv_0^2)) $, the trained CNN will generalize with small classification error within time $T^*$. Otherwise the test error remains above $\Theta(1)$.
    \end{itemize}
\end{theorem}

This theorem establishes nearly matching upper and lower bounds on the test error for training two-layer ReLU CNN models via gradient descents. In particular, it shows 
that the initialization scale of the output layer plays an important role in the training dynamics and the benign overfitting results. Specifically, when the initialization scale of the output layer $v_0$ is above a certain threshold, the generalization error as well as the conditions on the SNR is independent of $v_0$, which becomes similar to the setting with fixed output layer \citet{kou2023benign}. On the other hand, when the initialization is below the threshold, the benign overfitting results will be dependent on $v_0$, where the conditions on the SNR (i.e., $\|\boldsymbol{\mu}\|_2^2/\sigma_p$) become milder with a larger $v_0$. The detailed version of the above Theorem is presented in Section \ref{sec: mainresult}.

\subsection{Additional Related Works}

\paragraph{Benign overfitting in linear models.} A series of research works focus on understanding benign overfitting in linear/kernel/random feature models.
\citet{bartlett2020benign} established matching upper and lower risk bounds for the minimum norm interpolator, demonstrating that overparameterization is crucial for benign overfitting.  \citet{zou2021benign} provided a sharp excess risk bound, highlighting bias-variance decomposition in constant-stepsize SGD. Beyond linear models, \citet{liao2020random} extended this analysis to random Fourier feature regression when the sample size, data dimension, and number of random features maintain fixed ratios. \citet{adlam2022random} extended the model to include bias terms, finding that a mixture of nonlinearities can improve both the training and test errors over the best single nonlinearity. \citet{misiakiewicz2022spectrum}, \citet{hu2022sharp} study the generalization performance of kernel ridge regression, revealing a “double descent” behavior in the learning curve. \citet{mallinar2022benign} found that kernels with powerlaw spectra, including Laplace kernels and ReLU neural tangent kernels, exhibit tempered overfitting. \citet{tsigler2023benign} extended the results to ridge regression and provides general sufficient conditions under which the optimal regularization is negative. \cite{mei2022generalization} computed the precise asymptotics of the test error for random features model and showed that it reproduces all the qualitative features of the double-descent scenario. \citet{meng2024multiple} further extended the calculation to ``multiple random feature model” and discover multiple descent.

\paragraph{Benign overfitting in neural networks.} A line of recent works studied benign overfitting of neural networks. \citet{li2021towards} examined benign overfitting in  two-layer ReLU network interpolator and provided an upper bound on the bias with matching upper and lower bounds on the variance. \citet{montanari2022interpolation} investigated two-layer neural networks in NTK regime and characterized the generalization error. \citet{chatterji2023deep} bounded the excess risk of interpolating deep linear networks trained by gradient flow and revealed that
these models have exactly the same conditional variance as the minimum $l_2$-norm solution.  \citet{xu2023benign} also established benign overfitting results in XOR regime and relates with grokking phenomenon. \citet{frei2023benign} showed how satisfaction of these KKT conditions implies benign overfitting in linear classifiers and in two-layer leaky ReLU networks.
 \citet{kornowski2024tempered} studied the type of overfitting transitions from tempered to benign in over-parameterized regime and showed that the input dimension has a crucial role on the overfitting profile. \citet{karhadkar2024benign} investigated two-layer leaky ReLU networks and demonstrated that benign overfitting can occur without the need for the training data to be nearly orthogonal, and the dimensionality $d$ only required to be $\Omega(n)$. \citet{jiang2024unveil}, \cite{magen2024benign} further investigated the benign overfitting phenomenon in the attention mechanism of transformer based models.

\paragraph{Signal Learning via Two-layer Neural Networks.} Regarding the optimization methods, \citet{meng2023per} investigated the mechanisms of gradient regularization, uncovering its ability to enhance signal learning while suppressing noise memorization. Additionally, \citet{lu2024benign} discovered that training with stochastic gradient descent (SGD) using large learning rates can be beneficial to the generalization performance of neural networks. 
Moreover, some works considered more  complicated data models, leveraging similar theoretical frameworks to showcase the advantages of certain training modules and feature learning mechanisms \citep{allen2022feature,zou2023benefits,meng2024benign}.  \citet{allen2022feature} explored ensemble methods in neural networks, demonstrating how ensembling can enhance feature learning and improve generalization in over-parameterized models. \citet{zou2023benefits} considered a feature-noise data model and showed that Mixup training can effectively learn rare features, which can improve the generalization.

\section{Problem Setup}
In this section, we will introduce our data model, neural network architecture, and the training algorithm. 
Firstly, following a series of prior works \citep{cao2022benign,kou2023benign,meng2024benign}, our data model $\mathcal D$ is designed as a combination of signal and noise patches, which is formally defined as follows:
\begin{definition}[Data Model]
\label{def:data_model}
Let $\bmu\in\RR^{d}$ be a fixed signal vector. Each data point $(\bx,y)$ with the input features $\bx=[\bx^{(1)\top},\bx^{(2)^\top}]^\top\in\RR^{2d}$ and label $y\in\{\pm 1\}$ is generated from the following distribution $\cD$:
\begin{enumerate}
    \item The data label $y\in\{\pm1\}$ is generated as a Rademacher random variable.
    \item A noise vector $\bxi$ is generated from the Gaussian distribution $\cN\big(\zero,\sigma_p^2(\Ib-\bmu\bmu^\top/\|\bmu\|_2^2)\big)$. 
    \item One of $\bx^{(1)}$, $\bx^{(2)}$ is randomly selected and assigned as $y\cdot\bmu$ which is the signal part, and the other is assigned as $\bxi$ which is the noise part.
\end{enumerate}
\end{definition}
Such a data model has been widely considered in many recent literature,  which has become a standard platform for understanding the behavior of different learning algorithms.  Specifically, we consider the patch $y\cdot \bmu$ as the signal patch that corresponds to the label $y$, while the other patch $\bxi$ serves as the noise patch that is unrelated to the label. The variance of $\bxi$ is assumed to ensure that the noise is orthogonal to the signal vector, for simplicity. We remark that the orthogonality assumption can be removed by applying perturbation analysis, and this assumption provides us a useful and simplified framework for our analysis.

\paragraph{CNN Model.}
We consider a two-layer ReLU CNN model with both learnable hidden and output layers. Considering the data model in Definition \ref{def:data_model},  the neural network function for the logit $j$ is formulated as follows:
\begin{align*}
    F_j(\Wb,\vb;\bx) = \sum_{r=1}^{m} \sum_{p=1}^2 v_{j,r,p}\sigma(\la\wb_{j,r},\bx^{(p)}\ra).
\end{align*}
where $\sigma(z)=\max\{0,z\}$ denotes the ReLU activation function, $m$ denotes the number of filters, $j\in\{+1,-1\}$ denotes the index of the logit, $\wb_{j,r}\in\RR^d$ denotes the weight of the $r$-th filter corresponding to the $j$-th filter, and $v_{j,r,p}$ denotes the weight of the output layer that corresponds to the $p$-th patch, $r$-th filter, and $j$-th logit. Then, the output of the CNN model can be formulated as follows:
\begin{align}
\label{eq: CNN_model}
f(\Wb,\vb;\bx) = F_{+1}(\Wb_{+1},\vb_{+1};\bx) -  F_{-1}(\Wb_{-1},\vb_{-1};\bx).
\end{align}
Here, $\Wb_{+1}$ and $\Wb_{-1}$ collect all the weight vectors $\wb_{+1,r}$ and $\wb_{-1,r}$ respectively. Similarly, $\vb_{+1}$ and $\vb_{-1}$ collect all the scalars $v_{+1,r,p}$ and $v_{-1,r,p}$ for all $r$ and $p$, respectively.

\paragraph{Initialization.}
For initialization, we assume that each element in $\Wb^{(0)}$ is independently drawn from Gaussian distribution $\cN(0,\sigma_0^2)$ with variance $\sigma_0^2$, and we initialize all the elements in the output layer with the same value $v_0$. In the entire paper, we will refer to $\sigma_0$ and $v_0$ as the initialization scales of the hidden layer and output layer respectively.

\paragraph{Gradient Descent.}
Given $n$ i.i.d training data points $\{(\bx_i,y_i\}_{i\in[n]}$,  we define the empirical risk function as follows
\begin{align*}
L_S(\Wb,\vb) = \frac{1}{n}\sum_{i=1}^n \ell\big(y_if(\Wb,\vb;\bx_i)\big),
\end{align*}
where $\ell(z)=\log(1+\exp(-z))$ denotes the logistic loss. Then, the gradient descent is performed on the above empirical risk function. Let $\Wb$ and $\vb$ be the collections of parameters in the hidden layer and output layer respectively, starting from the initialization $\Wb^{(0)}$ and $\vb^{(0)}$, the gradient descent takes the following form
\begin{align*}
    &\Wb^{(t+1)}=\Wb^{(t)}-\eta\nabla_{\Wb^{(t)}}L_S (\Wb^{(t)},\vb^{(t)}),\\
    &\vb^{(t+1)}=\vb^{(t)}-\eta\nabla_{\vb^{(t)}}L_S (\Wb^{(t)},\vb^{(t)}).
\end{align*}
Here, $\eta>0$ is the learning rate, which is set to be the same for all trainable parameters. Readers may refer Appendix~\ref{sec: update_formula} for the detailed update form. 
Importantly, we remark that compared to the prior works \citep{cao2022benign,kou2023benign,meng2024benign} that only consider training the hidden layer parameters $\Wb$, we consider the training of both $\Wb$ and $\vb$. This will exhibit a fundamentally different training dynamics and lead to different generalization performances, which will be thoroughly investigated in the next section.

\paragraph{Notations.} Given two sequences ${x_n}$ and ${y_n}$, we denote $x_n = O(y_n)$ and $y_n = \Omega(x_n)$ if there exist constants $C_1 > 0$ and $N > 0$ such that $|x_n| \leq C_1 |y_n|$ for all $n \geq N$. We define $x_n = \Theta(y_n)$ if both $x_n = O(y_n)$ and $x_n = \Omega(y_n)$ hold. To hide logarithmic factors in these notations, we use $\widetilde{O}(\cdot)$, $\widetilde{\Omega}(\cdot)$, and $\widetilde{\Theta}(\cdot)$. The notation $\one(\cdot)$ represents the indicator variable of an event. Furthermore, we say $y = \poly(x_1, \dots, x_k)$ if $y = O(\max{|x_1|, \dots, |x_k|}^D)$ for some $D > 0$, and $y = \polylog(x)$ if $y$ is a polynomial function of  $\log(x)$.

\section{Main Results}\label{sec: mainresult}
In this section, we present our main theoretical results. Our results are based on the following conditions on the dimension $d$, 
sample size $n$, neural network width $m$, initialization scales for the weights in hidden layer (i.e., $\sigma_0$) and output layer (i.e., $v_0$), and learning rate $\eta$. In this paper, we consider 
the learning period $0 \le t \le T^*$, where $T^* = \eta^{-1} \mathrm{poly}(d, n, m,\varepsilon^{-1})$ is the maximum admissible iterations. Notably, such an iteration number $T^*$ can be arbitrarily large as long as it is polynomially with respect to the dimension, sample size, number of filters, and the inverse of target precision. This can be certainly satisfied in many practical scenarios.
\begin{condition}
\label{condition: main condition}
    Suppose there exists a sufficiently large constant $C$. For certain probability parameter $\delta \in (0,1)$, the following hold,
    \begin{enumerate}
        \item The dimension is sufficiently large: $d \ge C \max \{n \sigma_p^{-2}\|\bmu\|^2_2, n^2 \log(4n^2/\delta), n^{5/3}\sigma_p^2 \log(4n^2/\delta)\} \cdot (\log(T^*))^2$.
        \item Training sample size $n$ and neural network width satisfy: $n \ge C\log(m/\delta), m \ge C \log(n/\delta)$.
        \item The standard deviation of Gaussian initialization $\sigma_0$ is appropriately chosen such that: $\sigma_0 \le (C\max\{\sigma_p d/\sqrt{n}, (nd)^{1/4}(m\sigma_p)^{1/2},  \sqrt{\log(n/\delta)}\cdot \|\bmu\|_2\})^{-1}$.
        \item The initialization of output layer satisfies:
        
        $C \sigma_0 \sqrt{n\log(8mn/\delta)} \le v_0 \le (C\max\{m\sigma_0 \sigma_p \sqrt{d\log(8mn/\delta)}, n \sigma_p^2 \sqrt{m\log(4n^2/\delta)/d}\})^{-1}$.
        \item The learning rate satisfies: $\eta \le \big(C \max\{m^3 \sigma_p^2 d v_0^4/n, m \sigma_p^2 d v_0^2/n, \sigma_p^2 d^{3/2} v_0^2/(n^2 \sqrt{\log(4n^2/\delta)}\})\big)^{-1}$.
    \end{enumerate}
\end{condition}

The first condition on $d$, $n$, and $m$ is to ensure that the learning problem is in a sufficiently over-parameterized setting.
Here, the term merely involves additional logarithmic factors since  $T^*$ is at most a polynomial function of the relevant parameters. Besides, this assumption also states that $n\|\bmu\|_2^2\le C\cdot d\sigma_p^2$, which suggests that we focus on the small SNR setting since otherwise the benign overfitting can be trivially achieved. Similar conditions have been made in recent studies of over-parameterized models (\cite{chatterji2021finite};\cite{cao2022benign};\citet{meng2023per}). The combined condition for the sample size $n$ and neural network width $m$ guarantees that certain statistical characteristics of the training data and weight initialization hold with a high probability. This condition can be readily satisfied when  $n$ is sufficiently large and $m=\mathrm{polylog}(n)$. 
We then remark a set of technical conditions on the hyperparameters in our problem, including initialization scaling and learning rate. In particular, the condition on $\sigma_0$ ensures a small neural network initialization such that the training of the neural network is less influenced by the initialization and mainly depends on parameter updates. The condition on $v_0$ is a technical requirement for the analysis of training dynamics, particularly for the output layer. Finally, the condition on the learning rate $\eta$ is another technical requirement that ensures the stable convergence of gradient descent. 


The main difference between our work with previous works is that we have the trainable output layer, leading to a different behaviors in signal learning and noise memorization. Moreover, in the subsequent theoretical analysis, we will reveal the important role of the initialization scaling, especially for the output layer, to the model generalization.  The following theorem states the generalization performance when using a large initialization for the output layer.

\begin{theorem}
\label{thm: single_phase}
    Under condition \ref{condition: main condition}, if $v_0 = \Omega(n^{1/4}\sigma_p^{-1/2}d^{-1/4}m^{-1/2})$, then with probability at least $1-\delta$, the following results hold:
    \begin{itemize}
        \item \textbf{Convergence:} For any $\epsilon > 0$, there exists $t = \tilde O\Big(\frac{\sqrt{n}}{\eta \sigma_p \sqrt{d} \epsilon}\Big)$ such that the training loss converges to $\epsilon$: $L_{S}(\Wb^{(t)},\vb^{(t)}) \le \epsilon$.
        \item \textbf{Benign Overfitting:} When $ \|\bmu\|_2^4 / \sigma_p^4  = \tilde \Omega(d/n) $, the trained CNN will generalize with small classification error: $L_{D}(\Wb^{(t)}, \vb^{(t)}) = o(1)$ for any $t \ge \Theta\Big(\frac{n}{\eta \sigma_p^2 d m v_0^2}\Big)$.
        \item \textbf{Harmful Overfitting:} When $ \|\bmu\|_2^4 / \sigma_p^4  = \tilde{O}(d/n)$, 
        the test error $L_{D}(\Wb^{(t)}, \vb^{(t)}) \ge C_1$ for any $0 \le t \le \tilde O\Big(\frac{\sqrt{n}}{\eta \sigma_p \sqrt{d} \epsilon}\Big)$.
    \end{itemize}
    Here $C_1$ is some constant larger than $0$.
\end{theorem}
Theorem~\ref{thm: single_phase} shows the results when initialization of output layer is greater than the threshold of $\Theta(n^{1/4}\sigma_p^{-1/2}d^{-1/4}m^{-1/2})$. We can verify that this threshold falls into the range of the fourth statement in Condition \ref{condition: main condition}. In Theorem~\ref{thm: single_phase}, the first result states convergence of training loss and ensures that the obtained CNN performs perfect fitting on the training data set. The second and last results state the cases of benign and harmful overfitting respectively, which is determined by the ratio $\|\bmu\|_2^4/\sigma_p^4$. In this case, it is clear that the value of $v_0$ will not affect the generalization performance.

Moreover, we would like to remark that  for large $v_0$, the relative update of the output layer, i.e., the ratio of the weight changes, will be significantly smaller than that of the hidden layer. Therefore, training both layers will exhibit a similar behavior as training only the hidden layer. This is also consistent with the benign overfitting results with \citet{kou2023benign}, which concerns the case of fixed output layer. In their results, 
the SNR threshold between benign and harmful overfitting is $\|\bmu\|_2^4/\sigma_p^4 = \Theta(d/n)$, which matches our threshold in Theorem \ref{thm: single_phase} with up to logarithmic factors.

Then, we will consider the case of small initialization for the output layer, i.e., $v_0=O(n^{1/4}\sigma_p^{-1/2}d^{-1/4}m^{-1/2})$.
The following theorem shows that $v_0$ will play an important role in determining whether the trained model will achieve benign or harmful overfitting.
\begin{theorem}
\label{thm: double_phase}
    Under condition \ref{condition: main condition}, if $v_0 = {O}(n^{1/4}\sigma_p^{-1/2}d^{-1/4}m^{-1/2})$, with probability at least $1-\delta$ the following results hold:
    \begin{itemize}
        \item For any $\epsilon > 0$, there exists $t = \tilde O\Big(\frac{\sqrt{n}}{\eta \sigma_p \sqrt{d} \epsilon}\Big)$ such that the training loss converges to $\epsilon$: $L_{S}(\Wb^{(t)},\vb^{(t)}) \le \epsilon$.
        \item When $\|\bmu\|_2^2  / \sigma_p = \tilde \Omega(1/(mv_0^2)) $, the trained CNN will generalize with small classification error: $L_{D}(\Wb^{(t)}, \vb^{(t)})  = o(1)$ for any $t\geq \Theta\Big(\frac{n}{\eta \sigma_p^2 d m v_0^2}\Big)$.
        \item When $ \|\bmu\|_2^2 / \sigma_p = \tilde O(1/(mv_0^2))$, 
        the test error $L_{D}(\Wb^{(t)}, \vb^{(t)}) \ge C_2$ for any $0 \le t \le \tilde O\Big(\frac{\sqrt{n}}{\eta \sigma_p \sqrt{d} \epsilon}\Big)$.
    \end{itemize}
    Here $C_2$ is some constant greater than $0$.
\end{theorem}
\begin{wrapfigure}{R}{0.4\textwidth}
\vskip -0.2in
    \centering
\includegraphics[width=0.4\columnwidth]{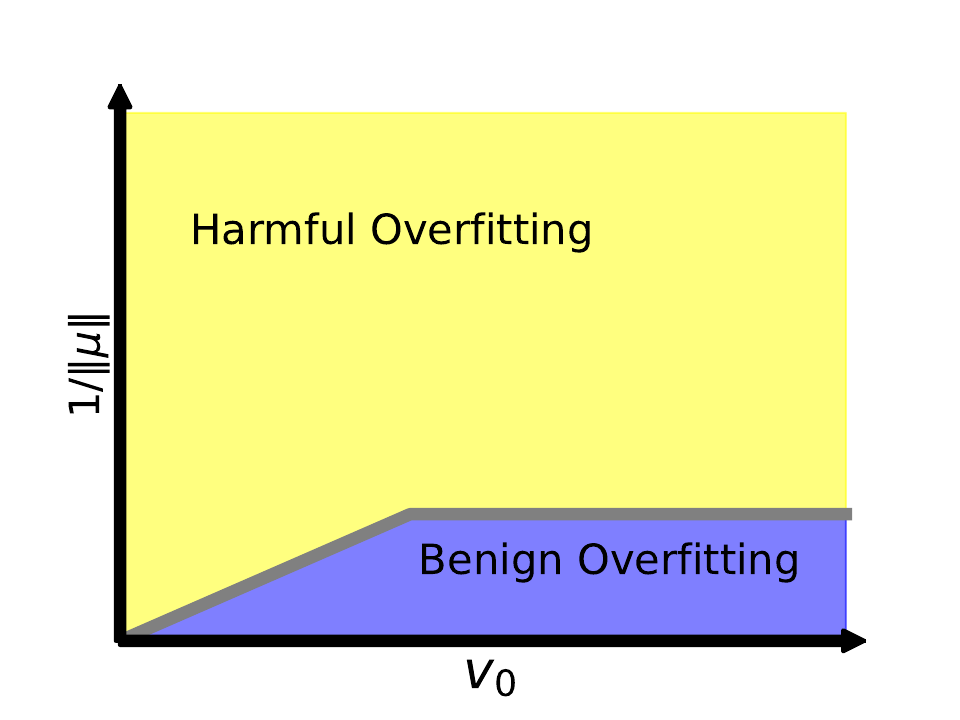}
\vskip -0.2in
    \caption{Illustration of the phase transition between small and large initialization of the output layer. The $x$-axis represents $v_0$, while the $y$-axis corresponds to $\|\bmu\|_2^{-1}$. }
    \label{fig: sim}
\end{wrapfigure}
In Theorem~\ref{thm: double_phase}, the first result ensures the convergence of the training loss when initialization in output layer is below the threshold $\Theta(n^{1/4}\sigma_p^{-1/2}d^{-1/4}m^{-1/2})$. Under this scenario, training iterations become more complicated, with both the output layer and  hidden layer evolving significantly during the training updates. The condition for benign/harmful overfitting in our analysis is also changed, and larger initialization in the output layer will be beneficial for benign overfitting. Specifically, when the initialization in the output layer is at the threshold $\Theta(n^{1/4}\sigma_p^{-1/2}d^{-1/4}m^{-1/2})$, the condition for benign overfitting in Theorem~\ref{thm: double_phase} matches that in Theorem~\ref{thm: single_phase}.

Combining Theorem~\ref{thm: double_phase} with Theorem~\ref{thm: single_phase}, it can be shown that varying the levels of initialization in output layer guarantees training convergence. However, these different initialization levels result in distinct training dynamics, significantly altering the conditions for benign and harmful overfitting. A notable phase transition occurs as we gradually increase the initialization levels in the output layer. While the condition $n\|\bmu\|_2^4/(\sigma_p^4d)=\tilde\Omega(1)$ is necessary for benign overfitting, the value of $v_{j,r,p}^{(0)}$ also significantly impacts its occurrence. It is evident that even if $n\|\bmu\|_2^4/(\sigma_p^4d)=\tilde\Omega(1)$, small values of $v_0$ can still lead to harmful overfitting during training iterations with both layers trainable. This phase transition can be illustrated in Figure \ref{fig: sim}. In Figure~\ref{fig: sim}, the purple region represents  the overfitted CNN trained under gradient descent that have small population loss, while the yellow region corresponds to the population loss remains of constant order. As predicted by the theory, when the initialization scale is below a specific threshold, the boundary between benign and harmful overfitting in Figure~\ref{fig: sim} follows the relation $v_0 \propto 1/\|\bmu\|_2$
When the initialization scale exceeds this threshold, the boundary maintains a near constant order.
\paragraph{Comparison with prior works.}
We further make a detailed comparison with the prior works \citep{frei2022benign,cao2022benign,kou2023benign}, which concerns only training the hidden layer, in terms of the conditions for achieving the benign overfitting. The results are summarized in Table \ref{tab:snr_compare}, where we consider $m=\mathrm{polylog}(n)$ and $\sigma_p=1$ for ease of comparison. In particular, for large $v_0=\tilde\Omega(n^{1/4}d^{-1/4})$, it is clear that our condition on the SNR (or the strength of the signal vector) resembles those in \citet{frei2022benign,kou2023benign} up to logarithmic factors, and is milder than that in \citet{cao2022benign}. For small  $v_0=\tilde O(n^{1/4}d^{-1/4})$, our condition on the SNR will depend on $v_0$, where smaller $v_0$ leads to a stronger requirement for the signal strength. Further note that we assumed $\|\bmu\|_2^2=O\big(d\sigma_p^2/n)=O(d/n)$ in Condition \ref{condition: main condition}, thus we need to set $v_0=\tilde \Omega(n^{1/2}/d^{1/2})$ in order to guarantee the existence of a data model  (governed by SNR) under Condition \ref{condition: main condition} that can provably achieve benign overfitting. Notably, this reduces to the SNR condition in \citet{cao2022benign} when $q\rightarrow 2^+$.


\begin{table}[!t]
\small
    \centering
        \caption{Comparisons of SNR threshold for benign overfitting with previous works. For the simplicity of comparison, we assume the noise scale $\sigma_p=1$ and network width $m=\mathrm{polylog}(n)$ across all settings.}
    \label{tab:snr_compare}
    \renewcommand{\arraystretch}{1.5} 
    \begin{tabular}{>{\centering\arraybackslash}m{1.7cm}|>{\centering\arraybackslash}m{4.0cm}|>{\centering\arraybackslash}m{2.8cm}|>{\centering\arraybackslash}m{3.1cm}|>{\centering\arraybackslash}m{2.7cm}}
        \hline
        Work & Ours & \citet{cao2022benign} & \citet{frei2022benign} & \citet{kou2023benign}  \\ \hline
        Model & $\text{ReLU}$ activation,\newline fully trainable layers & $\text{ReLU}^q$ activation &$\gamma-\text{Leaky, H-smooth}$ activation & $\text{ReLU}$ activation\\ \hline
        SNR Conditions & $v_0^4 \|\bmu\|_2^4=\tilde \Omega(1)$ (small $v_0$)\newline $\frac{n \|\bmu\|_2^4}{d}=\tilde \Omega(1)$ (large $v_0$) & $n\left(\frac{\|\bmu\|_2}{\sqrt{d}}\right)^q=\tilde \Omega(1)$ & $\frac{n \|\bmu\|_2^4}{d}=\Omega(1)$ & $\frac{n \|\bmu\|_2^4}{d}=\Omega(1)$ \\ \hline
    \end{tabular}
\end{table}
\section{Overview of Proof Technique}
\label{sec: overview_proof_tech}
In this section, we discuss the main challenges in studying benign overfitting under our setting, and explain some key techniques to overcome these challenges. The complete proofs of all the results are given in the appendix.

\paragraph{Main challenges:} The first challenge is that we have both trainable hidden and output layer. So during the training process, the dynamics of both layers will affect each other. Besides, we found that different initialization levels of the output layer lead to varying dynamics in the neural network. So we need to conduct specific analyses for different cases, which will be discussed in the following.

\subsection{Dynamics of two intertwined sequences}
To solve the first challenge, we introduce two intertwined sequences to simulate the dynamics of two layers. We have the update formulas of $\la\wb_{j,r}^{(t)},y_k \bmu\ra$ and $v_{j,r,1}^{(t)}$:
\begin{align*}
\la\wb_{j,r}^{(t+1)},y_k \bmu\ra &= \la\wb_{j,r}^{(t)},y_k \bmu\ra - j y_k \frac{\eta}{n} v_{j,r,1}^{(t)} \|\bmu\|_2^2 \sum_{i=1}^n\ell'^{(t)}_i \mathbb{I}(\la\wb_{j,r}^{(t)},y_i \cdot \bmu\ra >0),\\
v_{j,r,1}^{(t+1)} & = v_{j,r,1}^{(t)} - j y_k \frac{\eta}{n} \la \wb_{j,r}^{(t)},y_k \bmu \ra \sum_{i=1}^{n}\ell'^{(t)}_i \mathbb{I}(\la\wb_{j,r}^{(t)},y_i \cdot \bmu\ra >0).
\end{align*}
Focusing on the last term of the above two equations, we found that the increment in each iteration is proportional to the value of the counterpart, i.e. $\Delta \la\wb_{j,r}^{(t+1)},y_k \bmu\ra \propto v_{j,r,1}^{(t)}$, $\Delta v_{j,r,1}^{(t+1)} \propto \la\wb_{j,r}^{(t)},y_k \bmu\ra$. Therefore, we use the following sequences to simulate them:
\begin{align}
a_{t+1} &= a_t + A\cdot b_t,\nonumber\\
b_{t+1} &= b_t + B\cdot a_t.\label{eq: simulate_sequence}
\end{align}
In this way, we can study the simplified model to better analyze the dynamics of two layers. We have the following key technical lemma:
\begin{lemma}[Dynamic of two intertwined sequences]
\label{lemma: overview_interwined_sequence}
Consider two sequences $\{a_t\}$ and $\{b_t\}$ which satisfy (\ref{eq: simulate_sequence}),
where A,B are constant irrelevant with $t$ and satisfy $ AB \le 1$ and $a_0/b_0 = o(\sqrt{A/B})$. Then there exists an iteration $t_1=\Theta(1/\sqrt{AB})$ such that 
\begin{align*}
    a_{t_1}/b_{t_1} = \Theta(\sqrt{A/B});\quad a_{t_1} = \Theta(\sqrt{A/B}\cdot b_0);\quad b_{t_1} = \Theta(b_0).
\end{align*}
\end{lemma}
The first conclusion can also be interpreted as two sequences becoming ``balanced'' as the ratio of two sequences has a invariant order. However, the analysis of two layers becoming balanced requires the condition that the loss derivative is at constant order on all training data, $\ell'^{(t)}_i = \Theta(1)$ for all $i \in [n]$. We can make the following intuitive assumptions:
\begin{itemize}
    \item when the output layer initialization is \textbf{large}, it takes a long time for the hidden layer to grow, so $\ell'^{(t)}_i = o(1)$ before two layers reach the balancing state.
    \item when the output layer initialization is \textbf{small}, two layers will become balanced before $\ell'^{(t)}_i = o(1)$.
\end{itemize}
Therefore, we consider two cases respectively. 

\subsection{Large initialization case}
Similar to \cite{cao2022benign} and \cite{kou2023benign}, we divide the whole training process into two stages, where in \textbf{Stage 1} the loss derivative is at constant order on all training samples and in \textbf{Stage 2} the training loss is optimized to be arbitrarily small. In this case when output layer initialization is large, we find that although both layers are trainable, the output layer will not change much at the end of stage 1, which is formalized into the following proposition:
\begin{proposition}(Large Initialization, Stage 1)
\label{prop: overview_stage1_l}
    Under Condition \ref{condition: main condition}, with probability at least $1-\delta$ there exists $T_1 = \Theta\Big(\frac{n}{\eta \sigma_p^2 d m v_0^2}\Big)$ such that for every sample $k \in [n]$ the following hold:
    \begin{enumerate}
        \item Noise memorization reaches constant level: $\sum_{r=1}^m v^{(T_1)}_{y_k,r,2}\cdot\sigma(\la\wb_{y_k,r}^{(T_1)},\bxi_k\ra)= \Theta(1)$.
        \item Two layers of noise part are not balanced: $\la\wb_{y_k,r}^{(T_1)},\bxi_k\ra / v^{(T_1)}_{y_k,r,2} = o(\sigma_p \sqrt{d} / \sqrt{n})$.
        \item Signal learning reaches certain level: $\sum_{r=1}^m v_{y_k,r,1}^{(T_1)} \la\wb_{y_k,r}^{(T_1)},y_k \bmu\ra = \Theta\Big(\frac{n \|\bmu\|_2^2 }{\sigma_p^2 d}\Big)$.
        \item Two layers of signal part are not balanced: $ \la\wb_{y_k,r}^{(T_1)},y_k \bmu\ra / v^{(T_1)}_{y_k,r,1} = o(\|\bmu\|_2)$.
    \end{enumerate}
\end{proposition}
Proposition \ref{prop: overview_stage1_l} takes advantage of the condition that $\ell'^{(t)}_i = \Theta(1)$ when $t \in [0, T_1]$ so we can replace $\ell'^{(t)}_i$ factors by their constant upper and lower bounds during analysis. At the end of Stage 1, the noise memorization has reached constant order while the signal learning has reached a level that proportional to SNR. 

Based on the results above, we proceed to the second stage. In this stage, to give more careful analysis of signal learning and noise memorization at each layer, we adopt signal-noise decomposition similar as \cite{kou2023benign}:
\begin{definition}
\label{def: overview_w_decompose}
Let $\wb_{j,r}^{(t)}$ for $j \in \{\pm 1\}$, $r \in [m]$ be the convolution filters of the CNN at the t-th iteration. Then there exist unique coefficients $\gamma_{j,r}^{(t)}$ and $\rho_{j,r,i}^{(t)}$ such that 
\begin{align*}
    \wb_{j,r}^{(t)} = \wb_{j,r}^{(0)} + j \cdot \gamma_{j,r}^{(t)}\cdot \|\bmu\|_2^{-2} \cdot \bmu + \sum_{i=1}^n \rho_{j,r,i}^{(t)} \cdot \|\bxi_i\|_2^{-2} \cdot \bxi_i.
\end{align*}
Further denote $\overline{\rho}_{j,r,i}^{(t)} := \rho_{j,r,i}^{(t)} \mathbb{I}(\rho_{j,r,i}^{(t)} \ge 0)$, $\underline{\rho}_{j,r,i}^{(t)} := \rho_{j,r,i}^{(t)} \mathbb{I}(\rho_{j,r,i}^{(t)} \le 0)$. Then we have
\begin{align*}
    \wb_{j,r}^{(t)} = \wb_{j,r}^{(0)} + j \cdot \gamma_{j,r}^{(t)}\cdot \|\bmu\|_2^{-2} \cdot \bmu + \sum_{i=1}^n \overline{\rho}_{j,r,i}^{(t)} \cdot \|\bxi_i\|_2^{-2} \cdot \bxi_i + \sum_{i=1}^n \underline{\rho}_{j,r,i}^{(t)} \cdot \|\bxi_i\|_2^{-2} \cdot \bxi_i.
\end{align*}
\end{definition}
Here $\gamma_{j,r}^{(t)} \approx \la \wb_{j,r}^{(t)}, j\bmu \ra,\rho_{j,r,k}^{(t)} \approx \la \wb_{j,r}^{(t)}, \bxi_i \ra$ are decomposition coefficients. With this decomposition, we can estimate the scale of signal learning and noise memorization within the maximum admissible iterations $T^* = \eta^{-1} poly(d, n, m,\varepsilon^{-1})$:

\begin{proposition}
\label{prop: overview_stage2_analysis_c1}
    For any $T_1 \le t \le T^*$, it holds that,
    \begin{equation}
    \label{eq: overview_scale}
        \begin{split}
                    &0 \le v_{j,r,1}^{(t)} \le C_4 v_0 \alpha, \quad 0 \le v_{j,r,2}^{(t)} \le C_5 m v_0^2 \alpha, \quad 0 \le \gamma_{j,r}^{(t)} \le \frac{C_6 n \|\bmu\|_2^2 \alpha}{m v_0 \sigma_p^2 d},\\
        &0 \le \overline{\rho}_{j,r,k}^{(t)} \le \alpha,\quad 0 \ge \underline{\rho}_{j,r,k}^{(t)} \ge -2\sqrt{\log(8mn/\delta)}\cdot \sigma_0 \sigma_p\sqrt{d} - 8 \sqrt{\frac{\log(4n^2/\delta)}{d}} n\alpha \ge -\alpha ,
        \end{split}
    \end{equation}
    for all $j \in \{\pm 1\}, r \in [m]$ and $k \in [n]$. Here $\alpha =\frac{4\log(T^*)}{k_1 m v_0}$ where $k_1 = \min\limits_{k,j,r, t \le T_1} \Big\{\frac{v_{j,r,2}^{(t)} \sigma_p \sqrt{d}}{\sqrt{n}\la \wb_{j,r}^{(t)}, \bxi_k \ra}\Big\} = \Theta(1)$ and $C_4, C_5, C_6=\Theta(1)$.
\end{proposition}

During Stage 2, although the loss derivatives are no longer at constant level, we can show that they are still ``uniform" among all samples, i.e. $\ell'^{(t)}_i/\ell'^{(t)}_j \le C$ for $\forall i,j \in [n]$. Meanwhile, even though the two layers have not reached balancing state, we can prove that their ratio is bounded within certain level. We can summarize into the following lemma:

\begin{lemma}
\label{lemma: overview_key_lemma_stage2_c1}
    Under Condition \ref{condition: main condition}, suppose \eqref{eq: overview_scale} hold at iteration t. Then it holds that:
    \begin{enumerate}
        \item $\sum_{r=1}^m v_{y_k, r, 2}^{(t)} \sigma(\la \wb_{y_k, r}^{(t)}, \bxi_k \ra) - v_{y_i, r, 2}^{(t)} \sigma(\la \wb_{y_i, r}^{(t)}, \bxi_i \ra) \le \kappa$ and $\ell'^{(t)}_i / \ell'^{(t)}_k \le C_7$ for all $i, k \in [n]$ and $y_i = y_k$.
        \item $v_{y_k, r, 2}^{(t)} \ge C_8 v_0$ for all $r \in [m], k \in [n]$.
        \item Define $S_i^{(t)} := \{r \in [m]: \la w_{y_i,r},\bxi_k \ra >0\}$, and $S_{j,r}^{(t)}:=\{k\in [n]:y_k=j,\la w_{j,r}^{(t)}, \bxi_k \ra > 0\}$. For all $k \in [n], r \in [m]$ and $j \in \{\pm 1\}$, $S_k^{(0)} \subseteq S_k^{(t)}, S_{j,r}^{(0)} \subseteq S_{j,r}^{(t)}$.
        \item $\Theta \Big(\frac{1}{\|\bmu\|_2}\Big) \le \frac{v_{y_k, r, 1}^{(t)}}{\la \wb_{y_k, r}^{(t)}, y_k\bmu \ra} \le \Theta\Big(\frac{m v_0^2 \sigma_p^2 d}{n \|\bmu\|_2^2}\Big)$ for all $k \in [n], r \in [m]$.
        \item  $\Theta \Big(\frac{\sqrt{n}}{\sigma_p \sqrt{d}}\Big) \le \frac{v_{y_k,r,2}^{(t)}}{\la \wb_{y_k,r}^{(t)}, \bxi_k \ra} \le \Theta (mv_0^2)$ for all $k \in [n], r \in [m]$.
    \end{enumerate}
    Here, $\kappa, C_7, C_8 = \Theta(1)$. 
\end{lemma}
The first statement shows that the noise memorization is ``balanced'' among all training samples. The second statement ensures that the output layer remains above its original order and the third statement shows that the neurons activated by a noise patch at the beginning will remain activated throughout the process. 

 Assume a test sample $(\bx, y), \bx = (\bx^{(1)}, \bx^{(2)})^{\top}$ where $\bx^{(1)} = y \cdot \bmu$. To estimate the test error, it is equivalent to calculating $\mathbb{P}(y \cdot f(\Wb, \Vb; \bx) > 0)$. And from the output of our trained model, this can be expressed as:
    \begin{align*}
        y \cdot f(\Wb^{(t)}, \Vb^{(t)}; \bx) = & \sum_{r=1}^{m} [\underbrace{v_{y,r,1}^{(t)} \sigma(\la \wb_{y,r}^{(t)}, y\bmu \ra)}_{I_1} + \underbrace{v_{y,r,2}^{(t)}\sigma(\la \wb_{y,r}^{(t)}, \bxi \ra)}_{I_2}]\\
        &- \sum_{r=1}^{m} [\underbrace{v_{-y,r,1}^{(t)} \sigma(\la \wb_{-y,r}^{(t)}, y\bmu \ra)}_{I_3} + \underbrace{v_{-y,r,2}^{(t)}\sigma(\la \wb_{-y,r}^{(t)}, \bxi \ra)}_{I_4}].
    \end{align*}
$I_1 \sim I_4$ are all positive terms. And from our previous analysis, $\la \wb_{-y, r}^{(t)}, y\bmu \ra$ is non-increasing and its order is neglectable compared with $\la \wb_{y,r}^{(t)}, y\bmu \ra$ which is non-decreasing. Therefore, to accurately predict the test sample, the condition $y \cdot f(\Wb^{(t)}, \Vb^{(t)}; \bx) > 0$ can be simplified to $I_1 > I_4$, which indicates the output of the signal part should outweigh the noise part. $I_1$ can be lower bounded from the third statement in Proposition \ref{prop: overview_stage1_l}. For $I_4$, as the noise patch $\bxi$ is generated from Gaussian distribution, we have 
\begin{align*}
    \mathbb{E}\bigg[\sum_{r=1}^m v_{y,r,2}^{(t)} \sigma(\la \wb_{y,r}^{(t)}, \bxi \ra)\bigg] = \sum_{r=1}^{m} v_{-y,r,2}^{(t)} \frac{\|\wb_{-y,r}^{(t)}\|_2 \sigma_p}{\sqrt{2\pi}}.
\end{align*}
So the scale of $I_4$ is determined by $ v_{-y,r,2}^{(t)}$ and $\|\wb_{-y,r}^{(t)}\|_2$. From the signal-noise decomposition analysis, we can upper bound $\|\wb_{-y,r}^{(t)}\|_2$ as
$$
\|\wb_{j,r}^{(t)}\| \le  \|\wb_{j,r}^{(0)}\| + |\gamma_{j,r}^{(t)}\cdot \|\bmu\|_2^{-2} \cdot \bmu| + \bigg|\sum_{i=1}^n \rho_{j,r,i}^{(t)} \cdot \|\bxi_i\|_2^{-2} \cdot \bxi_i \bigg|,
$$
where $\gamma_{j,r}^{(t)}$, $\rho_{j,r,i}^{(t)}$ can be upper bounded by Proposition \ref{prop: overview_stage2_analysis_c1}. Based on these results, we can apply a standard technique which is also employed in \cite{cao2022benign}, \cite{meng2024benign} to complete the proof of Theorem \ref{thm: single_phase}. We defer the detailed analyses to the appendix.

Note that when $v_0=1/m$, our result is very much similar to \cite{kou2023benign}. However, the main difference is that our SNR bounds for benign and harmful overfitting include logarithmic factors. The reason is that since we have both trainable layers, their updates at each iteration are closely related to each other. So we cannot derive the same conclusion as them that the signal learning and noise memorization has a time-invariant ratio, i.e. $\sum_{i=1}^n \overline{\rho}_{j,r,i}^{(t)}/\gamma_{j,r}^{(t)} = \Theta(\text{SNR}^{-2})$. We leave the question of optimizing the SNR bound to be tighter as a future work.

\subsection{Small initialization case}
Then we consider the case when the output layer initialization is small. We also use two-stage analysis, but the process of the first stage is more complicated as two layers will become balanced before the end of this stage. To better characterize the dynamics before and after two layers reach balancing state, we divide the first stage into two phases: \textbf{linear phase} and \textbf{quadratic phase}.

The analysis for the first phase is similar to the case with large initialization. The output layer will not change much while the hidden layer will grow until two layers become ``balanced". The idea of balance between different layers is similar to the analyses of adjacent layers in deep networks studied by \citet{arora2018optimization} and \citet{ji2019implicit}. By utilizing lemma \ref{lemma: overview_interwined_sequence}, we have the main conclusion in this phase:
\begin{lemma}[Noise Memorization becomes balancing]
\label{lemma: overview_noise_balanced}
Under Condition \ref{condition: main condition}, if $|\ell'^{(t)}_i|=\Theta(1)$ for $\forall i \in [n]$, there exists a time $T_0 = \Theta(\frac{\sqrt{n}}{\eta \sigma_p \sqrt{d}}) \le T^{*,1}$ that for all $y_k=j$:
\begin{align*}
&v^{(T_0)}_{j,r,2}/\la\wb_{j,r}^{(T_0)},\bxi_k\ra=\Theta(\sqrt{n}/\sigma_p \sqrt{d}),\quad v^{(T_0)}_{j,r,2}\cdot\sigma(\la\wb_{j,r}^{(T_0)},\bxi_k\ra)=\Theta\Big(\frac{\sigma_p \sqrt{d} \cdot v_0^2}{\sqrt{n}}\Big);\\
&v^{(T_0)}_{j,r,1}\cdot\sigma(\la\wb_{j,r}^{(T_0)},y_k\bmu\ra)=\Theta\Big(\frac{\sqrt{n} \|\bmu\|_2^2 v_0^2}{\sigma_p \sqrt{d}}\Big).
\end{align*}
\end{lemma}
Lemma \ref{lemma: overview_noise_balanced} points out that the noise memorization will grow faster and reach balancing state earlier than signal learning. From then on, the value of $v^{(t)}_{j,r,2}/\la\wb_{j,r}^{(t)},\bxi_k\ra$ will remain this order, which indicates two layers will jointly grow. Therefore, the growth rate of noise memorization will become more quadratic-like. We define all iterations from $T_0$ to the end of stage 1 as quadratic phase and have the following results:

\begin{lemma}[Noise Memorization becomes nearly constant]
\label{lemma: overview_noise_output_constant}
Under Condition \ref{condition: main condition}, if $|\ell'^{(t)}_i|=\Theta(1)$ for $\forall i \in [n]$, there exists a time $T_1$ that $T_0 < T_1= \tilde O\Big(\frac{\sqrt{n}}{\eta \sigma_p \sqrt{d}}\Big) \le T^{*,1}$, and for all $y_k=j$ we have:
\begin{itemize}
\item $v^{(T_1)}_{j,r,2}\cdot\sigma(\la\wb_{j,r}^{(T_1)},\bxi_k\ra)= \Theta(1)$;
\item 
$\la\wb_{j,r}^{(T_1)},y_k \bmu\ra / v^{(T_1)}_{j,r,1} = o(\|\bmu\|_2)$;
\item $v^{(T_1)}_{j,r,1}\cdot\sigma(\la\wb_{j,r}^{(T_1)},y_k \bmu\ra)=\tilde O\Big(\frac{\sqrt{n} \|\bmu\|_2^2 v_0^2}{\sigma_p \sqrt{d}}\Big)$.
\end{itemize}
\end{lemma}
Lemma \ref{lemma: overview_noise_output_constant} shows that when the noise memorization has reached constant level at the end of Stage 1, two layers of signal learning has not become balanced yet. This can be explained that once the noise part become balanced, its output will grow in a much faster rate than in the linear phase, so it will reach constant level in a relatively short time. We can summarize the results of two phases as the main proposition for stage 1:
\begin{proposition}[Small Initialization, Stage 1]
\label{prop: overview_stage1_s}
 Under Condition~\ref{condition: main condition}, with probability at least $1-\delta$ there exists $T_1 = \tilde O \Big(\frac{\sqrt{n}}{\eta \sigma_p \sqrt{d}}\Big)$ such that for every sample $k \in [n]$ with label $y_k = j$:
 \begin{enumerate}
     \item Noise memorization reaches constant level: $v^{(T_1)}_{y_k,r,2}\cdot\sigma(\la\wb_{y_k,r}^{(T_1)},\bxi_k\ra)= \Theta(1)$.
     \item Two layers of noise part become balanced: $\la\wb_{y_k,r}^{(T_1)},\bxi_k\ra / v^{(T_1)}_{y_k,r,2} = \Theta(\sigma_p \sqrt{d} / \sqrt{n})$ .
     \item Signal learning reaches certain level: $v^{(T_1)}_{y_k,r,1}\cdot\sigma(\la\wb_{y_k,r}^{(T_1)},y_k \bmu\ra)=\tilde \Theta\Big(\frac{\sqrt{n} \|\bmu\|_2^2 v_0^2}{\sigma_p \sqrt{d}}\Big)$.
     \item Two layers of signal part are not balanced: $ \la\wb_{y_k,r}^{(T_1)},y_k \bmu\ra / v^{(T_1)}_{y_k,r,1} = o(\|\bmu\|_2)$.
 \end{enumerate}
\end{proposition}
Note that Proposition \ref{prop: overview_stage1_s} is also based on the condition that $\ell'^{(t)}_i = \Theta(1)$ for $\forall i \in [n]$. Then we can adopt similar method as in the large initialization case to analyze Stage 2. We have the following proposition to estimate the scale of signal learning and noise memorization:
\begin{proposition}
\label{prop: overview_stage2_analysis_c2}
    For any $T_1 \le t \le T^*$, it holds that,
    \begin{align*}
        &0 \le v_{j,r,1}^{(t)} \le C_4 v_0 \alpha, \quad 0 \le \gamma_{j,r}^{(t)} \le \frac{ C_5 \sqrt{n}  v_0 \|\bmu\|_2^2 \alpha}{\sigma_p \sqrt{d}},\quad 0 \le \overline{\rho}_{j,r,k}^{(t)} \le \alpha,\\
        &0 \ge \underline{\rho}_{j,r,k}^{(t)} \ge -2\sqrt{\log(8mn/\delta)}\cdot \sigma_0 \sigma_p\sqrt{d} - 8 \sqrt{\frac{\log(4n^2/\delta)}{d}} n\alpha \ge -\alpha 
    \end{align*}
    for all $j \in \{\pm 1\}, r \in [m]$ and $k \in [n]$. Here $\alpha = 4\sqrt{\frac{\sigma_p \sqrt{d} \log(T^*)}{k_1 m \sqrt{n}}}$ where $k_1 = \min\limits_{k \in [n], t \le T_1} \Big\{\frac{v_{j,r,2}^{(t)} \sigma_p \sqrt{d}}{\sqrt{n}\la \wb_{j,r}^{(t)}, \bxi_k \ra}\Big\} = \Theta(1)$ and $C_4, C_5=\Theta(1)$.
\end{proposition}
Compared with Proposition \ref{prop: overview_stage2_analysis_c1}, Proposition \ref{prop: overview_stage2_analysis_c2} does not need to bound the value of $v_{j,r,2}^{(t)}$ as it can be directly calculated from balancing ratio of two layers. The remaining part to estimate test error is similar as the large initialization case and we defer the detailed proof to the appendix.

\section{Experiments}
\label{sec: experiment}

In this section, we present the empirical results on synthetic data to validate our theoretical analysis. 
Here we generate synthetic data following Definition \ref{def:data_model}. Since the learning problem is rotation-invariant, without loss of generality, we set $\bmu = \|\bmu\|_2 \cdot [1, 0, ..., 0]^{\top}$. Then we generate the noise vector $\bxi = [0, \bxi']^{\top}$ where $\bxi' \sim \mathcal{N}(0, \sigma_p^2 I_{d-1})$.

We consider the 2-layer CNN following the definition (\ref{eq: CNN_model}) with both trainable layers. To train the model, we set the number of filters $m=10$ and initialize the first layer using entry-wise Gaussian random initialization $\mathcal{N}(0, \sigma_0^2)$ with $\sigma_0 = 0.01$. Meanwhile, we set the learning rate to $0.01$ and run gradient descent for $T=200$ iterations. After training the model, we estimate the test error on $1000$ i.i.d test examples.

\paragraph{Different initialization scale.}

\begin{figure}[!t]
\centering 
\subfigure[different initialization scale]{
\label{fig: exp1_1}
\includegraphics[width=0.3\textwidth]{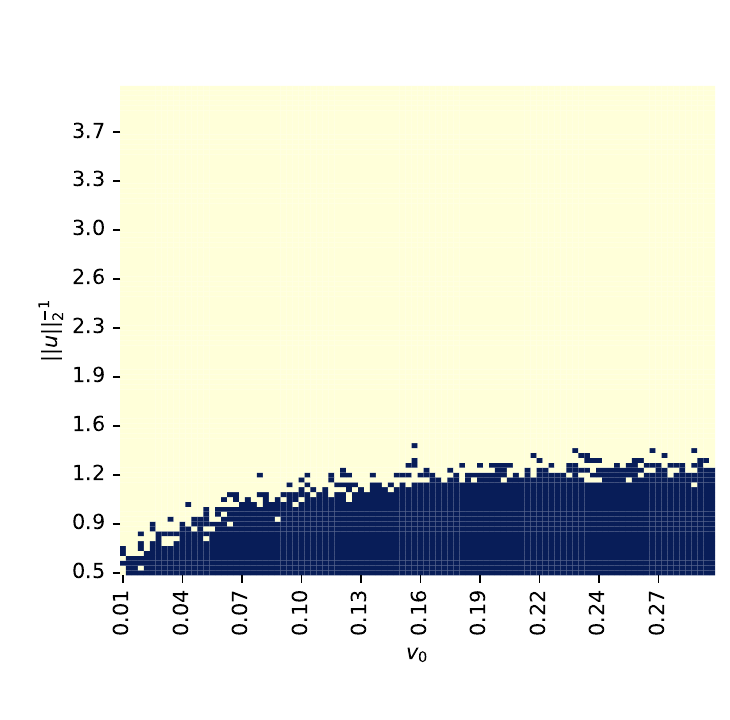}}
\subfigure[large initialization case]{
\label{fig: exp1_2}
\raisebox{5pt}{ 
\includegraphics[width=0.3\textwidth]{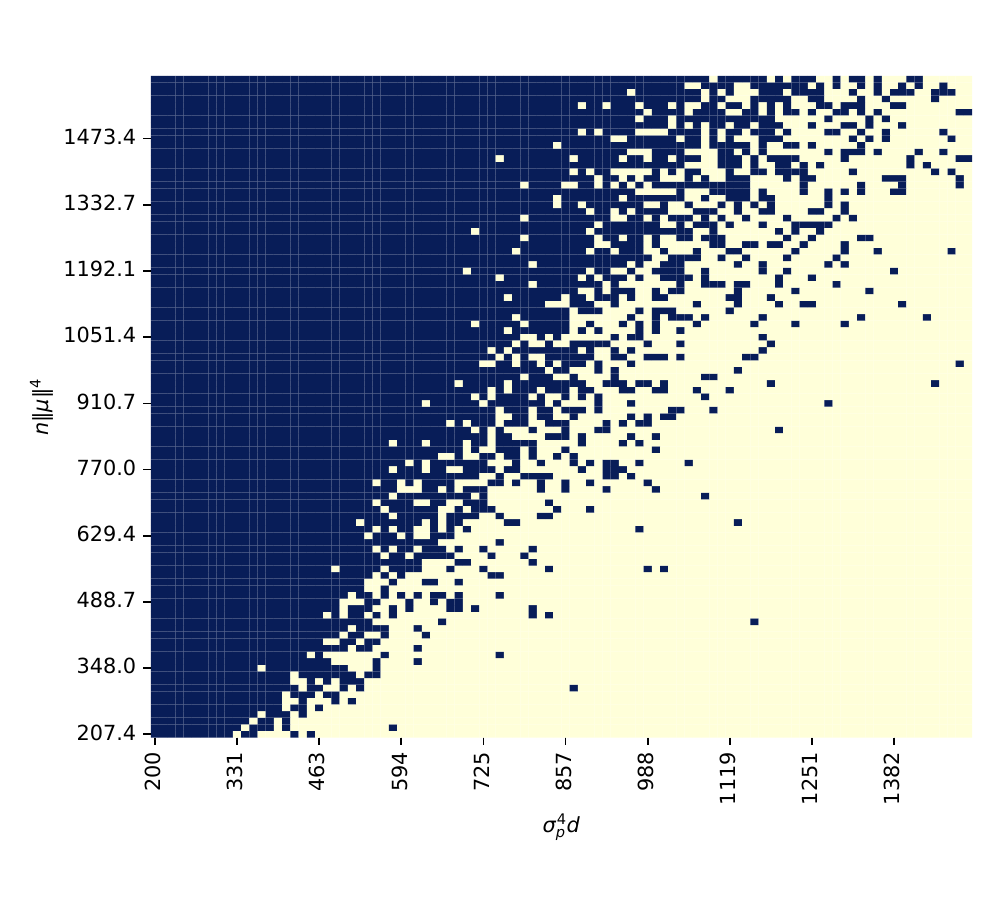}}}
\subfigure[small initialization case]{
\label{fig: exp1_3}
\includegraphics[width=0.3\textwidth]{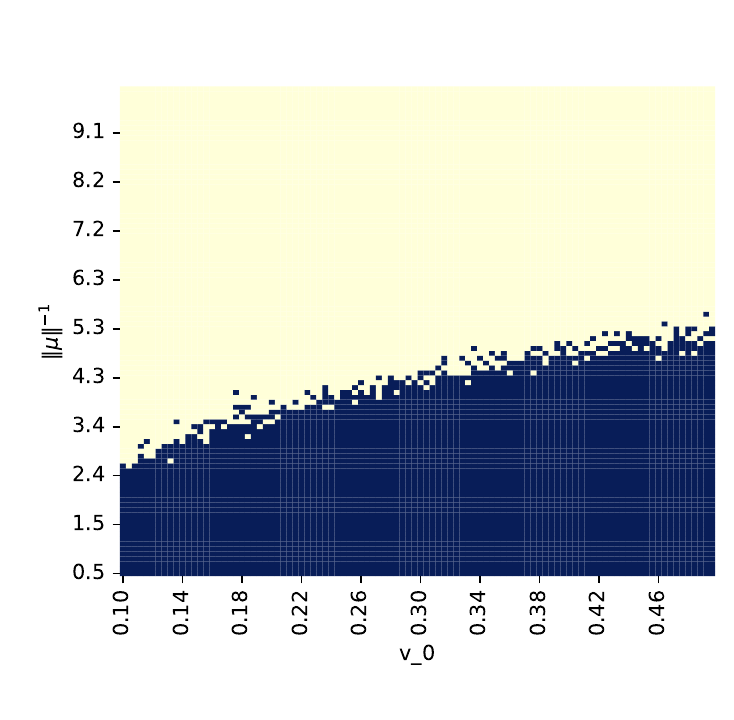}}
\caption{Figure~\ref{fig: exp1_1} is the truncated heatmap of test error on synthetic data under different $v_0$, where accuracy higher than 0.95 is colored blue, otherwise is colored yellow. The shape of the contour aligns with our theoretical prediction in Figure \ref{fig: sim}, indicating a phase transition between small $v_0$ and large $v_0$ as predicted by Theorem~\ref{thm: single_phase} and \ref{thm: double_phase}. Figure~\ref{fig: exp1_2} is the truncated heatmap of test error under a large fixed $v_0=5$ with varying $\|\bmu\|_2$ and $d$, where the test accuracy higher than 0.8 is colored blue, otherwise is colored yellow. The contours of the test accuracy are straight lines in the spaces $(\sigma_p^4 d, n \|\bmu\|_2^4)$ which validates Theorem \ref{thm: single_phase}. Figure~\ref{fig: exp1_3} is the truncated heatmap of test error under different $\|\bmu\|_2$ and $v_0$, where accuracy higher than 0.8 is colored blue, otherwise is colored yellow. The contours of the test accuracy are straight lines in the spaces $(v_0, \|\bmu\|_2^{-1})$ which validates Theorem \ref{thm: double_phase}.}
\label{fig: exp1}
\end{figure}

We first observe the dynamic of SNR boundary between benign and harmful overfitting as $v_0$ grows. According to Theorem \ref{thm: single_phase} and \ref{thm: double_phase}, there exists a phase transition when $v_0$ is at the threshold value $\Theta(n^{1/4}\sigma_p^{-1/2}d^{-1/4}m^{-1/2})$. When $v_0$ is above the threshold, the SNR boundary is $\frac{n\|\bmu\|_2^4}{\sigma_p^4 d} = \tilde\Theta(1)$, which indicates $\|\bmu\|_2$ remains invariant along this boundary as the output layer initialized with different value $v_0$. When $v_0$ is below this threshold, the SNR boundary becomes $\frac{m v_0^2 \|\bmu\|_2^2}{\sigma_p} = \tilde \Theta(1)$, implying that $1/\|\bmu\|_2$ will increase linearly with $v_0$. Therefore, in the test accuracy heatmap, we use horizontal axis to denote $v_0$ and vertical axis to denote the value $1/\|\bmu\|_2$. We fix $d=1000, n=100$, $\sigma_p=1$ and report the test accuracy for $v_0$ range from $0.01$ to $0.3$ while increasing $1/\|\bmu\|_2$ from $0.5$ to $4$ while remaining other parameters unchanged. The results are shown in Figure \ref{fig: exp1_1}.

\setlength{\abovecaptionskip}{2pt}
\setlength{\belowcaptionskip}{2pt}

Figure \ref{fig: exp1_1} perfectly corroborates our hypothesis presented in Figure \ref{fig: sim}. As shown in Figure \ref{fig: exp1_1}, there is a noticeable phase transition around \( v_0 = 0.1 \). When the value of $v_0$ is below this threshold, the contour of test accuracy is a straight line in the space $(v_0, \|\bmu\|_2^{-1})$. This observation is consistent with Theorem \ref{thm: double_phase}, which states that the SNR bound between benign and harmful overfitting is $\|\bmu\|_2^2/\sigma_p = \tilde \Theta(1/(mv_0^2))$, indicating that $v_0 \propto \|\bmu\|_2^{-1}$ on the boundary.  When the value of $v_0$ exceeds this threshold, the SNR bound maintains a consistent order of $\tilde \Theta(n\|\bmu\|_2^4/\sigma_p^4d)$, which is irrelevant with $v_0$.  This can be theoretically verified from Theorem \ref{thm: single_phase}, where the SNR bound is given by $\|\bmu\|_2^2/\sigma_p^4 = \tilde  \Theta(d/n)$, and empirically from the horizontal line in the latter half of the figure. The observation is clearly demonstrated with the truncated heatmap and aligns well with our theoretical results.

Next, we conduct experiments for $v_0$ with large and small initialization scale respectively.\\
\noindent\textbf{\underline{Large $v_0$ case.}} For the specific case $v_0=5$, we report the test accuracy between different choices of $\|\bmu\|_2$ and $d$. We fix $n=100$, $\sigma_p=1$ and change the y-axis by increasing $\|\bmu\|_2$ from $1.2$ to $2$ while changing the x-axis by increasing d from $200$ to $1500$. We convert the test accuracy to binary values based on a truncation threshold of 0.8. The results are shown in Figure \ref{fig: exp1_2}.

According to Figure \ref{fig: exp1_2}, increasing the signal strength $\|\bmu\|_2$ or decreasing the dimension \( d \) both lead to an increase in test accuracy. Furthermore, we can see from the heatmap that the contour of test accuracy form a straight line, indicating that $n\|\bmu\|_2^4 \propto \sigma_p^4 d$ on the SNR boundary. This matches the horizontal line for larger 
$v_0$ in Figure 1 and aligns with our results in Theorem \ref{thm: single_phase}.

\noindent\textbf{\underline{Small $v_0$ case.}} For $v_0$ with small initialization, as the SNR boundary is relevant with initialization scale $v_0$, we report the test accuracy between different choices of $\|\bmu\|_2$ and $v_0$. To better characterize their relation, we use the horizontal axis to denote $v_0$ and vertical axis to denote the value $1/\|\bmu\|_2$. We fix $d=1000$, $n=100$, $\sigma_p=0.1$ and change the y-axis by increasing $1/\|\bmu\|_2$ from $0.5$ to $10$ while increasing $v_0$ from $0.1$ to $0.5$. Similarly, we report the truncated heatmap based on the threshold of 0.8. The results are shown in Figure \ref{fig: exp1_3}.

According to Figure \ref{fig: exp1_3}, increasing the signal strength $\|\bmu\|_2$ or the initialization scale \( v_0 \) both lead to an increase in test accuracy. Additionally, the contour of test accuracy is a straight line in the space $(v_0, \|\bmu\|_2^{-1})$. This observation is consistent with Theorem \ref{thm: double_phase}, which states that the SNR bound between benign and harmful overfitting is $\|\bmu\|_2^2/\sigma_p = \Theta(1/(mv_0^2))$, indicating that $v_0 \propto \|\bmu\|_2^{-1}$ on the boundary.

\begin{figure}[!t]
\centering 
\subfigure[\scriptsize layer scale and ratio when $v_0=1$]{
\label{fig: exp4_0}
\includegraphics[width=0.315\textwidth]{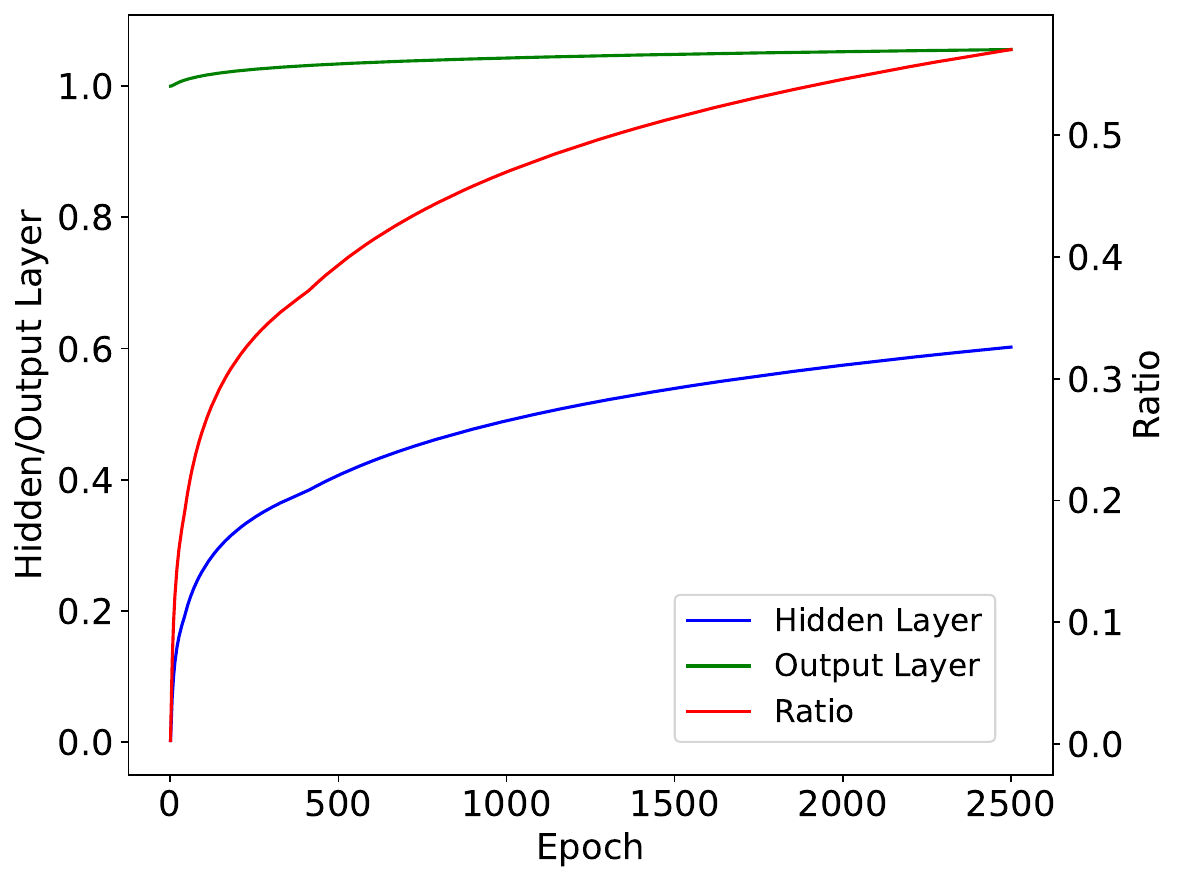}}
\subfigure[\scriptsize layer scale and ratio when $v_0=0.1$]{
\label{fig: exp4_1}
\includegraphics[width=0.32\textwidth]{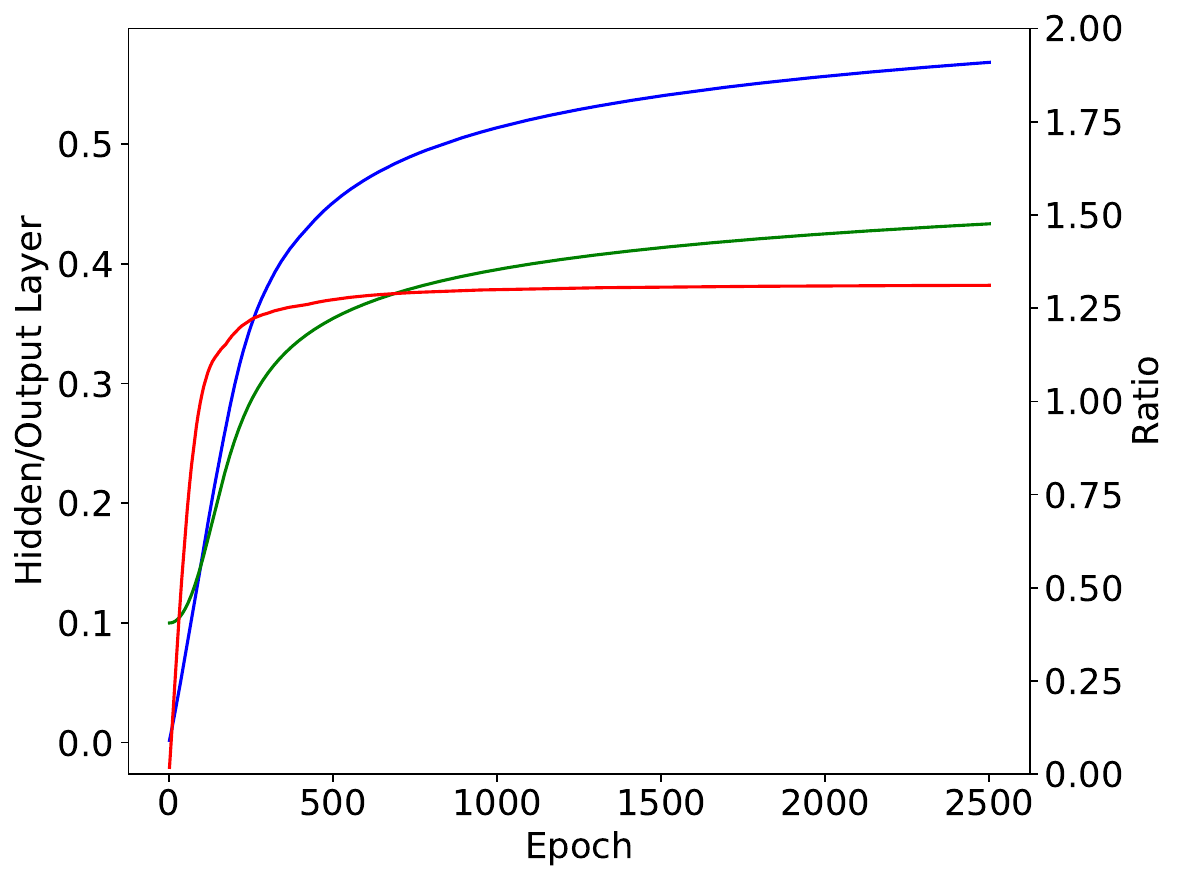}}
\subfigure[\scriptsize layer scale and ratio when $v_0=0.0005$]{
\label{fig: exp4_2}
\includegraphics[width=0.32\textwidth]{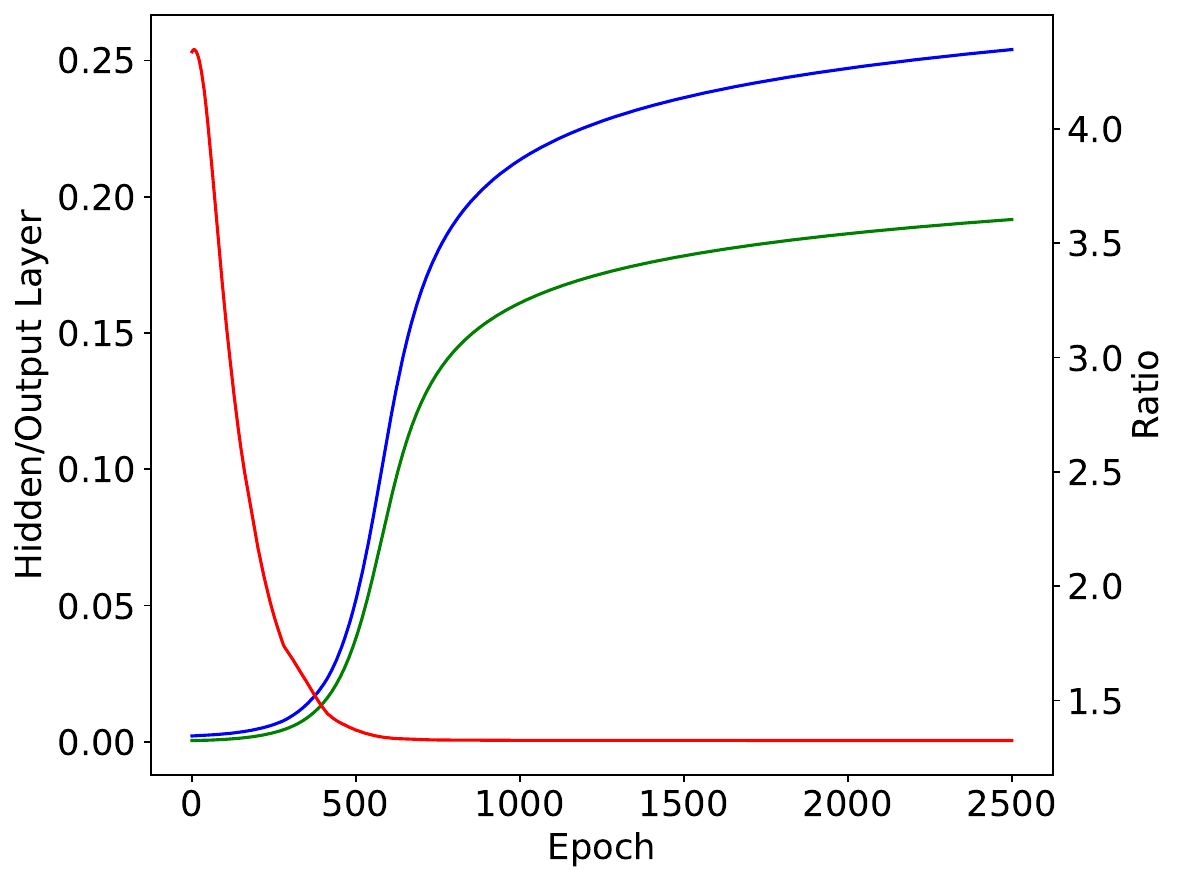}}
\caption{Scales of the hidden layer ($\max_{i \in [n]} \la\wb_{y_i,r}^{(t)}, \bxi_i\ra$, blue) and output layer ($\max_{i \in [n]} v^{(t)}{y_i,r,2}$, green) throughout the training process. These are referenced by the y-axis on the left. The red curve represents the ratio $\la\wb{y_i,r}^{(t)}, \bxi_i\ra / v^{(t)}_{y_i,r,2}$, referenced by the y-axis on the right. When $v_0$ is large (Figure \ref{fig: exp4_0}), the ratio of the two layers keeps growing, indicating an imbalance. When $v_0$ is smaller (Figures \ref{fig: exp4_1} and \ref{fig: exp4_2}), the scales of the two layers grow consistently over time, and their ratio converges to a constant value, validating the ``balanced'' results predicted by our theory.}
\label{fig: exp4}
\end{figure}

\paragraph{\textbf{Verification of the balancing state of output and hidden layers.}}
Finally, we conduct experiments to verify our detailed analysis in Section \ref{sec: overview_proof_tech}. According to Proposition \ref{prop: overview_stage1_s}, there exists a phase transition when the initialization scale of the output layer is small and the ratio of two layers(i.e.$\la\wb_{y_i,r}^{(t)},\bxi_i\ra / v^{(t)}_{y_i,r,2}$) will remain consistent in the second phase (i.e. quadratic phase). To demonstrate this, we fix $\|\bmu\|_2 = 1$, $d=1000, n=100, \sigma_p=0.2, \sigma_0=0.0001$ and plot the scale of hidden layer $\max_{i \in [n]}\la\wb_{y_i,r}^{(t)},\bxi_i\ra$ and output layer $\max_{i \in [n]}v^{(t)}_{y_i,r,2}$ for each iteration when $v_0$ is smaller than the initialization threshold to exhibit phase transition. We conduct experiments for both scenarios when the initialization of output layer is greater and smaller than the initialization threshold to exhibit phase transition. For the first scenario, we set $v_0=1$ and from Figure \ref{fig: exp4_0}, the scale of hidden layer grows dramatically while the output layer remains around its initialization value, so their ratio keeps growing. This is consistent with the second point in Proposition \ref{prop: overview_stage1_l}, which states that:

\begin{itemize} \item The two layers of the noise component are not balanced and $\la\wb_{y_k,r}^{(T_1)},\bxi_k\ra / v^{(T_1)}_{y_k,r,2} = o(\sigma_p \sqrt{d} / \sqrt{n})$. \end{itemize}

As a result, the two layers have not yet reached a balanced state, and their ratio continues to increase.
For the second scenario, we also conduct experiments for both cases when the output layer is larger than the hidden layer($v_0 = 0.1$) at initialization and the opposite case($v_0 = 0.0005$). As shown in Figure \ref{fig: exp4_1} and \ref{fig: exp4_2}, in both cases, two layers become ``balanced" and grow at a consistent rate when $t$ is sufficiently large. Their ratio also converges to a constant value. This also aligns with the second point in Proposition \ref{prop: overview_stage1_s} that:
\begin{itemize}
    \item Two layers of noise part become balanced: $\la\wb_{y_k,r}^{(T_1)},\bxi_k\ra / v^{(T_1)}_{y_k,r,2} = \Theta(\sigma_p \sqrt{d} / \sqrt{n})$ .
\end{itemize}
It is important to note that the convergence value depends only on $d$, $n$, and $\sigma_p$, and is independent of $v_0$. As we can see from figure \ref{fig: exp4_1} and \ref{fig: exp4_2}, their ratio converges to a similar value regardless of the different choices of $v_0$.
These results are well consistent with our theoretical analysis.

\section{Conclusion}
This paper investigates the benign overfitting for two-layer ReLU CNNs with fully trainable layers. We consider training the neural network using gradient descent on the cross-entropy loss. In our theory, we show that the initialization scale of the output layer significantly affects training dynamics: large scales lead to behavior similar to fixed-output models, while small scales result in more complex layer interactions. For both cases, we establish tight bounds on test errors, providing precise theoretical insights into the occurrence of benign overfitting. An important future direction is to explore benign overfitting in neural networks trained on
more complex data distributions, e.g., XOR data, and consider different optimizers such as Adam. Besides, it is also interesting to investigate more practical neural network models, such as the model with skip connection and the model with the normalization layer.
\bibliography{ref}

\newpage
\appendix
\section{The update formula}
\label{sec: update_formula}
In this section, we give the update formula. Specifically, in Section~\ref{subsec:calcu_gradient} we give the update formula of the gradient descent; in Section~\ref{subsec:signal_noise_decomp_formula}, we provide the analysis of signal-noise decomposition in $\wb_{j,r}^{(t)}$ which will further be applied in our analysis.

\subsection{Update formula of gradient descent}
\label{subsec:calcu_gradient}
We provide the calculation of the update rule in this section. Without loss of generality, we assume that the number of filters with index $j=+1$ is equal to the number of filters with index $j=-1$. Therefore, the neural network function, with respect to the $j$-th output, is formulated as follows:
\begin{align*}
F_j(\Wb_j,\vb_j;\bx) = \sum_{r=1}^m\sum_{p=1}^{2}v_{j,r,p}\sigma(\la\wb_{j,r},\bx^{(p)}\ra).
\end{align*}

With direct calculation, the update rule of $\wb_{j,r}^{(t)}$ and $v_{j,r,p}^{(t+1)}$ is given by
\begin{align*}
\wb_{j,r}^{(t+1)} &= \wb_{j,r}^{(t)} - \eta \nabla_{\wb_{j,r}} L_S(\Wb^{(t)})\notag\\
& = \wb_{j,r}^{(t)} - \frac{\eta }{n}\sum_{i=1}^n jy_i\ell'^{(t)}_{i}\cdot  \sum_{p=1}^2\big[v_{j,r,p}^{(t)}\cdot\sigma'(\la\wb_{j,r}^{(t)},\bx^{(p)}\ra)\cdot\bx_i^{(p)}\big].\notag\\
v_{j,r,p}^{(t+1)} &= v_{j,r,p}^{(t)} - \eta \nabla_{v_{j,r,p}} L_S(\Wb^{(t)}) = v_{j,r,p}^{(t)} - \frac{\eta }{n}\sum_{i=1}^n jy_i\ell'^{(t)}_{i}\cdot \sigma(\la\wb_{j,r}^{(t)},\bx_i^{(p)}\ra).
\end{align*}
Then, we consider the dynamics of $v_{j,r,p}^{(t)}\sigma(\la\wb_{j,r}^{(t)},\bx_k^{(p)}\ra)$. In particular, we have
\begin{align*}
\la\wb_{j,r}^{(t+1)},\ub\ra& = \la\wb_{j,r}^{(t)},\ub\ra - \frac{\eta }{n}\sum_{i=1}^n j y_i\ell'^{(t)}_{i}\cdot \sum_{p=1}^2\big[v_{j,r,p}^{(t)}\cdot\sigma'(\la\wb_{j,r}^{(t)},\bx^{(p)}\ra)\cdot\la\bx_i^{(p)},\ub\ra\big]\\
v_{j,r,p}^{(t+1)} &= v_{j,r,p}^{(t)} - \frac{\eta }{n}\sum_{i=1}^n jy_i\ell'^{(t)}_{j,i}\cdot  \sigma(\la\wb_{j,r}^{(t)},\bx_i^{(p)}\ra).
\end{align*}
Combining the  results above, for any  vector $\ub$ we have
\begin{align*}
&v_{j,r,q}^{(t+1)}\cdot\la\wb_{j,r}^{(t+1)},\ub\ra =v_{j,r,q}^{(t)}\cdot\la\wb_{j,r}^{(t)},\ub\ra - \frac{\eta}{n}\cdot\la\wb_{j,r}^{(t)},\ub\ra\cdot\sum_{i=1}^n \ell'^{(t)}_{i}\cdot jy_i\cdot \sigma(\la\wb_{j,r}^{(t)},\bx_i^{(q)}\ra)\notag\\
&\qquad -\frac{\eta}{n}\cdot v_{j,r,q}^{(t)}\cdot \sum_{i=1}^n \ell'^{(t)}_{i}\cdot jy_i\cdot \sum_{p=1}^2\big[v_{j,r,p}^{(t)}\cdot\sigma'(\la\wb_{j,r}^{(t)},\bx^{(p)}\ra)\cdot\la\bx_i^{(p)},\ub\ra\big]\notag\\
& \qquad +\frac{\eta^2}{n^2}\cdot\bigg[\sum_{i=1}^n\ell'^{(t)}_{i}\cdot jy_i\cdot \sigma(\la\wb_{j,r}^{(t)},\bx_i^{(q)}\ra)\bigg]\cdot \bigg[\sum_{i=1}^n \ell'^{(t)}_{i}\cdot jy_i\sum_{p=1}^2\big[v_{j,r,p}^{(t)}\cdot\sigma'(\la\wb_{j,r}^{(t)},\bx^{(p)}\ra)\cdot\la\bx_i^{(p)},\ub\ra\big]\bigg].
\end{align*}

\subsection{Signal-noise decomposition}
\label{subsec:signal_noise_decomp_formula}
In this section, we analyze the dynamics of the coefficients in the signal-noise decomposition. We first give the following definition:
\begin{definition}
\label{def: w_decompose}
Let $\wb_{j,r}^{(t)}$ for $j \in \{\pm 1\}$, $r \in [m]$ be the convolution filters of the CNN at the t-th iteration. Then there exist unique coefficients $\gamma_{j,r}^{(t)}$ and $\rho_{j,r,i}^{(t)}$ such that 
\begin{align*}
    \wb_{j,r}^{(t)} = \wb_{j,r}^{(0)} + j \cdot \gamma_{j,r}^{(t)}\cdot \|\bmu\|_2^{-2} \cdot \bmu + \sum_{i=1}^n \rho_{j,r,i}^{(t)} \cdot \|\bxi_i\|_2^{-2} \cdot \bxi_i.
\end{align*}
Further denote $\overline{\rho}_{j,r,i}^{(t)} := \rho_{j,r,i}^{(t)} \mathbb{I}(\rho_{j,r,i}^{(t)} \ge 0)$, $\underline{\rho}_{j,r,i}^{(t)} := \rho_{j,r,i}^{(t)} \mathbb{I}(\rho_{j,r,i}^{(t)} \le 0)$. Then we have
\begin{align*}
    \wb_{j,r}^{(t)} = \wb_{j,r}^{(0)} + j \cdot \gamma_{j,r}^{(t)}\cdot \|\bmu\|_2^{-2} \cdot \bmu + \sum_{i=1}^n \overline{\rho}_{j,r,i}^{(t)} \cdot \|\bxi_i\|_2^{-2} \cdot \bxi_i + \sum_{i=1}^n \underline{\rho}_{j,r,i}^{(t)} \cdot \|\bxi_i\|_2^{-2} \cdot \bxi_i.
\end{align*}
\end{definition}

Based on this definition, we give the following lemma to characterize the dynamics of the decomposition coefficients:
\begin{lemma}
\label{lemma: coef_update}
    The coefficients $\gamma_{j,r}^{(t)}, \overline{\rho}_{j,r,i}^{(t)}$ and $\underline{\rho}_{j,r,i}^{(t)}$ satisfy the following iterative equations:
    \begin{align*}
        &\gamma_{j,r}^{(0)}, \overline{\rho}_{j,r,i}^{(0)}, \underline{\rho}_{j,r,i}^{(0)} = 0\\
        &\gamma_{j,r}^{(t+1)} = \gamma_{j,r}^{(t)} - \frac{\eta}{n} v_{j,r,1}^{(t)} \|\bmu\|_2^2 \sum_{i=1}^n \ell'^{(t)}_{i} \sigma'(\la \wb_{j,r}^{(t)}, y_i \bmu \ra)\\
        &\overline{\rho}^{(t+1)}_{j,r,i} = \overline{\rho}_{j,r,i}^{(t)} - \frac{\eta}{n} v_{j,r,2}^{(t)} \cdot \ell'^{(t)}_{i} \cdot \sigma'(\la \wb_{j,r}^{(t)}, \bxi_i \ra) \cdot \|\bxi_i\|_2^2 \cdot \mathbb{I}(y_i = j)\\
        &\underline{\rho}^{(t+1)}_{j,r,i} = \underline{\rho}_{j,r,i}^{(t)} + \frac{\eta}{n} v_{j,r,2}^{(t)} \cdot \ell'^{(t)}_{i} \cdot \sigma'(\la \wb_{j,r}^{(t)}, \bxi_i \ra) \cdot \|\bxi_i\|_2^2 \cdot \mathbb{I}(y_i = -j)
    \end{align*}
\end{lemma}

\begin{proof}[Proof of Lemma \ref{lemma: coef_update}]
    Recall the update equation of $\wb_{j,r}^{(t)}$:
    \begin{align*}
        \wb_{j,r}^{(t+1)} &= \wb_{j,r}^{(t)} - \eta \nabla_{\wb_{j,r}} L_S(\Wb^{(t)}, \btheta^{(t)})\\
        &= \wb_{j,r}^{(t)} - \frac{\eta}{n} v_{j,r,2}^{(t)} \sum_{i=1}^n \ell'^{(t)}_i \sigma'(\la \wb_{j,r}^{(t)}, \bxi_i \ra) \cdot j y_i \bxi_i - \frac{\eta}{n} v_{j,r,1}^{(t)} \sum_{i=1}^n \ell'^{(t)}_i \sigma'(\la \wb_{j,r}^{(t)}, y_i \bmu \ra) \cdot j \bmu\\
        &= \wb_{j,r}^{(0)} - \frac{\eta}{n} \sum_{s=0}^t \sum_{i=1}^n v_{j,r,2}^{(s)}  \ell'^{(s)}_i \sigma'(\la \wb_{j,r}^{(s)}, \bxi_i \ra) \cdot j y_i \bxi_i - \frac{\eta}{n} \sum_{s=0}^t \sum_{i=1}^n v_{j,r,1}^{(s)}  \ell'^{(s)}_i \sigma'(\la \wb_{j,r}^{(s)}, y_i \bmu \ra) \cdot j \bmu.
    \end{align*}
    According to Definition \ref{def: w_decompose}, we can get the unique representation of $\gamma_{j,r}^{(t)}$ and $\rho_{j,r,i}^{(t)}$ that
    \begin{align*}
        \gamma_{j,r}^{(t)} &= - \frac{\eta}{n} \sum_{s=0}^t \sum_{i=1}^n v_{j,r,1}^{(s)}  \ell'^{(s)}_i \sigma'(\la \wb_{j,r}^{(s)}, y_i \bmu \ra) \cdot \|\bmu\|_2^2,\\
        \rho_{j,r,i}^{(t)} &= -\frac{\eta}{n} \sum_{s=0}^t \sum_{i=1}^n v_{j,r,2}^{(s)}  \ell'^{(s)}_i \sigma'(\la \wb_{j,r}^{(s)}, \bxi_i \ra) \cdot j y_i \|\bxi_i\|_2^2.\\
    \end{align*}
    Moreover, with the notation $\overline{\rho}_{j,r,i}^{(t)} := \rho_{j,r,i}^{(t)} \mathbb{I}(\rho_{j,r,i}^{(t)} \ge 0)$, $\underline{\rho}_{j,r,i}^{(t)} := \rho_{j,r,i}^{(t)} \mathbb{I}(\rho_{j,r,i}^{(t)} \le 0)$, we have
    \begin{align*}
        \overline{\rho}_{j,r,i}^{(t)} &= -\frac{\eta}{n} \sum_{s=0}^t \sum_{i=1}^n v_{j,r,2}^{(s)}  \ell'^{(s)}_i \sigma'(\la \wb_{j,r}^{(s)}, \bxi_i \ra) \cdot \|\bxi_i\|_2^2 \mathbb{I}(y_i = j),\\
        \underline{\rho}_{j,r,i}^{(t)} &= -\frac{\eta}{n} \sum_{s=0}^t \sum_{i=1}^n v_{j,r,2}^{(s)}  \ell'^{(s)}_i \sigma'(\la \wb_{j,r}^{(s)}, \bxi_i \ra) \cdot \|\bxi_i\|_2^2 \mathbb{I}(y_i = -j).
    \end{align*}
    Therefore, we can get the single-step update equation of $\gamma_{j,r}^{(t)}, \overline{\rho}_{j,r,i}^{(t)}$ and $\underline{\rho}_{j,r,i}^{(t)}$ to complete the proof.
\end{proof}

\section{Preliminary Lemmas}
In this section, we will give some technical lemmas which are applied in our analysis. 
\subsection{Concentration lemmas}
In this section,  we present some key lemmas that give some important properties of the data and the neural network parameters at their random initialization.

The first lemma estimates the norms of the noise vectors $\bxi_k$,$k \in [n]$, and bounds their inner products with each other. The proof is similar to that presented in \cite{kou2023benign} and \cite{meng2024benign} We provide the proof here for the convenience of readers. 
\begin{lemma}
\label{lemma: noise bound}
    Suppose that $\delta > 0$ and $d=\Omega(\log(4n/\delta))$. Then with probability at least $1-\delta$,
    \begin{align*}
    &\sigma_p^2 d/2 \le \|\bxi_i\|_2^2 \le 3\sigma_p^2 d/2,\\
    &|\la \bxi_i,\bxi_j \ra | \le 2 \sigma_p^2 \cdot \sqrt{d \log(4n^2/\delta)}
    \end{align*}
    for all $i,j \in [n].$
\end{lemma}

\begin{proof}[Proof of Lemma \ref{lemma: noise bound}]
By Bernstein’s inequality, with probability at least $1-\delta/(2n)$ we have
\begin{align*}
    \big|\|\bxi_k\|_2^2-\sigma_p^2d \big| = O\big(\sigma_p^2 \cdot \sqrt{d\log(4n/\delta)}\big).
\end{align*}
Therefore, if we set appropriately $d=\Omega(\log(4n/\delta))$, we get
\begin{align*}
    \sigma_p^2 d/2 \le \|\bxi_k\|_2^2 \le 3\sigma_p^2 d/2.
\end{align*}
Moreover, clearly $\la \bxi_i, \bxi_j \ra$ has mean zero. For any i,j with $i \neq j$, by Bernstein’s inequality, with probability at least $1-\delta/(2n)^2$ we have
\begin{align*}
    |\la \bxi_i,\bxi_j \ra | \le 2 \sigma_p^2 \cdot \sqrt{d \log(4n^2/\delta)}
\end{align*}
Applying a union bound completes the proof.
\end{proof}

Next, define $S_{j,r}^{(t)}:=\{i\in [n]:y_i=j,\la \wb_{j,r}^{(t)}, \bxi_i \ra > 0\}$, $S_i^{(t)} := \{r \in [m]: \la w_{y_i,r},\bxi_k \ra >0\}$. We have the following lemma characterizing their sizes.
\begin{lemma}
\label{lemma: preliminary_S_jr_intialization}
    Suppose that $\delta > 0$ and $n \ge 32 \log(4m/\delta)$. Then with probability at least $1-\delta$, for all $j \in \{\pm 1\}$, $r \in [m]$ it holds that
    \begin{align*}
        \big|S_{j,r}^{(0)}\big| \ge n/8.
    \end{align*}
\end{lemma}

\begin{proof}[Proof of Lemma \ref{lemma: preliminary_S_jr_intialization}] 
Note that $\big|S_{j,r}^{(0)}\big| = \sum_{i=1}^{n} \mathbb{I}(y_i=j) \cdot \mathbb{I}(\la \wb^{(t)}_{j,r}, \bxi_i \ra \ge 0)$. And $\mathbb{P}(y_i=j, \la \wb^{(t)}_{j,r}, \bxi_i \ra \ge 0) = 1/4$. So by Hoeffding’s inequality, with probability at least $1-\delta/(2m)$, we have
\begin{align*}
    \big| |S_{j,r}^{(0)}| - n/4 \big| \le \sqrt{\frac{n\log(4m/\delta)}{2}}
\end{align*}
Therefore, as $n \ge C\log(m/\delta)$, by applying union bound, we have with probability at least $1-\delta$ for all $j \in \{\pm 1\}$, $r \in [m]$,
    \begin{align*}
        \big|S_{j,r}^{(0)}\big| \ge n/8.
    \end{align*}
\end{proof}

\begin{lemma}
\label{lemma: preliminary_S_i_initialization}
    Suppose that $\delta > 0$ and $m \ge 50 \log(2n/\delta)$. Then with probability at least $1 - \delta$, for all $i \in [n]$
    \begin{align*}
        |S_{i}^{(0)}| \ge 0.4m.
    \end{align*}
\end{lemma}

\begin{proof}[Proof of Lemma \ref{lemma: preliminary_S_i_initialization}]
    Note that $|S_i^{(0)}| = \sum_{r=1}^m \mathbb{I}(\la \wb_{y_i, r}^{(0)}, \bxi_i \ra > 0)$ and $\mathbb{P}(\la \wb_{y_i, r}^{(0)}, \bxi_i \ra > 0) = 1/2$. So by Hoeffding’s inequality, with probability at least $1-\delta/(n)$, we have
    \begin{align*}
        \bigg|\frac{|S_i^{(0)}}{m} - \frac{1}{2} \bigg| \le \sqrt{\frac{\log(2n/\delta)}{2m}}.
    \end{align*}
    Therefore, as $m \ge 50 \log(2n/\delta)$, by applying union bound, we have with probability at least $1-\delta$ for all $i \in [n]$
    \begin{align*}
        |S_{i}^{(0)}| \ge 0.4m.
    \end{align*}
\end{proof}

Then the following lemma bounds the inner product between a randomly initialized CNN $\wb_{j,r}^{(0)}$ and the signal in the training data.
\begin{lemma}
\label{lemma: preliminary_initial inner product}
    Suppose that $d \ge \Omega(\log(mn/\delta)), m=\Omega(\log(1/\delta))$. Then with probability at least $1-\delta$,
    \begin{align*}
        &1/2 \sigma_0^2d \le \|\wb_{j,r}^{(0)}\| \le 3/2 \sigma_0^2 d\\
        &|\la \wb_{j,r}^{(0)}, \bmu \ra| \le \sqrt{2\log(8m/\delta)} \cdot \sigma_0\|\bmu\|_2,\\
        &|\la \wb_{j,r}^{(0)}, \bxi_i \ra| \le 2\sqrt{\log(8mn/\delta)}\cdot \sigma_0 \sigma_p\sqrt{d}
    \end{align*}
    for all $r\in [m], j\in\{\pm 1\}$ and $i\in [n]$.
\end{lemma}

\begin{proof}[Proof of Lemma \ref{lemma: preliminary_initial inner product}]
First of all, as $\wb \sim \mathcal{N}(0, \sigma_0 I)$, by Bernstein’s inequality, with probability at least $1 - \delta/(4m)$ we have
\begin{align*}
    |\|\wb_{j,r}^{(0)}\|_2^2 - \sigma_0^2 d| = O(\sigma_0^2 \cdot \sqrt{d \log(8m/\delta)}).
\end{align*}
Therefore, if we set appropriately $d = \Omega(\log(mn/\delta))$, we have with probability at least $1 - \delta/3$, for all $j \in \|\pm 1\}$ and $r \in [m]$,
\begin{align*}
    1/2 \sigma_0^2d \le \|\wb_{j,r}^{(0)}\| \le 3/2 \sigma_0^2 d.
\end{align*}

Next, it is clear that for each $r \in [m]$, $j\la \wb_{j,r}^{(0)}, \bmu \ra$ is a Gaussian random variable with mean zero and variance $\sigma_0^2 \|\bmu\|_2^2$. Therefore, by Gaussian tail bound and union bound, with probability at least $1-\delta/4$, for all $r\in [m], j\in\{\pm 1\}$,
\begin{align*}
    j\la \wb_{j,r}^{(0)}, \bmu \ra \le |\la \wb_{j,r}^{(0)}, \bmu \ra| \le \sqrt{2 \log(8m/\delta)} \cdot \sigma_0 \|\bmu\|_2.
\end{align*}
Similarly, based on the results of Lemma \ref{lemma: noise bound}, we have for $\la \wb_{j,r}^{(0)}, \bxi_i \ra$
 \begin{align*}
      |\la \wb_{j,r}^{(0)}, \bxi_i \ra| &\le 2\sqrt{\log(8mn/\delta)}\cdot \sigma_0 \sigma_p\sqrt{d}.
 \end{align*}
\end{proof}

\subsection{Technical lemmas}
\label{sec: tech_lemmas}
As both layers are trainable in our setting, the change of one layer will impact the other, so we need to consider their dynamics jointly. 
In the following, we give some technical lemmas to describe the dynamic of two intertwined sequences, which will be used in our main proof later.

\begin{restatable}[Dynamic of two intertwined sequence]{lemma}{lemmaseq}\label{lemma: interwined_sequence}
Consider two sequences $\{a_t\}$ and $\{b_t\}$ which satisfy
\begin{align*}
a_{t+1} &= a_t + A\cdot b_t,\\
b_{t+1} &= b_t + B\cdot a_t,
\end{align*}
where A,B are constant irrelevant with $t$ and satisfy $ AB \le 1$ and $a_0/b_0 = o(\sqrt{A/B})$. Then there exists an iteration $t_1=\Theta(1/\sqrt{AB})$ such that 
\begin{itemize}
    \item $a_{t_1}/b_{t_1} = \Theta(\sqrt{A/B});$
    \item $a_{t_1} = \Theta(\sqrt{A/B}\cdot b_0);$
    \item $b_{t_1} = \Theta(b_0).$
\end{itemize}
The first conclusion can also be interpreted as two sequences becoming "balanced".
\end{restatable}

\begin{proof}
First, we define the ratio of two sequences' elements with the same index, $a_t/b_t$ as the balancing factor. We could derive the one-step update equation of the balancing factor:
\begin{align*}
\frac{a_{t+1}}{b_{t+1}} &= \frac{a_t + A\cdot b_t}{b_t + B\cdot a_t} = \frac{\frac{a_t}{b_t} + A}{1 + B \cdot \frac{a_t}{b_t}}.
\end{align*}
So the characteristic equation of the sequence is:
\begin{align*}
    x = \frac{x + A}{1 + Bx}.
\end{align*}
Solving the equation, we could get the limit of balancing factor $\lim \limits_{t\rightarrow \infty}a_{t}/b_t = \sqrt{A/B}$.

Then, let $t_1$ be an iteration such that $b_t\in[2b_0, 3b_0]$, then we have for all $t\le t_1$, it holds that 
\begin{align*}
a_t\le a_0 + t_1\cdot A\cdot 3b_0.
\end{align*}
So it follows that for any $t\le t_1$,
\begin{align*}
b_t&\le b_0 + B\cdot t_1\cdot (a_0 + t_1\cdot A\cdot  3b_0)\\
&\le b_0 + B \cdot t_1 \cdot \Theta(t_1 \cdot A \cdot b_0) \\
&\le b_0 + \Theta(AB\cdot t_1^2 \cdot b_0),
\end{align*}
where the second inequity is due to $a_0/b_0=o(A)$.
As $2b_0 \le b_t \le 3b_0$, we have $t_1=\Omega(1/\sqrt{AB})$.

Similarly, for $t\in[t_1/2, t_1]$ it holds that
\begin{align*}
a_t\ge a_0 + b_0\cdot t_1/2\cdot A,
\end{align*}
implying that 
\begin{align*}
b_t &\ge b_0 + B\cdot t_1/2\cdot (a_0 + b_0\cdot t_1/2\cdot A)\\
&\ge b_0 + B\cdot t_1/2 \cdot b_0 \cdot t_1/2\cdot A\\
&= b_0 + AB \cdot b_0^2 \cdot t_1^2/4.
\end{align*}
Same as above, this further implies that $t_1 = O(1/\sqrt{AB})$.
Combining the above two results, we know that $t_1=\Theta(1/\sqrt{AB})$.
Meanwhile, consider the scale of $a_{t_1}$:
\begin{align*}
a_{t_1} &= a_0 + \Theta(A \cdot t_1 \cdot b_0)\\
&= a_0 + \Theta(\sqrt{A/B}\cdot b_0)\\
&= \Theta(\sqrt{A/B}\cdot b_0),
\end{align*}
where the last equity is due to $a_0/b_0 = o(A)$ and $AB \le 1$.
Therefore, the balancing factor $a_{t_1}/b_{t_1} = \Theta(\sqrt{A/B})$ and will remain at this level from then on, which enters the balancing stage.
\end{proof}

Then we give the lemma to compare two sets of intertwined sequences.
\begin{restatable}[Comparison of two sequences]{lemma}{lemmacompare}\label{lemma: compare_sequence}
Consider the following two sets of sequences:
\begin{equation*}
\begin{cases}
    \overline{a}_{t+1} = \overline{a_t} + \overline{A} \cdot \overline{b_t}\\
    \overline{b}_{t+1} = \overline{b_t} + \overline{B} \cdot \overline{a_t}
\end{cases}
\end{equation*}

\begin{equation*}
\begin{cases}
    \underline{a}_{t+1} = \underline{a_t} + \underline{A} \cdot \underline{b_t}\\
    \underline{b}_{t+1} = \underline{b_t} + \underline{B} \cdot \underline{a_t}
\end{cases}
\end{equation*}
where $\overline{A}, \overline{B}, \underline{A}, \underline{B}$ are constants irrelevant with t, and $\underline{A} = O(\overline{A})$, $\underline{B} = O(\overline{B})$. If $\overline{a_0} = O(\underline{a_0})$, $\overline{b_0} = O(\underline{b_0})$, then we have
\begin{itemize}
    \item $\underline{a_t} = O(\overline{a_t})$;
    \item $\underline{b_t} = O(\overline{b_t})$.
\end{itemize}
\end{restatable}

\begin{proof}
We use induction to prove the lemma. The conclusion holds naturally at time 0. Suppose that the conclusion also holds at all $0 \le t \le \tilde t-1$, now we prove for the time when $t = \tilde t$:
\begin{align*}
\underline{a}_{\tilde t} &= \underline{a}_{\tilde t-1} + \underline{A} \cdot \underline{b}_{\tilde t-1}\\
&\le \underline{a}_{\tilde t-1} + \Theta(\overline{A}) \cdot \underline{b}_{\tilde t-1}\\
&= O(\overline{a}_{\tilde t-1} + \overline{A}\cdot \overline{b}_{\tilde t-1}) = O(\overline{a}_{\tilde t}).
\end{align*}
The second inequity is due to $\underline{A} \le \Theta(\overline{A})$ and the second equality is from the induction hypothesis. Similarly, we have
\begin{align*}
\underline{b}_{\tilde t} &= \underline{b}_{\tilde t-1} + \underline{B} \cdot \underline{a}_{\tilde t-1}\\
&\le \overline{b}_{\tilde t-1} +\Theta(B \cdot \overline{a}_{\tilde t-1})\\
&= O(b_{\tilde t-1} + B\cdot a_{\tilde t-1}) = O(b_{\tilde t}).
\end{align*}
The second inequity is due to $\underline{B} \le \Theta(\overline{B})$ and the second equality is from the induction hypothesis.
Therefore, the conclusion also holds at time $\tilde t$.
\end{proof}

\section{Proof of Large Initialization}
In this section, we consider the case when the output layer initialization $v_0 \ge \Theta(n^{1/4}\sigma_p^{-1/2}d^{-1/4}m^{-1/2})$.

To better understand the whole training process, we decouple it into two stages to better analyze the complex dynamics of the signal and noise coefficients.
\begin{itemize}
    \item Stage 1: $\ell'^{(t)}_{i} = \Theta(1)$

     During the first stage, as we have a small initialization, the output of the model is also small. So $\ell'^{(t)}_i = \Theta(1)$ for all $i \in [n]$. We can show that the dynamic of signal learning and noise memorization will be different depending on the initialization scale of output layer.

     \item Stage 2: $L_S \rightarrow 0$

     At the end of stage 1, the output of noise part has already reached constant order so the training loss will be small. Based on the results of stage 1, we can analyze the convergence of training loss and the test error of trained model.
\end{itemize}

\subsection{First stage}
In Stage 1, consider the training period $0 \le t \le T^{*,1}$, and define $T^{*,1}$ as:
\begin{equation}
\label{eq: define_T1}
    T^{*,1}= \max\limits_{k,j,r}\bigg\{t: \sum_{r=1}^m v^{(t)}_{j,r,2}\cdot\sigma(\la\wb_{j,r}^{(t)},\bxi_k\ra) \le 0.1\bigg\}
\end{equation}
First we present the main proposition during Stage 1.

\begin{proposition}(Large Initialization, Stage 1)
\label{prop: stage1_l}
    Under Condition \ref{condition: main condition}, with probability at least $1-\delta$ there exists $T_1 = \Theta\Big(\frac{n}{\eta \sigma_p^2 d m v_0^2}\Big)$ such that for every sample $k \in [n]$ the following hold:
    \begin{enumerate}
        \item Noise memorization reaches constant level: $\sum_{r=1}^m v^{(T_1)}_{y_k,r,2}\cdot\sigma(\la\wb_{y_k,r}^{(T_1)},\bxi_k\ra)= 0.05$.
        \item Two layers of noise part are not balanced: $\la\wb_{y_k,r}^{(T_1)},\bxi_k\ra / v^{(T_1)}_{y_k,r,2} = o(\sigma_p \sqrt{d} / \sqrt{n})$.
        \item Signal learning reaches specific level: $\sum_{r=1}^m v_{y_k,r,1}^{(T_1)} \la\wb_{y_k,r}^{(T_1)},y_k \bmu\ra = \Theta\Big(\frac{n \|\bmu\|_2^2 }{\sigma_p^2 d}\Big)$.
        \item Two layers of signal part are not balanced: $ \la\wb_{y_k,r}^{(T_1)},y_k \bmu\ra / v^{(T_1)}_{y_k,r,1} = o(\|\bmu\|_2)$.
    \end{enumerate}
\end{proposition}

We first introduce a lemma to characterize the dynamic of output layer:
\begin{lemma}
\label{lemma: positive_v}
Under Condition \ref{condition: main condition}, if $|\ell'^{(t)}_i| \in [0.4, 0.6]$ for any $0 \le t \le T^{*,1}$, $i \in [n]$, we have that
\begin{align}
    &v_{j,r,2}^{(t)} \ge 0.5 \cdot v_0.\label{eq: v_bound}
\end{align}
\end{lemma}

\begin{proof}[Proof of Lemma \ref{lemma: positive_v}]
    We consider two cases: $0 \le t \le T_0$ and $T_0 \le t \le T^{*,1}$, where $T_0 = \frac{0.5 v_0}{\eta \sqrt{\log(8mn/\delta)}\cdot \sigma_0 \sigma_p\sqrt{d}}$. For the first case, from signal-noise decomposition analysis, for $y_k \neq j$ we have
    \begin{align*}
        \la \wb_{j,r}^{(t-1)}, \bxi_k \ra &= \la \wb_{j,r}^{(0)}, \bxi_k \ra + \underline{\rho}_{j,r,k}^{(t-1)} +\sum_{i=1, i\neq k}^n \rho_{j,r,i}^{(t-1)} \frac{\la \bxi_i, \bxi_k \ra}{\|\bxi_i\|_2^2}.
    \end{align*}
    Recall the update equation of $v_{j,r,2}^{(t)}$:
\begin{align*}
    v_{j,r,2}^{(t)}  =& v_{j,r,2}^{(t-1)} - \frac{j \eta}{n}\sum_{i=1}^{n}{\ell'^{(t-1)}_i}y_i\la \wb_{j,r}^{(t-1)},\bxi_i \ra \mathbb{I}(\la\wb_{j,r}^{(t-1)},\bxi_i\ra >0)\\
    =& v_{j,r,2}^{(t-1)} +\frac{\eta}{n} \sum_{k=1,y_k\neq j}^n {\ell'^{(t-1)}_k} \la \wb_{j,r}^{(t-1)},\bxi_k \ra \mathbb{I}(\la\wb_{j,r}^{(t-1)},\bxi_k\ra >0) \\
    &- \frac{\eta}{n} \sum_{k=1,y_k=j}^n {\ell'^{(t-1)}_k} \la \wb_{j,r}^{(t-1)},\bxi_k \ra \mathbb{I}(\la\wb_{j,r}^{(t-1)},\bxi_k\ra >0)\\
    \ge& v_{j,r,2}^{(t-1)} - \frac{\eta}{n} \sum_{k=1, y_k \neq j}^n 0.6 \bigg(\la \wb_{j,r}^{(0)}, \bxi_k \ra + \underline{\rho}_{j,r,k}^{(t-1)} +\sum_{i=1, i\neq k}^n \rho_{j,r,i}^{(t-1)} \frac{\la \bxi_i, \bxi_k \ra}{\|\bxi_i\|_2^2}\bigg)\\
    &+ \frac{\eta}{n} \sum_{k=1, y_k = j}^n 0.4 \bigg(\la \wb_{j,r}^{(0)}, \bxi_k \ra + \overline{\rho}_{j,r,k}^{(t-1)} +\sum_{i=1, i\neq k}^n \rho_{j,r,i}^{(t-1)} \frac{\la \bxi_i, \bxi_k \ra}{\|\bxi_i\|_2^2}\bigg)\\
    \ge& v_{j,r,2}^{(t-1)} - \frac{\eta}{n} \sum_{k=1, y_k \neq j}^n 0.6 \bigg(\la \wb_{j,r}^{(0)}, \bxi_k \ra + \underline{\rho}_{j,r,k}^{(t-1)} +\sum_{i=1, i\neq k}^n \overline{\rho}_{j,r,i}^{(t-1)} \frac{\la \bxi_i, \bxi_k \ra}{\|\bxi_i\|_2^2}\bigg)\\
    &+ \frac{\eta}{n} \sum_{k=1, y_k = j}^n 0.4 \bigg(\la \wb_{j,r}^{(0)}, \bxi_k \ra + \overline{\rho}_{j,r,k}^{(t-1)} +\sum_{i=1, i\neq k}^n \underline{\rho}_{j,r,i}^{(t-1)} \frac{\la \bxi_i, \bxi_k \ra}{\|\bxi_i\|_2^2}\bigg)\\
    =& v_{j,r,2}^{(t-1)} - \frac{\eta}{n} \sum_{k=1, y_k \neq j}^n 0.6 \la \wb_{j,r}^{(0)}, \bxi_k \ra + \underline{\rho}_{j,r,k}^{(t)}\bigg(0.6 - \sum_{i=1, i \neq k}^n \frac{|\la \bxi_i, \bxi_k \ra|}{\|\bxi_i\|_2^2}\bigg)\\
    &+\frac{\eta}{n}
    \sum_{k=1, y_k = j}^n 0.4 \la \wb_{j,r}^{(0)}, \bxi_k \ra + \overline{\rho}_{j,r,k}^{(t)}\bigg(0.4  - \sum_{i=1, i \neq k}^n \frac{|\la \bxi_i, \bxi_k \ra|}{\|\bxi_i\|_2^2}\bigg)\\
    \ge& v_{j,r,2}^{(t-1)} - \frac{\eta}{n} \sum_{k=1, y_k \neq j}^n 1.2 \sqrt{\log(8mn/\delta)}\cdot \sigma_0 \sigma_p\sqrt{d} + \underline{\rho}_{j,r,k}^{(t)}(0.6 - \sum_{i=1}^n 4 \sigma_p^2 \cdot \sqrt{\log(4n^2/\delta)/d})\\
    &+ \frac{\eta}{n} \sum_{k=1, y_k = j}^n -0.8 \sqrt{\log(8mn/\delta)}\cdot \sigma_0 \sigma_p\sqrt{d} + \underline{\rho}_{j,r,k}^{(t)}(0.4 - \sum_{i=1}^n 4 \sigma_p^2 \cdot \sqrt{\log(4n^2/\delta)/d})\\
    \ge& v_{j,r,2}^{(t-1)} - \eta \sqrt{\log(8mn/\delta)}\cdot \sigma_0 \sigma_p\sqrt{d} - \frac{\eta}{n}\sum_{k=1, y_k \neq j}^n 0.3 \underline{\rho}_{j,r,k}^{(t)} + \frac{\eta}{n} \sum_{k=1, y_k = j}^n 0.2 \overline{\rho}_{j,r,k}^{(t)}\\
    \ge& v_{j,r,2}^{(t-1)} - \eta \sqrt{\log(8mn/\delta)}\cdot \sigma_0 \sigma_p\sqrt{d}\\
    \ge& v_0 - \sum_{s=1}^{t} \eta \sqrt{\log(8mn/\delta)}\cdot \sigma_0 \sigma_p\sqrt{d}\\
    \ge& v_0 - t \cdot \eta \sqrt{\log(8mn/\delta)}\cdot \sigma_0 \sigma_p\sqrt{d}\\
    \ge& 0.5 v_0.
\end{align*}
    The first inequality is from $\ell'^{(t)} \in [0.4, 0.6]$; The second inequality is from the monotonicity of $\rho$ so that $\overline{\rho} \ge 0$ and $\underline{\rho} \le 0$; The third inequality is from Lemma \ref{lemma: noise bound} and Lemma \ref{lemma: preliminary_initial inner product}; The fourth inequality is from Condition \ref{condition: main condition} the definition of d; The last inequality is from Condition \ref{condition: main condition} the definition of $d$ and the monotonicity of $\rho$; The fifth inequality is again from the monotonicity of $\rho$ and the last inequality is from the definition of $T_0$. Therefore, we complete the proof for the first case.
    
    Then we use induction to prove for the second case. The conclusion holds obviously at the end of case 1 when $t=T_0$. Assume that the conclusion holds for time $T_0 \le t \le \tilde t-1$ and we consider the case at time $t = \tilde
     d$. Recall the update equation of $\overline{\rho}_{j,r,i}^{(t)}$
     \begin{align*}
         \overline{\rho}^{(t+1)}_{j,r,i} = \overline{\rho}_{j,r,i}^{(t)} - \frac{\eta}{n} v_{j,r,2}^{(t)} \cdot \ell'^{(t)}_{i} \cdot \sigma'(\la \wb_{j,r}^{(t)}, \bxi_i \ra) \cdot \|\bxi_i\|_2^2 \cdot \mathbb{I}(y_i = j).
     \end{align*}
     So for $t = T_0$, we must have
     \begin{align}
     \label{eq: over_rho_lower}
         \overline{\rho}^{(t)}_{j,r,i} &= \overline{\rho}_{j,r,i}^{(0)} - \sum_{s=1}^t \frac{\eta}{n} v_{j,r,2}^{(s)} \cdot \ell'^{(s)}_{i} \cdot \sigma'(\la \wb_{j,r}^{(s)}, \bxi_i \ra) \cdot \|\bxi_i\|_2^2 \cdot \mathbb{I}(y_i = j)\nonumber\\
         &\ge \sum_{s=1}^t \frac{\eta}{n} 0.5 v_0 \cdot
          0.4 \cdot 1/2 \sigma_p^2 d \cdot 2\sqrt{\log(8mn/\delta)}\cdot \sigma_0 \sigma_p\sqrt{d}\nonumber\\
          &= 0.2 T_0 v_0 \sigma_p^2 d \sqrt{\log(8mn/\delta)}\cdot \sigma_0 \sigma_p\sqrt{d}\nonumber\\
          &\ge 20\sqrt{\log(8mn/\delta)}\cdot \sigma_0 \sigma_p\sqrt{d}.
     \end{align}
     The first inequality is from $\ell'^{(t)}_i \ge 0.4$, $v_{j,r,2}^{(t)} \ge 0.5v_0$ we just proved and Lemma \ref{lemma: noise bound}; The second inequality is from the definition of $T_0$ and Condition \ref{condition: main condition} so $T_0 = \Omega\Big(\frac{1}{v_0 \sigma_p^2 d}\Big)$.
     
    Therefore, we have
\begin{align*}
    v_{j,r,2}^{(\tilde t)} 
    \ge& v_{j,r,2}^{(\tilde t-1)} - \eta \sqrt{\log(8mn/\delta)}\cdot \sigma_0 \sigma_p\sqrt{d} - \frac{\eta}{n}\sum_{k=1, y_k \neq j}^n 0.3 \underline{\rho}_{j,r,k}^{(\tilde t)} + \frac{\eta}{n} \sum_{k=1, y_k = j}^n 0.2 \overline{\rho}_{j,r,k}^{(\tilde t)}\\
    \ge& v_{j,r,2}^{(\tilde t-1)} - \frac{\eta}{n}\sum_{k=1, y_k \neq j}^n 0.3 \underline{\rho}_{j,r,k}^{(\tilde t)}\\ 
    \ge& v_{j,r,2}^{(\tilde t-1)} . 
\end{align*}
The first inequality is derived as case 1; The second inequality  is from \ref{eq: over_rho_lower} and $\overline{\rho}$ is non-decreasing; The last inequality is from the monotonicity of $\underline{\rho}$ so $\underline{\rho} \le 0$. 
Therefore, we prove for the case at $\tilde t$.
\end{proof}

Then we could prove that during stage 1, $\la \wb_{j,r}^{(t)}, \bxi_k \ra$ is balanced for all $y_k = j$, which indicates that the neurons capture the features of samples with the same label evenly.
\begin{lemma}
\label{lemma: inner_product_balance}
If $|\ell'^{(t)}_i| \in [0.4, 0.6]$ for all $i \in [n]$ within time $[T_*, T^{*,1}]$, then for any $i,k$ that $y_i = y_k = j$,
\begin{align*}
    \frac{1}{9} \le \frac{\la \wb_{j,r}^{(t)}, \bxi_i \ra}{\la \wb_{j,r}^{(t)}, \bxi_k \ra} \le 18
\end{align*}
where $T_*=\Theta\Big(\frac{\sigma_0 n \sqrt{\log(8mn/\delta)}}{\eta v_0\sigma_p \sqrt{d}}\Big)$ and $T^{*,1}$ is the same as defined in (\ref{eq: define_T1}).
\end{lemma}

\begin{proof}[Proof of Lemma \ref{lemma: inner_product_balance}]
We use induction to prove this lemma. When $t=T_*$, it holds that
\begin{align*}
\la \wb_{j,r}^{(T_*)} , \bxi_i \ra &= \la \wb_{j,r}^{(0)}, \bxi_i \ra +\frac{\eta}{n} \sum_{t=1}^{T_*} v_{j,r,2}^{(t)} \|\bxi_i\|_2^2 \ell'^{(t)}_i + \frac{j\eta}{n} \sum_{t=1}^{T_*} v_{j,r,2}^{(t)} \sum_{i'=1,i'\neq i}^n y_{i'} {\ell'}_{i'}^{(t)} \la \bxi_i, \bxi_{i'} \ra \mathbb{I}(\la\wb_{j,r}^{(t)},\bxi_{i'}\ra >0).
\end{align*}
So we have
\begin{align*}
\bigg|\la \wb_{j,r}^{(T_*)} , \bxi_i \ra - \frac{\eta}{n} \sum_{t=1}^{T_*} v_{j,r,2}^{(t)} \|\bxi_i\|_2^2 \ell'^{(t)}_i\bigg| &= |\la \wb_{j,r}^{(0)}, \bxi_i \ra + \frac{j\eta}{n} \sum_{t=1}^{T_*} v_{j,r,2}^{(t)} \sum_{i'=1,i'\neq i}^n y_{i'} {\ell'}_{i'}^{(t)} \la \bxi_i, \bxi_{i'} \ra \mathbb{I}(\la\wb_{j,r}^{(t)},\bxi_{i'}\ra >0)|\\
&\le |\la \wb_{j,r}^{(0)}, \bxi_i \ra| + 2 \frac{\eta}{n} \sum_{t=1}^{T_*} \sigma_p^2 \cdot \sqrt{d \log(4n^2/\delta)} \sum_{i'=1}^n |{\ell'}_{i'}^{(0)}| \cdot  v_{j,r,2}^{(t)}\\
&\le |\la \wb_{j,r}^{(0)}, \bxi_i \ra| + 1.2 \eta \sum_{t=1}^{T_*} \cdot \sigma_p^2 \sqrt{d \log(4n^2/\delta)}\cdot  v_{j,r,2}^{(t)}\\
&\le |\la \wb_{j,r}^{(0)}, \bxi_i \ra| + \frac{\eta}{10n} \sum_{t=1}^{T_*} \sigma_p^2 d  v_{j,r,2}^{(t)}\\
&\le |\la \wb_{j,r}^{(0)}, \bxi_i \ra| + \frac{\eta}{4n} \sum_{t=1}^{T_*} v_{j,r,2}^{(t)} \|\bxi_i\|_2^2 \cdot |\ell'^{(t)}_i|,
\end{align*}
where the first inequity is due to lemma \ref{lemma: noise bound}, the second inequity is due to $|\ell'^{(t)}_i| \le 0.6$, the third inequity is due to $d = \Omega(n^2 \log(4n^2/\delta))$ and the last inequity is due to Lemma \ref{lemma: noise bound} that $\|\bxi_i\|_2^2 \ge 1/2 \sigma_p^2 d$ and $|\ell'^{(t)}_i| \ge 0.4$ with probability at least $1-\delta$. 

We then further bound the first term on the right side that:
\begin{align*}
    |\la \wb_{j,r}^{(0)}, \bxi_i \ra| &\le 2\sqrt{\log(8mn/\delta)}\cdot \sigma_0 \sigma_p\sqrt{d}\\
    &\le \frac{\eta T_*}{4n} 0.5 v_0 \cdot \frac{1}{2} \sigma_p^2 d \cdot0.4\\
    &\le \frac{\eta}{4n} \sum_{t=1}^{T_*} v_{j,r,2}^{(t)} \|\bxi_i\|_2^2 \cdot |\ell'^{(t)}_i|,
\end{align*}
where the first inequity is due to Lemma \ref{lemma: noise bound}, the second inequity is due to $T_*=\Theta(\frac{\sigma_0 n \sqrt{\log(8mn/\delta)}}{\eta v_0\sigma_p \sqrt{d}})$ and the last inequity is due to Lemma \ref{lemma: noise bound} that $\|\bxi_i\|_2^2 \ge 1/2 \sigma_p^2 d$ and $|\ell'^{(t)}_i| \ge 0.4$ with probability at least $1-\delta$. 

Therefore, we have
\begin{align*}
    &\bigg|\la \wb_{j,r}^{(T_*)} , \bxi_i \ra - \frac{\eta}{n} \sum_{t=1}^{T_*} v_{j,r,2}^{(t)} \|\bxi_i\|_2^2 \ell'^{(t)}_i\bigg| \le \frac{\eta}{2n} \sum_{t=1}^{T_*} v_{j,r,2}^{(t)} \|\bxi_i\|_2^2 \cdot \big|\ell'^{(t)}_i\big|,\\
    &\frac{\eta}{2n} \sum_{t=1}^{T_*} v_{j,r,2}^{(t)} \|\bxi_i\|_2^2 \cdot |\ell'^{(t)}_i| \le \la \wb_{j,r}^{(T_*)} , \bxi_i \ra \le \frac{3\eta}{2n} \sum_{t=1}^{T_*} v_{j,r,2}^{(t)} \|\bxi_i\|_2^2 \cdot |\ell'^{(t)}_i|.
\end{align*}

Then we could bound the noise memorization of two samples as
\begin{align*}
    \frac{\la \wb_{j,r}^{(T_*)} , \bxi_i \ra}{\la \wb_{j,r}^{(T_*)}, \bxi_k \ra} &\le \max_{t \le T_*}\Bigg\{\frac{\frac{3}{2} \|\bxi_i\|_2^2 \ell'^{(t)}_i}{\frac{1}{2}  \|\bxi_k\|_2^2 \ell'^{(t)}_{k}}\Bigg\}
    \le \frac{3 \cdot 0.6 \cdot 2 \sigma_p^2 d}{0.4 \cdot \frac{1}{2}\sigma_p^2 d}
    = 18,\\
    \frac{\la \wb_{j,r}^{(T_*)} , \bxi_i \ra}{\la \wb_{j,r}^{(T_*)}, \bxi_k \ra}&\ge \min_{t \le T_*} \Bigg\{\frac{\frac{1}{2}  \|\bxi_i\|_2^2 \ell'^{(t)}_i}{\frac{3}{2} \|\bxi_k\|_2^2 \ell'^{(t)}_{k}}\Bigg\} \le \frac{0.4 \cdot \sigma_p^2 d}{3 \cdot 0.6 \cdot 2 \sigma_p^2 d} = \frac{1}{9}.
\end{align*}

So the conclusion holds at time $T_*$. 

For $T_* < t \le T^{*,1}$, suppose that the conclusion holds at $\tilde t-1$. We prove for the case at time $\tilde t$. Recall the update equation of $\la \wb_{j,r}^{(t)}, \bxi_i \ra$:
\begin{align*}
    \la \wb_{j,r}^{(t)}, \bxi_i \ra =& \la \wb_{j,r}^{(t-1)}, \bxi_i \ra + \frac{\eta}{n} v_{j,r,2}^{(t-1)} \|\bxi_i\|_2^2 \ell'^{(t-1)}_{i} \\
    &+ \frac{j\eta}{n} v_{j,r,2}^{(t-1)} \sum_{i'=1,i'\neq i}^n y_{i'} {\ell'}_{i'}^{(t-1)} \la \bxi_i, \bxi_{i'} \ra \mathbb{I}(\la\wb_{j,r}^{(t-1)},\bxi_{i'}\ra >0).
\end{align*}
We could follow the proof above and get that:
\begin{align*}
    \frac{\eta}{2n} v_{j,r,2}^{(\tilde t)} \|\bxi_i\|_2^2 \cdot |\ell'^{(\tilde t)}_i| \le \la \wb_{j,r}^{(\tilde t)} , \bxi_i \ra - \la \wb_{j,r}^{(\tilde t-1)} , \bxi_i \ra \le \frac{3\eta}{2n} v_{j,r,2}^{(\tilde t)} \|\bxi_i\|_2^2 \cdot |\ell'^{(\tilde t)}_i|.
\end{align*}
So we have
\begin{align*}
    \frac{\la \wb_{j,r}^{(\tilde t)}, \bxi_i \ra}{\la \wb_{j,r}^{(\tilde t)}, \bxi_k \ra} &\le \max\Bigg\{\frac{\la \wb_{j,r}^{(\tilde t-1)} , \bxi_i \ra}{\la \wb_{j,r}^{(\tilde t-1)}, \bxi_k \ra}, \frac{\frac{3}{2} \|\bxi_i\|_2^2 \ell'^{(\tilde t-1)}_{i}}{\frac{1}{2}  \|\bxi_k\|_2^2 \ell'^{(\tilde t-1)}_{k}}\Bigg\}
    \le \frac{3 \cdot 0.6 \cdot 2 \sigma_p^2 d}{0.4 \cdot \frac{1}{2}\sigma_p^2 d}
    = 18,
\end{align*}
where the first inequity is due to $\frac{a+b}{c+d} \le \max\{\frac{a}{c}, \frac{b}{d} \}$ for any $a,b,c,d \ge 0$. The second inequity is due to the induction hypothesis. Similarly, we could lower bound as
\begin{align*}
    \frac{\la \wb_{j,r}^{(\tilde t)}, \bxi_i \ra}{\la \wb_{j,r}^{(\tilde t)}, \bxi_k \ra} &\ge \min\Bigg\{\frac{\la \wb_{j,r}^{(\tilde t-1)}, \bxi_i \ra}{\la \wb_{j,r}^{(\tilde t-1)}, \bxi_k \ra}, \frac{\frac{1}{2}  \|\bxi_i\|_2^2 \ell'^{(\tilde t-1)}_{i}}{\frac{3}{2} \|\bxi_k\|_2^2 \ell'^{(\tilde t-1)}_{k}}\Bigg\} \le \frac{0,4 \cdot \sigma_p^2 d}{3 \cdot 0.6 \cdot 2 \sigma_p^2 d} = \frac{1}{9},
\end{align*}
where the first inequity is due to $\frac{a+b}{c+d} \ge \min\{\frac{a}{c}, \frac{b}{d} \}$ for any $a,b,c,d \ge 0$ and the second inequity is due to the induction hypothesis. So the conclusion also holds at time $\tilde t$.
\end{proof}

Then we can provide the main proof:
\begin{proof}[\textbf{\underline{Proof of Proposition~\ref{prop: stage1_l}}}]

Note that the above two lemmas hold when $\ell'^{(t)}_i \in [0.4, 0.6]$ for any $i \in [n]$, we will prove that within time $T^{*,1}$, $\ell'^{(t)}_i \in [0.4, 0.6]$ for any $i \in [n]$ holds.

Recall that we define $T^{*,1} = \max\limits_{k,j,r}\{t:\max\limits_k \{v^{(t)}_{j,r,2}\cdot\sigma(\la\wb_{j,r}^{(t)},\bxi_k\ra)\} \le 0.1/m\}$ in (\ref{eq: define_T1}). We could bound the value of $|\ell'^{(t)}_i|$ within $T^{*,1}$ for all $i \in [n]$:
\begin{align*}
    |0.5 - |\ell'^{(t)}_i||&= \bigg|0.5 - \frac{1}{1+\exp\{y_i \cdot[F_{+1}(\bW_{+1}^{(t)},V_{+1},x_i)-F_{-1}(\bW_{-1}^{(t)},V_{-1},x_i)]\}}\bigg|\\
    &\le \bigg|\frac{\exp\{y_i \cdot[F_{+1}(\bW_{+1}^{(t)},V_{+1},x_i)-F_{-1}(\bW_{-1}^{(t)},V_{-1},x_i)]\}}{4}\bigg|,
\end{align*}
where the last inequity is due to $|\frac{1}{2} - \frac{1}{1+e^x}|\le |\frac{x}{4}|$ for any $x$. Then we have
\begin{align*}
    |\ell'^{(t)}_i| &\ge 0.5 - \frac{\exp\{y_i \cdot[F_{+1}(\bW_{+1}^{(t)},V_{+1},x_i)-F_{-1}(\bW_{-1}^{(t)},V_{-1},x_i)]\}}{4}\\
    &= 0.5 - \Theta(\exp\{y_i \cdot[F_{+1}(\bW_{+1}^{(t)},V_{+1},x_i)-F_{-1}(\bW_{-1}^{(t)},V_{-1},x_i)]\})\\
    &= 0.5 - \Theta\bigg(\max\limits_{j,r}\bigg\{\sum_{r=1}^m v^{(t)}_{j,r,2}\cdot\sigma(\la\wb_{j,r}^{(t)},\bxi_i\ra), \sum_{r=1}^m v^{(t)}_{j,r,1}\cdot\sigma(\la\wb_{j,r}^{(t)},\bmu\ra)\bigg\}\bigg)\\
    &= 0.5 - \Theta(\max\limits_{j,r}\{\sum_{r=1}^m v^{(t)}_{j,r,2}\cdot\sigma(\la\wb_{j,r}^{(t)},\bxi_i\ra))\\
    &= 0.5 - 0.1\\
    &= 0.4.
\end{align*}
 Similarly we could derive $|\ell'^{(t)}_i| \le 0.6$. So we have $|\ell'^{(t)}_i|\in [0.4, 0.6]$ for all $t \le T^{*,1}$.

Then we can prove the four statements. We prove the first two statements together. Here we use the technical lemma which depicts the dynamic of two intertwined sequence which has been proved in section \ref{sec: tech_lemmas}:

\lemmaseq*

Recall the update of $\la\wb_{j,r}^{(t)},\bxi_k\ra$ and $v_{j,r,2}^{(t)}$:
\begin{align*}
 \la\wb_{j,r}^{(t+1)},\bxi_k\ra &= \la\wb_{j,r}^{(t)},\bxi_k\ra - \frac{j y_k \eta}{n} v_{j,r,2}^{(t)} \|\bxi_k\|_2^2 \ell'^{(t)}_{k} -\frac{j \eta}{n} v_{j,r,2}^{(t)} \sum_{i=1,i\neq k}^n y_i \ell'^{(t)}_i \la \bxi_k,\bxi_i \ra \mathbb{I}(\la\wb_{j,r}^{(t)},\bxi_i\ra >0),\\ 
    v_{j,r,2}^{(t+1)} & = v_{j,r,2}^{(t)} - \frac{j \eta}{n}\sum_{i=1}^{n}\ell'^{(t)}_iy_i\la \wb_{j,r}^{(t)},\bxi_i \ra \mathbb{I}(\la\wb_{j,r}^{(t)},\bxi_i\ra >0).
\end{align*}

Take $a_0, b_0, A, B$ as $\la\wb_{j,r}^{(0)},\bxi_k\ra, v_0, 0.4 \frac{\eta}{n}  \|\bxi_k\|_2^2, \eta/180 $, from Condition \ref{condition: main condition} we know that these parameters satisfy the condition of Lemma \ref{lemma: interwined_sequence}. So we could estimate the time when the noise part becomes balanced:
\begin{align*}
&T_0 = \Theta(1 / \sqrt{AB}) = \frac{3\sqrt{20n}}{\eta \|\bxi_k\|_2} = \Theta\bigg(\frac{\sqrt{n}}{\eta \sigma_p \sqrt{d}}\bigg).
\end{align*}
Meanwhile, from lemma \ref{lemma: interwined_sequence} we also know that $v^{(T_0)}_{j,r,2} = \Theta(v_0)$,
 $\la\wb_{j,r}^{(T_0)},\bxi_k\ra = \Theta(\sqrt{A/B} \cdot v_0)$, and we could compute the output of noise part at time $T_0$:
\begin{align*}
&v^{(T_0)}_{j,r,2}\cdot\sigma(\la\wb_{j,r}^{(T_0)},\bxi_k\ra) = \Theta(\sqrt{A/B} \cdot b_0 ^2) = \Theta\Bigg(\frac{\|\bxi_k\|_2 \cdot v_0^2}{\sqrt{n}}\Bigg) = \Theta\Bigg(\frac{\sigma_p \sqrt{d} \cdot v_0^2}{\sqrt{n}}\Bigg) \ge 0.1/m,
\end{align*}
where the last inequity is from  the output layer initialization $v_0 = v_0  \ge \Theta(n^{1/4}\sigma_p^{-1/2}d^{-1/4}m^{-1/2})$

Therefore, when the noise memorization reaches constant level, two layers of noise part have not balanced yet, i.e. $\la\wb_{y_k,r}^{(T_1)},\bxi_k\ra / v^{(T_1)}_{y_k,r,2} = o(\sigma_p \sqrt{d} / \sqrt{n})$.

Then we prove the first conclusion. Recall the update of $\la\wb_{j,r}^{(t)},\bxi_k\ra$ when $j = y_k$:
\begin{align*}
 \la\wb_{j,r}^{(t+1)},\bxi_k\ra &= \la\wb_{j,r}^{(t)},\bxi_k\ra - \frac{j y_k \eta}{n} v_{j,r,2}^{(t)} \|\bxi_k\|_2^2 \ell'^{(t)}_{k} -\frac{j \eta}{n} v_{j,r,2}^{(t)} \sum_{i=1,i\neq k}^n y_i \ell'^{(t)}_i \la \bxi_k,\bxi_i \ra \mathbb{I}(\la\wb_{j,r}^{(t)},\bxi_i\ra >0)\\
 &\ge \la\wb_{j,r}^{(t)},\bxi_k\ra + \frac{0.1 \eta}{n} v_{j,r,2}^{(t)} \|\bxi_k\|_2^2 \\
 &= \la\wb_{j,r}^{(0)},\bxi_k\ra + \sum_{s=1}^t \frac{0.1 \eta}{n} v_{j,r,2}^{(s)} \|\bxi_k\|_2^2\\
 &\ge \la\wb_{j,r}^{(0)},\bxi_k\ra + \frac{0.1 \eta t}{n}  \|\bxi_k\|_2^2 \cdot 0.5 v_0\\
 &= \Theta\bigg(\frac{\eta t \sigma_p^2 d v_0}{n}\bigg).
\end{align*}
The first inequality is due to $\ell'^{(t)}_i \ge 0.4$ in Stage 1 and the definition of $d$ in Condition \ref{condition: main condition}; The second inequality is from Lemma \ref{lemma: positive_v} and the last equality is from Lemma \ref{lemma: preliminary_initial inner product}.
So $\la\wb_{j,r}^{(t)},\bxi_k\ra$ will grow linearly with $t$. When $\sum_{r=1}^m v^{(T_1)}_{y_k,r,2}\cdot\sigma(\la\wb_{y_k,r}^{(T_1)},\bxi_k\ra)= 0.05$, it suffices that $ \la\wb_{j,r}^{(t)},\bxi_k\ra$ reaches level of $\Theta\Big(\frac{1}{m v_0}\Big)$. Let $\Theta\Big(\frac{\eta t \sigma_p^2 d v_0}{n}\Big) = \Theta\Big(\frac{1}{m v_0}\Big)$ we could get that $t = \Theta\Big(\frac{n}{\eta \sigma_p^2 d m v_0^2}\Big)$. 

Note that if we take $v_0 = 1/m$, the model is the same as that of \cite{kou2023benign}. So we could derive that $t = \Theta(\frac{n m}{\eta \sigma_p^2 d})$, which corresponds to $T_1$ when $\overline{\rho} \ge 2$ in their Lemma D.1.

Next we prove for the last two statements. Recall the update of $\la\wb_{j,r}^{(t)},y_k \bmu\ra$ and $v_{j,r,1}^{(t)}$:
\begin{align*}
\la\wb_{j,r}^{(t+1)},y_k \bmu\ra &= \la\wb_{j,r}^{(t)},y_k \bmu\ra - j y_k \frac{\eta}{n} v_{j,r,1}^{(t)} \|\bmu\|_2^2 \sum_{i=1}^n\ell'^{(t)}_i \mathbb{I}(\la\wb_{j,r}^{(t)},y_i \cdot \bmu\ra >0),\\
v_{j,r,1}^{(t+1)} & = v_{j,r,1}^{(t)} - j \frac{\eta}{n}\sum_{i=1}^{n}\ell'^{(t)}_i \la \wb_{j,r}^{(t)},\bmu \ra \mathbb{I}(\la\wb_{j,r}^{(t)},y_i \cdot \bmu\ra >0).
\end{align*}

Here we use another technical lemma that compares the scale of two intertwined sequences in Section \ref{sec: tech_lemmas}:



\lemmacompare*

We use lemma \ref{lemma: compare_sequence} by taking $\overline{a_0}=\la\wb_{j,r}^{(0)},\bxi_k\ra, \overline{b_0} = v_0$, $\overline{A} = 0.4 \frac{\eta}{n}  \|\bxi_k\|_2^2, \overline{B} = \eta/180$ and $\underline{a_0}=\la\wb_{j,r}^{(0)},\bmu\ra, \underline{b_0} = v_0$, $\underline{A} = 0.3 \eta  \|\bmu\|_2^2$, $\underline{B} = 0.3 \eta$. From the Condition \ref{condition: main condition}, these parameters satisfy the condition of lemma \ref{lemma: compare_sequence}. So we could get that there exists a constant c that $v_{j,r,1}^{(t)} \le \underline{b_t} = O(\overline{b_t}) \le O(v_{j,r,2}^{(t)}) = c \cdot v_0$. 

Meanwhile, It is obvious that $\la\wb_{j,r}^{(t)},y_k \bmu\ra \le \underline{a_t}$, $v_{j,r,1}^{(t)} \le \underline{b_t}$ and $\la\wb_{j,r}^{(t)},\bxi_k\ra \ge \overline{a_t}, v_{j,r,2}^{(t)} \ge \overline{b_t}$. So we have:
\begin{align*}
\la\wb_{j,r}^{(t)},y_k \bmu\ra &\le \underline{a_t} \le \underline{a_0} + t\cdot \underline{A} \cdot c \cdot v_0,\\
\la\wb_{j,r}^{(t)},y_k \bmu\ra &\ge \underline{a_0} + t/2 \cdot \underline{A} \cdot v_0,
\end{align*}
which implies that $\la\wb_{j,r}^{(t)},y_k \bmu\ra = \Theta(t \eta b_0 \|\bmu\|_2^2)$ and $v_{j,r,1}^{(t_1)} = \Theta(b_0)$. So by the time $T_0$ the balancing factor:
\begin{align*}
\la\wb_{j,r}^{(T_0)},y_k \bmu\ra / v_{j,r,1}^{(T_0)} = \eta T_0 \|\bmu\|_2^2 = \Theta\bigg(\frac{\sqrt{n}\|\bmu\|_2^2}{\sigma_p \sqrt{d}}\bigg) = o(\|\bmu\|_2).
\end{align*}
This suggests that the signal part is not balancd. Moreover, we could estimate the output of signal at this time to prove the third statement.
\begin{align*}
    \sum_{r=1}^m v_{j,r,1}^{(t)} \la\wb_{j,r}^{(t)},y_k \bmu\ra = \Theta(t \eta m v_0^2 \|\bmu\|_2^2) = \Theta\Bigg(\frac{n \|\bmu\|_2^2 }{\sigma_p^2 d}\Bigg).
\end{align*}
 Again the result corresponds to the Signal Learning $O(\hat{\gamma}) = O(n \cdot SNR^2) = O(\frac{n \|\bmu\|_2^2}{\sigma_p^2 d})$ at time $T_1$ in \cite{kou2023benign}.

Last we need to check that $T_1 \le T^{*,1}$. This is obvious because from the first statement we have $\sum_{r=1}^m v^{(T_1)}_{y_k,r,2}\cdot\sigma(\la\wb_{y_k,r}^{(T_1)},\bxi_k\ra)= 0.05 \le 0.1$. So $T_1$ must be within the bound of $T^{*,1}$.

\end{proof}

\subsection{Second stage}
In this section, we give the analysis from $T^{*,1}$ to the time when the training loss converges to 0. The main difference between the first stage and the second stage is that the loss derivative is not constant in the second stage.

From previous analysis, when $v_0 = v_0 > \Theta(n^{1/4}\sigma_p^{-1/2}d^{-1/4}m^{-1/2})$, there is only linear phase before the noise output reaches constant level. We first give the main proposition to estimate the signal and noise output. Here $T^* = \eta^{-1} poly(d, n, m,\varepsilon^{-1})$ is the maximum admissible iterations.

\begin{proposition}
\label{prop: stage2_analysis_c1}
    For any $T_1 \le t \le T^*$, it holds that,
    \begin{align}
        &0 \le v_{j,r,1}^{(t)} \le C_4 v_0 \alpha, \label{eq: v1_bound_c1}\\
        &0 \le v_{j,r,2}^{(t)} \le C_5 m v_0^2 \alpha, \label{eq: v2_bound_c1}\\
        &0 \le \gamma_{j,r}^{(t)} \le \frac{C_6 n \|\bmu\|_2^2 \alpha}{m v_0 \sigma_p^2 d},\label{eq: signal_output_bound_c1}\\
        &0 \le \overline{\rho}_{j,r,k}^{(t)} \le \alpha,\label{eq: pos_noise_bound_c1}\\
        &0 \ge \underline{\rho}_{j,r,k}^{(t)} \ge -2\sqrt{\log(8mn/\delta)}\cdot \sigma_0 \sigma_p\sqrt{d} - 8 \sqrt{\frac{\log(4n^2/\delta)}{d}} n\alpha \ge -\alpha \label{eq: neg_noise_bound_c1}
    \end{align}
    for all $j \in \{\pm 1\}, r \in [m]$ and $k \in [n]$. Here $\alpha =\frac{4\log(T^*)}{k_1 m v_0}$ where $k_1 = \min\limits_{k,j,r, t \le T_1} \Big\{\frac{v_{j,r,2}^{(t)} \sigma_p \sqrt{d}}{\sqrt{n}\la \wb_{j,r}^{(t)}, \bxi_k \ra}\Big\} = \Theta(1)$ and $C_4, C_5, C_6=\Theta(1)$.
\end{proposition}
We use induction to prove Proposition \ref{prop: stage2_analysis_c1}. We introduce several technical lemmas.

\begin{lemma}
\label{lemma: inner_rho_diff_c1}
    Under Condition \ref{condition: main condition}, suppose (\ref{eq: v1_bound_c1}) to (\ref{eq: neg_noise_bound_c1}) hold at iteration t. Then for all $k \in [n]$ it holds that
    \begin{align}
        &\Big| \la \wb_{-y_k, r}^{(t)} - \wb_{-y_k, r}^{(0)}, \bxi_k \ra - \underline{\rho}_{-y_k, r, k}^{(t)}\Big| \le 4n \alpha \sigma_p^2 \cdot \sqrt{\log(4n^2/\delta)/d}\label{eq: under_rho_diff_c1}\\
        &\Big| \la \wb_{y_k, r}^{(t)} - \wb_{y_k, r}^{(0)}, \bxi_k \ra - \overline{\rho}_{y_k, r, k}^{(t)}\Big| \le 4n \alpha \sigma_p^2 \cdot \sqrt{\log(4n^2/\delta)/d} \label{eq: over_rho_diff_c1}
    \end{align}
\end{lemma}

\begin{proof}[Proof of Lemma \ref{lemma: inner_rho_diff_c1}]
    First from signal-noise decomposition analysis, for $y_k \neq j$ we have
    \begin{align*}
        \la \wb_{j,r}^{(t)} - \wb_{j,r}^{(0)}, \bxi_k \ra &= \underline{\rho}_{j,r,k}^{(t)} +\sum_{i=1, i\neq k}^n \rho_{j,r,i}^{(t)} \frac{\la \bxi_i, \bxi_k \ra}{\|\bxi_i\|_2^2}.
    \end{align*}
    So we have
    \begin{align*}
        \Big|\la \wb_{j,r}^{(t)} - \wb_{j,r}^{(0)}, \bxi_k \ra - \underline{\rho}_{j,r,k}^{(t)}\Big| &\le \sum_{i=1, i\neq k}^n \rho_{j,r,i}^{(t)} \frac{|\la \bxi_i, \bxi_k \ra|}{\|\bxi_i\|_2^2}\\
        &\le \sum_{i=1, i\neq k}^n \overline{\rho}_{j,r,i}^{(t)} \frac{|\la \bxi_i, \bxi_k \ra|}{\|\bxi_i\|_2^2}\\
        &\le 4n \alpha \sigma_p^2 \cdot \sqrt{\log(4n^2/\delta)/d}
    \end{align*}
    The first inequality is from triangle inequality; The second inequality is from the monotonicity of $\rho$ and the last inequality is from our induction hypothesis and Lemma \ref{lemma: noise bound}.

    The proof for (\ref{eq: over_rho_diff_c1}) is the same as (\ref{eq: under_rho_diff_c1}) so we omit it for convenience.
\end{proof}

\begin{lemma}
\label{lemma: output_leading_term_c1}
    Under Condition \ref{condition: main condition}, suppose (\ref{eq: v1_bound_c1}) to (\ref{eq: neg_noise_bound_c1}) hold at iteration t. Then for all $k \in [n]$ it holds that
    \begin{align*}
        &F_{-y_k}(\Wb_{-y_k}^{(t)}, \bV_{-y_k}^{(t)}; \bx_k) \le 0.25,\\
        &-0.25 + \sum_{r=1}^m v_{y_k, r, 2}^{(t)} \sigma(\la \wb_{y_k, r}^{(t)}, \bxi_k \ra) \le F_{y_k}(\Wb_{y_k}^{(t)}, \bV_{y_k}^{(t)}; \bx_k) \le 0.25 + \sum_{r=1}^m v_{y_k, r, 2}^{(t)} \sigma(\la \wb_{y_k, r}^{(t)}, \bxi_k \ra).
    \end{align*}

    Further, as $y_k f(\Wb^{(t)}, \bV^{(t)}; \bx_k) = F_{y_k}(\Wb_{y_k}^{(t)}, \bV_{y_k}^{(t)}; \bx_k) - F_{-y_k}(\Wb_{-y_k}^{(t)}, \bV_{-y_k}^{(t)}; \bx_k)$, we have
    \begin{equation}
    \label{eq: output_leading_term_c1}
        -0.5 + \sum_{r=1}^m v_{y_k, r, 2}^{(t)} \sigma(\la \wb_{y_k, r}^{(t)}, \bxi_k \ra) \le y_k \cdot f(\Wb^{(t)}, \bV^{(t)}; \bx_k) \le 0.5 + \sum_{r=1}^m v_{y_k, r, 2}^{(t)} \sigma(\la \wb_{y_k, r}^{(t)}, \bxi_k \ra).
    \end{equation}
\end{lemma}

\begin{proof}[Proof of Lemma \ref{lemma: output_leading_term_c1}]
    Without loss of generality, we may assume $y_k = +1$. We have
    \begin{align*}
        F_{-y_k}(\bW_{-y_k}^{(t)}, \bV_{-y_k}^{(t)}; \bx_k) =& \sum_{r=1}^{m} v_{-y_k, r, 1}^{(t)}\sigma(\wb_{-y_k, r}^{(t)}, y_k \bmu \ra) + v_{-y_k, r, 2}^{(t)}\sigma(\la \wb_{-y_k, r}^{(t)}, \bxi_k \ra)\\
        =& \sum_{r=1}^m v_{-y_k,r,1}^{(t)} \cdot (\gamma_{-y_k,r}^{(t)} + \la \wb_{-y_k, r}^{(0)}, y_k \bmu \ra)\\
        &+ v_{-y_k, r, 2}^{(t)} \cdot (\la \wb_{-y_k, r}^{(0)}, \bxi_k \ra + \underline{\rho}_{-y_k, r, k}^{(t)} + 4n \alpha \sigma_p^2 \cdot \sqrt{\log(4n^2/\delta)/d}).
    \end{align*}
    The last equality is from Lemma \ref{lemma: inner_rho_diff_c1}. Therefore, we have
    \begin{align*}
         F_{-y_k}(\bW_{-y_k}^{(t)}, \bV_{-y_k}^{(t)}; \bx_k)=& \sum_{r=1}^m [v_{-y_k,r,1}^{(t)} \sigma(\la \wb_{-y_k, r}^{(t)}, \bxi_k \ra) + v_{-y_k,r,2}^{(t)} \sigma(\la \wb_{-y_k, r}^{(t)}, \bxi_k \ra)]\\
         \le& 4m \cdot \max\{C_4 v_{-y_k, r, 1}^{(0)} \alpha, C_5 m v_{-y_k, r, 2}^{(0)2} \alpha\} \cdot \max\bigg\{ |\la \wb_{-y_k, r}^{(0)}, y_k \bmu \ra|, |\la \wb_{-y_k, r}^{(0)}, \bxi_k \ra|,\\
         &\gamma_{-y_k,r}^{(t)},  4n \alpha \sigma_p^2 \cdot \sqrt{\log(4n^2/\delta)/d}\bigg\}\\
         \le& 4m \cdot \max\{C_4 v_{-y_k, r, 1}^{(0)} \alpha, C_5 m v_{-y_k, r, 2}^{(0)2} \alpha\} \cdot \max\bigg\{ \sqrt{2\log(8m/\delta)} \cdot \sigma_0\|\bmu\|_2,\\
         & 2\sqrt{\log(8mn/\delta)}\cdot \sigma_0 \sigma_p\sqrt{d}, \frac{C_6 n \|\bmu\|_2^2 \alpha}{m v_0 \sigma_p^2 d}, 4n \alpha \sigma_p^2 \cdot \sqrt{\frac{\log(4n^2/\delta)}{d}}\bigg\}\\
         \le& 0.25.
    \end{align*}
    The second inequality is from our induction hypothesis and the last inequality is from Condition \ref{condition: main condition}.

    For the analysis of $F_{y_k}(\wb_{y_k}^{(t)}, \bV_{y_k}^{(t)}; \bx_k)$, we first have
    \begin{align*}
        F_{y_k}(\wb_{y_k}^{(t)}, \bV_{y_k}^{(t)}; \bx_k) &= \sum_{r=1}^{m} v_{y_k, r, 1}^{(t)}\sigma(\wb_{y_k, r}^{(t)}, y_k \bmu \ra) + v_{y_k, r, 2}^{(t)}\sigma(\la \wb_{y_k, r}^{(t)}, \bxi_k \ra).
    \end{align*}
    And for $v_{y_k, r, 2}^{(t)}\sigma(\la \wb_{y_k, r}^{(t)}, \bxi_k \ra)$ we have
    \begin{align*}
        \bigg|\sum_{r=1}^{m} v_{y_k, r, 1}^{(t)}\sigma(\wb_{y_k, r}^{(t)}, y_k \bmu \ra)\bigg| &\le 2m \cdot C_4 v_0 \alpha \cdot \max\bigg\{|\la \wb_{y_k, r}^{(0)}, y_k \bmu \ra|, \gamma_{y_k, r}^{(t)}\bigg\}\\
        &\le 2m \cdot C_4 v_0 \cdot \max\bigg\{\sqrt{2\log(8m/\delta)} \cdot \sigma_0\|\bmu\|_2, \frac{C_6 n \|\bmu\|_2^2 \alpha}{m v_0 \sigma_p^2 d}\bigg\}\\
        &\le 0.25.
    \end{align*}
    The last inequality is also from Condition \ref{condition: main condition}. So we could get
    \begin{align*}
        -0.25 + \sum_{r=1}^m v_{y_k, r, 2}^{(t)} \sigma(\la \wb_{y_k, r}^{(t)}, \bxi_k \ra) \le F_{y_k}(\wb_{y_k}^{(t)}, \bV_{y_k}^{(t)}; \bx_k) \le 0.25 + \sum_{r=1}^m v_{y_k, r, 2}^{(t)} \sigma(\la \wb_{y_k, r}^{(t)}, \bxi_k \ra).
    \end{align*}
    This completes the proof.
\end{proof}

\begin{lemma}
\label{lemma: key_lemma_stage2_c1}
    Under Condition \ref{condition: main condition}, suppose (\ref{eq: signal_output_bound_c1}) to (\ref{eq: neg_noise_bound_c1}) hold at iteration t. Then it holds that:
    \begin{enumerate}
        \item $\sum_{r=1}^m v_{y_k, r, 2}^{(t)} \sigma(\la \wb_{y_k, r}^{(t)}, \bxi_k \ra) - v_{y_i, r, 2}^{(t)} \sigma(\la \wb_{y_i, r}^{(t)}, \bxi_i \ra) \le \kappa$ and $\ell'^{(t)}_i / \ell'^{(t)}_k \le C_7$ for all $i, k \in [n]$ and $y_i = y_k$.
        \item $v_{y_k, r, 2}^{(t)} \ge C_8 v_0$ for all $r \in [m], k \in [n]$.
        \item Define $S_i^{(t)} := \{r \in [m]: \la w_{y_i,r},\bxi_k \ra >0\}$, and $S_{j,r}^{(t)}:=\{k\in [n]:y_k=j,\la w_{j,r}^{(t)}, \bxi_k \ra > 0\}$. For all $k \in [n], r \in [m]$ and $j \in \{\pm 1\}$, $S_k^{(0)} \subseteq S_k^{(t)}, S_{j,r}^{(0)} \subseteq S_{j,r}^{(t)}$.
        \item $\Theta \Big(\frac{1}{\|\bmu\|_2}\Big) \le \frac{v_{y_k, r, 1}^{(t)}}{\la \wb_{y_k, r}^{(t)}, y_k\bmu \ra} \le \Theta\Big(\frac{m v_0^2 \sigma_p^2 d}{n \|\bmu\|_2^2}\Big)$ for all $k \in [n], r \in [m]$.
        \item  $\Theta \Big(\frac{\sqrt{n}}{\sigma_p \sqrt{d}}\Big) \le \frac{v_{y_k,r,2}^{(t)}}{\la \wb_{y_k,r}^{(t)}, \bxi_k \ra} \le \Theta (mv_0^2)$ for all $k \in [n], r \in [m]$.
    \end{enumerate}
    Here, $\kappa, C_7, C_8$ can be taken as $1, \exp(1.5)$ and $0.5$.
\end{lemma}

\begin{proof}[Proof of Lemma \ref{lemma: key_lemma_stage2_c1}]
    We use induction to prove this lemma. We first prove that all conclusions hold when $t = T_1$. From Proposition \ref{prop: stage1_l} we have
    \begin{align*}
        v^{(T_1)}_{y_k,r,2}\cdot\sigma(\la\wb_{y_k,r}^{(T_1)},\bxi_k\ra) \le 0.1/m.
    \end{align*}
    Therefore, 
    \begin{align*}
        \sum_{r=1}^m v_{y_k, r, 2}^{(t)} \sigma(\la \wb_{y_k, r}^{(t)}, \bxi_k \ra) - v_{y_i, r, 2}^{(t)} \sigma(\la \wb_{y_i, r}^{(t)}, \bxi_i \ra) &\le \sum_{r=1}^m |v_{y_k, r, 2}^{(t)} \sigma(\la \wb_{y_k, r}^{(t)}, \bxi_k \ra)| + \sum_{r=1}^m |v_{y_i, r, 2}^{(t)} \sigma(\la \wb_{y_i, r}^{(t)}, \bxi_i \ra)|\\
        &\le 0.1+0.1\\
        &= 0.2\\
        &\le \kappa.
    \end{align*}
    Meanwhile, as $\ell'^{(t)}_k \in [0.4, 0.6]$ for all $k \in [n]$, we have $\ell'^{(t)}_i / \ell'^{(t)}_k \le 1.5 \le C_1$. So the first conclusion holds.

    And from Lemma \ref{lemma: active neuron}, we could directly derive the second and fourth conclusions.

    Now, suppose that there exists $\tilde t \le T^*$ such that five conditions hold for any $0 \le t \le \tilde t-1$, we prove that these conditions also hold for $t = \tilde t$.

    We first prove conclusion 1. From (\ref{eq: output_leading_term_c1}) we have
    \begin{align}
        \bigg|y_k \cdot f(\wb^{(t)}, \bV^{(t)}; \bx_k) - y_i \cdot f(\wb^{(t)}, \bV^{(t)}; \bx_i) - \sum_{r=1}^m \Big[v_{y_k, r, 2}^{(t)} \sigma(\la \wb_{y_k, r}^{(t)}, \bxi_k \ra) - v_{y_i, r, 2}^{(t)} \sigma(\la \wb_{y_i, r}^{(t)}, \bxi_i \ra)\Big]\bigg| \le 0.5.\label{eq: output_diff_c1}
    \end{align}

    Recall the update equation of $v_{y_k, r, 2}^{(t)} \sigma(\la \wb_{y_k, r}^{(t)}, \bxi_k \ra)$:
    \begin{align*}
        v_{y_k, r, 2}^{(\tilde t)} \sigma(\la \wb_{y_k, r}^{(\tilde t)}, \bxi_k \ra) =& v_{y_k, r, 2}^{(\tilde t-1)} \sigma(\la \wb_{y_k, r}^{(\tilde t-1)}, \bxi_k \ra) - \frac{\eta}{n} v_{y_k, r, 2}^{(\tilde t-1)2} \|\bxi_k\|_2^2 \ell'^{(\tilde t-1)}_k\\
        &- \frac{y_k \eta}{n} \la \wb_{y_k, r}^{(\tilde t-1)}, \bxi_k \ra \sum_{i=1}^n \ell'^{(\tilde t-1)}_i y_i \la \wb_{y_k, r}^{(\tilde t-1)}, \bxi_i \ra \mathbb{I}(\la \wb_{y_k, r}^{(\tilde t-1)}, \bxi_i \ra > 0)\\
        &+ \frac{\eta^2}{n^2} v_{y_k, r, 2}^{(\tilde t-1)} \sum_{i=1, i\neq k}^n y_i \ell'^{(\tilde t-1)}_i \la \bxi_k, \bxi_i \ra \mathbb{I}(\la \wb_{y_k, r}^{(\tilde t-1)}, \bxi_i \ra \ge 0) \sum_{i=1}^n \ell'^{(\tilde t-1)}_i y_i \la \wb_{y_k, r}^{(\tilde t-1)}, \bxi_i \ra \mathbb{I}(\la \wb_{y_k, r}^{(\tilde t-1)}, \bxi_i \ra \ge 0).
    \end{align*}
    From the last induction hypothesis, we have $\Theta \Big(\frac{\sqrt{n}}{\sigma_p \sqrt{d}}\Big) \le \frac{v_{y_k,r,2}^{(t)}}{\la \wb_{y_k,r}^{(t)}, \bxi_k \ra} \le \Theta (mv_0^2)$ for all $t \le \tilde t -1$, and we could assume that $\frac{k_1 \sqrt{n}}{\sigma_p \sqrt{d}} \le  \frac{v^{(t)}_{j,r,2}}{\la\wb_{j,r}^{(t)},\bxi_k\ra} \le k_2 m v_0^2$ for all $j = y_k$, $k \in [n]$ and $t \le \tilde t -1$, thus,
    \begin{align*}
        v_{y_k, r, 2}^{(\tilde t)} \sigma(\la \wb_{y_k, r}^{(\tilde t)}, \bxi_k \ra) \le& v_{y_k, r, 2}^{(\tilde t-1)} \sigma(\la \wb_{y_k, r}^{(\tilde t-1)}, \bxi_k \ra) - \frac{\eta}{n} v_{y_k, r, 2}^{(\tilde t-1)2} \|\bxi_k\|_2^2 \ell'^{(\tilde t-1)}_k\\
        &- \frac{\eta}{n} \sum_{i=1}^n \ell'^{(\tilde t-1)}_i \cdot \frac{\sigma_p^2 d v_{y_k,r,2}^{(\tilde t-1)2}}{k_1^2 n} \cdot \mathbb{I}(\la \wb_{y_k, r}^{(\tilde t-1)}, \bxi_i \ra > 0)\\
        \le& v_{y_k, r, 2}^{(\tilde t-1)} \sigma(\la \wb_{y_k, r}^{(\tilde t-1)}, \bxi_k \ra) - \frac{\eta}{n} v_{y_k, r, 2}^{(\tilde t-1)2} \ell'^{(\tilde t-1)}_k \bigg(\|\bxi_k\|_2^2 + (C_1/k_1^2) \sigma_p^2 d \cdot (|S_{y_k, r}^{(\tilde t-1)}|/n)\bigg),
    \end{align*}
    where the last inequality is from the first induction hypothesis.

    We then lower bound $v_{y_i, r, 2}^{(\tilde t)} \sigma(\la \wb_{y_i, r}^{(\tilde t)}, \bxi_i \ra)$ that
    \begin{align*}
        v_{y_i, r, 2}^{(\tilde t)} \sigma(\la \wb_{y_i, r}^{(\tilde t)}, \bxi_i \ra) \ge& v_{y_i, r, 2}^{(\tilde t-1)} \sigma(\la \wb_{y_i, r}^{(\tilde t-1)}, \bxi_i \ra) - \frac{\eta}{n} v_{y_i, r, 2}^{(\tilde t-1)2} \|\bxi_i\|_2^2 \ell'^{(\tilde t-1)}_i\\
        &- \frac{\eta}{n} \sum_{i'=1}^n \ell'^{(\tilde t-1)}_{i'} \cdot \frac{v_{y_i,r,2}^{(\tilde t-1)2}}{k_2^2 m^2 v_{y_i, r, 2}^{(0)4}} \cdot \mathbb{I}(\la \wb_{y_i, r}^{(\tilde t-1)}, \bxi_{i'} \ra > 0)\\
        \ge& v_{y_i, r, 2}^{(\tilde t-1)} \sigma(\la \wb_{y_i, r}^{(\tilde t-1)}, \bxi_i \ra) - \frac{\eta}{n} v_{y_i, r, 2}^{(\tilde t-1)2} \ell'^{(\tilde t-1)}_i \bigg(\|\bxi_i\|_2^2 + (C_1 k_2^2 m^2 v_{y_i, r, 2}^{(0)4})^{-1} \cdot (|S_{y_i, r}^{(\tilde t-1)}|/n)\bigg).
    \end{align*}
The last inequality is from the first induction hypothesis.
Combining the above two inequalities, we have
    \begin{align*}
        &\sum_{r=1}^m v_{y_k, r, 2}^{(\tilde t)} \sigma(\la \wb_{y_k, r}^{(\tilde t)}, \bxi_k \ra) - v_{y_i, r, 2}^{(\tilde t)} \sigma(\la \wb_{y_i, r}^{(\tilde t)}, \bxi_i \ra) \\
        \le& \sum_{r=1}^m v_{y_k, r, 2}^{(\tilde t-1)} \sigma(\la \wb_{y_k, r}^{(\tilde t-1)}, \bxi_k \ra) - v_{y_i, r, 2}^{(\tilde t-1)} \sigma(\la \wb_{y_i, r}^{(\tilde t-1)}, \bxi_i \ra)\\
        -& \bigg[\frac{\eta}{n} v_{y_k, r, 2}^{(\tilde t-1)2} \ell'^{(\tilde t-1)}_k \bigg(\|\bxi_k\|_2^2 + (C_1/k_1^2) \sigma_p^2 d \cdot (|S_{y_k, r}^{(\tilde t-1)}|/n)\bigg) - \frac{\eta}{n} v_{y_i, r, 2}^{(\tilde t-1)2} \ell'^{(\tilde t-1)}_i \bigg(\|\bxi_i\|_2^2 + (C_1 k_2^2 m^2 v_{y_i, r, 2}^{(0)4})^{-1} \cdot (|S_{y_i, r}^{(\tilde t-1)}|/n)\bigg)\bigg].
    \end{align*}
    We consider two cases: $\sum_{r=1}^m v_{y_k, r, 2}^{(\tilde t-1)} \sigma(\la \wb_{y_k, r}^{(\tilde t-1)}, \bxi_k \ra) - v_{y_i, r, 2}^{(\tilde t-1)} \sigma(\la \wb_{y_i, r}^{(\tilde t-1)}, \bxi_i \ra) \le 0.9\kappa$ and $\sum_{r=1}^m v_{y_k, r, 2}^{(\tilde t-1)} \sigma(\la \wb_{y_k, r}^{(\tilde t-1)}, \bxi_k \ra) - v_{y_i, r, 2}^{(\tilde t-1)} \sigma(\la \wb_{y_i, r}^{(\tilde t-1)}, \bxi_i \ra) > 0,9\kappa$. When $\sum_{r=1}^m v_{y_k, r, 2}^{(\tilde t-1)} \sigma(\la \wb_{y_k, r}^{(\tilde t-1)}, \bxi_k \ra) - v_{y_i, r, 2}^{(\tilde t-1)} \sigma(\la \wb_{y_i, r}^{(\tilde t-1)}, \bxi_i \ra) \le 0.9\kappa$, we can first upper bound $v_{y_k, r, 2}^{(\tilde t-1)}$ as
    \begin{align*}
        v_{y_k, r, 2}^{(\tilde t-1)} &\le k_2 m v_0^2 \cdot \la \wb_{y_k, r}^{(\tilde t-1)}, \bxi_k \ra\\
        &\le k_2 m v_0^2 \cdot \bigg[\la\wb_{j,r}^{(0)},\bxi_k\ra + \rho_{j,r,k}^{(\tilde t-1)} + 4 \sqrt{\frac{\log(4n^2/\delta)}{d}} n\alpha \bigg]\\
        &\le 2 \alpha k_2 m v_0^2.
    \end{align*}
    The first inequality is from our last induction hypothesis; The second inequality is from Lemma \ref{lemma: inner_rho_diff_c1} and the last inequality is from (\ref{eq: pos_noise_bound_c1}).
    So we have
    \begin{align*}
        \sum_{r=1}^m v_{y_k, r, 2}^{(\tilde t)} \sigma(\la \wb_{y_k, r}^{(\tilde t)}, \bxi_k \ra) - v_{y_i, r, 2}^{(\tilde t)} \sigma(\la \wb_{y_i, r}^{(\tilde t)}, \bxi_i \ra) &\le 0.9\kappa - \frac{\eta}{n} \sum_{r=1}^m v_{y_k, r, 2}^{(\tilde t-1)2} \ell'^{(\tilde t-1)}_k \bigg(\|\bxi_k\|_2^2 + (C_7/k_1^2) \sigma_p^2 d \cdot (|S_{y_k, r}^{(\tilde t-1)}|/n)\bigg)\\
        &\le 0.9\kappa + \frac{\eta}{n} \sum_{r=1}^m v_{y_k, r, 2}^{(\tilde t-1)2} (\|\bxi_k\|_2^2 + (C_7/k_1^2) \sigma_p^2 d)\\
        &\le 0.9 \kappa + \frac{\eta m}{n} \Big(2 \alpha k_2 m v_0^2\Big)^2 (3/2 + C_7/k_1^2) \sigma_p^2 d\\
        &\le 0.9 \kappa + 0.1 \kappa\\
        &\le \kappa.
    \end{align*}
    The first inequality is due to $\ell'^{(\tilde t-1)}_i \le 0$; The second inequality is due to $|S_{y_k, r}^{(\tilde t -1)}| \le n$ and the last inequality is from Condition \ref{condition: main condition} that $\eta \le C^{-1} nm^{-3} \sigma_p^{-2} d^{-1} v_{j,r,2}^{(0)-4}$.

    On the other hand, when  $v_{y_k, r, 2}^{(\tilde t-1)} \sigma(\la \wb_{y_k, r}^{(\tilde t-1)}, \bxi_k \ra) - v_{y_i, r, 2}^{(\tilde t-1)} \sigma(\la \wb_{y_i, r}^{(\tilde t-1)}, \bxi_i \ra) > 0.9\kappa$, from (\ref{eq: output_diff_c1}) we have
    \begin{align*}
        y_k \cdot f(\wb^{(\tilde t-1)}, \bV^{(\tilde t-1)}; \bx_k) - y_i \cdot f(\wb^{(\tilde t-1)}, \bV^{(\tilde t-1)}; \bx_i) &\ge \sum_{r=1}^m \Big[v_{y_k, r, 2}^{(\tilde t-1)} \sigma(\la \wb_{y_k, r}^{(\tilde t-1)}, \bxi_k \ra) - v_{y_i, r, 2}^{(\tilde t-1)} \sigma(\la \wb_{y_i, r}^{(\tilde t-1)}, \bxi_i \ra)\Big] - 0.5\\
        &\ge 0.9 \kappa - 0.5 \kappa\\
        &\ge 0.4 \kappa.
    \end{align*}
    The second inequality is from $\kappa = 1$. So we have
    \begin{align*}
        \frac{\ell'^{(\tilde t-1)}_i}{\ell'^{(\tilde t-1)}_k} \le \exp(y_k \cdot f(\wb^{(\tilde t-1)}, \bV^{(\tilde t-1)}; \bx_k) - y_i \cdot f(\wb^{(\tilde t-1)}, \bV^{(\tilde t-1)}; \bx_i)) \le \exp(-0.4 \kappa).
    \end{align*}
    The first inequality is due to $\frac{1 + \exp(b)}{1 + \exp(a)} \le \exp(b-a)$ for $b \ge a \ge 0$.
    
    Therefore, for all $r \in [m]$ we have
    \begin{align*}
        &\frac{v_{y_i, r, 2}^{(\tilde t-1)} \ell'^{(\tilde t-1)}_i (\|\bxi_i\|_2^2 + (C_7 k_2^2 m^2 v_{y_i, r, 2}^{(0)4})^{-1} \cdot (|S_{y_i, r}^{(\tilde t-1)}|/n))}{v_{y_k, r, 2}^{(\tilde t-1)} \ell'^{(\tilde t-1)}_k (\|\bxi_k\|_2^2 + (C_7/k_1^2) \sigma_p^2 d \cdot (|S_{y_k, r}^{(\tilde t-1)}|/n))}\\
        =& \frac{\ell'^{(\tilde t-1)}_i (\|\bxi_i\|_2^2 + (C_7 k_2^2 m^2 v_{y_i, r, 2}^{(0)4})^{-1} \cdot (|S_{y_i, r}^{(\tilde t-1)}|/n))}{\ell'^{(\tilde t-1)}_k (\|\bxi_k\|_2^2 + (C_7/k_1^2) \sigma_p^2 d \cdot (|S_{y_k, r}^{(\tilde t-1)}|/n))}\\
        \le& \frac{\ell'^{(\tilde t-1)}_i}{\ell'^{(\tilde t-1)}_k} \cdot \max\bigg\{\frac{\|\bxi_i\|_2^2}{\|\bxi_k\|_2^2}, \frac{1}{C_7^2 k_1^2 k_2^2 m^2 v_{y_i, r, 2}^{(0)4} \sigma_p^2 d}\bigg\}\\        
        \le& \frac{\ell'^{(\tilde t-1)}_i \|\bxi_i\|_2^2}{\ell'^{(\tilde t-1)}_k \|\bxi_k\|_2^2} \\
        \le& \exp(-0.4 \kappa)\\
        \le& 1.
    \end{align*}
    The first equality is from $y_i = y_k$, so $v_{y_i, r, 2}^{(\tilde t-1)} = v_{y_k, r, 2}^{(\tilde t-1)}$; The first inequality is from $\frac{a+b}{c+d} \le \max\{\frac{a}{c}, \frac{b}{d}\}$ and $|S_{y_i, r}^{(\tilde t-1)}| = |S_{y_k, r}^{(\tilde t-1)}|$; The second inequality is due to $v_0 > \Theta(n^{1/4}\sigma_p^{-1/2}d^{-1/4}m^{-1/2})$, so $(C_7^2 k_1^2 k_2^2 m^2 v_0^4 \sigma_p^2 d)^{-1} = o(1)$; The third inequality is from Lemma \ref{lemma: noise bound} that
    \begin{align*}
         \big|\|\bxi_k\|_2^2-\sigma_p^2d \big| = O\big(\sigma_p^2 \cdot \sqrt{d\log(4n/\delta)}\big).
    \end{align*}
    So it follows that
    \begin{align*}
        \Big|v_{y_i, r, 2}^{(\tilde t-1)} \ell'^{(\tilde t-1)}_i (\|\bxi_i\|_2^2 + (C_7 k_2^2)^{-1} \sigma_p^2 d \cdot (|S_{y_i, r}^{(\tilde t-1)}|/n))\Big| \le \Big|v_{y_k, r, 2}^{(\tilde t)} \ell'^{(\tilde t-1)}_k (\|\bxi_k\|_2^2 + (C_7/k_1^2) \sigma_p^2 d \cdot (|S_{y_k, r}^{(\tilde t-1)}|/n))\Big|.
    \end{align*}
    Therefore,
    \begin{align*}
        v_{y_k, r, 2}^{(\tilde t)} \sigma(\la \wb_{y_k, r}^{(\tilde t)}, \bxi_k \ra) - v_{y_i, r, 2}^{(\tilde t)} \sigma(\la \wb_{y_i, r}^{(\tilde t)}, \bxi_i \ra)
        \le v_{y_k, r, 2}^{(\tilde t-1)} \sigma(\la \wb_{y_k, r}^{(\tilde t-1)}, \bxi_k \ra) - v_{y_i, r, 2}^{(\tilde t-1)} \sigma(\la \wb_{y_i, r}^{(\tilde t-1)}, \bxi_i \ra) \le \kappa.
    \end{align*}
    Meanwhile, from \ref{eq: output_diff_c1} we have
    \begin{align*}
    &y_k \cdot f(\wb^{(\tilde t)}, \bV^{(\tilde t)}; \bx_k) - y_i \cdot f(\wb^{(\tilde t)}, \bV^{(\tilde t)}; \bx_i) \le 0.5 + \kappa, \\
        &\frac{\ell'^{(\tilde t)}_i}{\ell'^{(\tilde t)}_k} \le \exp(y_k \cdot f(\wb^{(\tilde t)}, \bV^{(\tilde t)}; \bx_k) - y_i \cdot f(\wb^{(\tilde t)}, \bV^{(\tilde t)}; \bx_i)) \le \exp(0.5 + \kappa) = C_7.
    \end{align*}

    Then we prove conclusion 2. Denote $\ell'^{(t)}_{min} = -\min\limits_{i \in [n]} |\ell'^{(t)}_i|$ and $\ell'^{(t)}_{max} = -\max\limits_{i \in [n]} |\ell'^{(t)}_i|$ . Then we could lower bound $v_{j,r,2}^{(t)}$ as
\begin{align*}
    v_{j,r,2}^{(t)}  =& v_{j,r,2}^{(t-1)} - \frac{j \eta}{n}\sum_{i=1}^{n}{\ell'^{(t-1)}_i}y_i\la \wb_{j,r}^{(t-1)},\bxi_i \ra \mathbb{I}(\la\wb_{j,r}^{(t-1)},\bxi_i\ra >0)\\
    =& v_{j,r,2}^{(t-1)} +\frac{\eta}{n} \sum_{k=1,y_k\neq j}^n {\ell'^{(t-1)}_k} \la \wb_{j,r}^{(t-1)},\bxi_k \ra \mathbb{I}(\la\wb_{j,r}^{(t-1)},\bxi_k\ra >0) \\
    &- \frac{\eta}{n} \sum_{k=1,y_k=j}^n {\ell'^{(t-1)}_k} \la \wb_{j,r}^{(t-1)},\bxi_k \ra \mathbb{I}(\la\wb_{j,r}^{(t-1)},\bxi_k\ra >0)\\
    \ge& v_{j,r,2}^{(t-1)} + \frac{\eta}{n} \sum_{k=1, y_k \neq j}^n \ell'^{(t)}_{max} \bigg(\la \wb_{j,r}^{(0)}, \bxi_k \ra + \underline{\rho}_{j,r,k}^{(t-1)} +\sum_{i=1, i\neq k}^n \rho_{j,r,i}^{(t-1)} \frac{\la \bxi_i, \bxi_k \ra}{\|\bxi_i\|_2^2}\bigg)\\
    &- \frac{\eta}{n} \sum_{k=1, y_k = j}^n \ell'^{(t)}_{min} \bigg(\la \wb_{j,r}^{(0)}, \bxi_k \ra + \overline{\rho}_{j,r,k}^{(t-1)} +\sum_{i=1, i\neq k}^n \rho_{j,r,i}^{(t-1)} \frac{\la \bxi_i, \bxi_k \ra}{\|\bxi_i\|_2^2}\bigg)\\
    \ge& v_{j,r,2}^{(t-1)} + \frac{\eta}{n} \sum_{k=1, y_k \neq j}^n \ell'^{(t)}_{max} \bigg(\la \wb_{j,r}^{(0)}, \bxi_k \ra + \underline{\rho}_{j,r,k}^{(t-1)} +\sum_{i=1, i\neq k}^n \overline{\rho}_{j,r,i}^{(t-1)} \frac{\la \bxi_i, \bxi_k \ra}{\|\bxi_i\|_2^2}\bigg)\\
    &- \frac{\eta}{n} \sum_{k=1, y_k = j}^n \ell'^{(t)}_{min} \bigg(\la \wb_{j,r}^{(0)}, \bxi_k \ra + \overline{\rho}_{j,r,k}^{(t-1)} +\sum_{i=1, i\neq k}^n \underline{\rho}_{j,r,i}^{(t-1)} \frac{\la \bxi_i, \bxi_k \ra}{\|\bxi_i\|_2^2}\bigg)\\
    =& v_{j,r,2}^{(t-1)} + \frac{\eta}{n} \ell'^{(t)}_{max} \sum_{k=1, y_k \neq j}^n \la \wb_{j,r}^{(0)}, \bxi_k \ra + \underline{\rho}_{j,r,k}^{(t)}\bigg(1 - \frac{\ell'^{(t)}_{min}}{\ell'^{(t)}_{max}} \sum_{i=1, i \neq k}^n \frac{|\la \bxi_i, \bxi_k \ra|}{\|\bxi_i\|_2^2}\bigg)\\
    &-\frac{\eta}{n} \ell'^{(t)}_{min}
    \sum_{k=1, y_k = j}^n  \la \wb_{j,r}^{(0)}, \bxi_k \ra + \overline{\rho}_{j,r,k}^{(t)}\bigg(1  - \frac{\ell'^{(t)}_{max}}{\ell'^{(t)}_{min}} \sum_{i=1, i \neq k}^n \frac{|\la \bxi_i, \bxi_k \ra|}{\|\bxi_i\|_2^2}\bigg)\\
    \ge& v_{j,r,2}^{(t-1)} + \frac{\eta}{n} {\ell'^{(t)}_{max}} \sum_{k=1, y_k \neq j}^n 2 \sqrt{\log(8mn/\delta)}\cdot \sigma_0 \sigma_p\sqrt{d} + \underline{\rho}_{j,r,k}^{(t)}(1 - C_7 \sum_{i=1}^n 4 \sigma_p^2 \cdot \sqrt{\log(4n^2/\delta)/d})\\
    &- \frac{\eta}{n} {\ell'^{(t)}_{min}}\sum_{k=1, y_k = j}^n -2 \sqrt{\log(8mn/\delta)}\cdot \sigma_0 \sigma_p\sqrt{d} + \underline{\rho}_{j,r,k}^{(t)}(1 - \frac{1}{C_7} \sum_{i=1}^n 4 \sigma_p^2 \cdot \sqrt{\log(4n^2/\delta)/d})\\
    \ge& v_{j,r,2}^{(t-1)} - \eta \sqrt{\log(8mn/\delta)}\cdot \sigma_0 \sigma_p\sqrt{d} - \frac{\eta}{n}\sum_{k=1, y_k \neq j}^n 0.5 \underline{\rho}_{j,r,k}^{(t)} + \frac{\eta}{n} \sum_{k=1, y_k = j}^n 0.5 \overline{\rho}_{j,r,k}^{(t)}\\
    \ge& v_{j,r,2}^{(t-1)}.
\end{align*}
The first inequality is from the definition of $\ell'^{(t)}_{min}$ and $\ell'^{(t)}_{max}$; The second inequality is from the monotonicity of $\rho$ so that $\overline{\rho} \ge 0$ and $\underline{\rho} \le 0$; The third inequality is from Lemma \ref{lemma: noise bound} and Lemma \ref{lemma: preliminary_initial inner product} as well as the first induction hypothesis; The fourth inequality is from Condition \ref{condition: main condition} the definition of d; The last inequality is from Condition \ref{condition: main condition} the definition of $\eta$ and the monotonicity of $\rho$; The last equality also comes from the monotonicity of $\rho$ so the negative term is canceled out.

    Next we prove conclusion 3. Recall the update equation of $\la \wb_{j,r}^{(t)}, \bxi_k \ra$ when $y_k = j$:
    \begin{align*}
    \la\wb_{j,r}^{(t)},\bxi_k\ra &= \la\wb_{j,r}^{( t-1)},\bxi_k\ra - \frac{\eta}{n} v_{j,r,2}^{( t-1)} \|\bxi_k\|_2^2 {\ell'}_{k}^{( t-1)} -\frac{j \eta}{n} v_{j,r,2}^{( t-1)} \sum_{i=1,i\neq k}^n y_i {\ell'}_{i}^{( t-1)} \la \bxi_k,\bxi_i \ra \mathbb{I}(\la\wb_{j,r}^{( t-1)},\bxi_i\ra >0).
    \end{align*}
    Note that $S_k^{(0)} \subseteq S_k^{(\tilde t)}, S_{j,r}^{(0)} \subseteq S_{j,r}^{(\tilde t)}$ both equal to $\la\wb_{j,r}^{(\tilde t)},\bxi_k\ra > 0$ so we only need to prove this. From induction hypothesis, we have $\la\wb_{j,r}^{(\tilde t-1)},\bxi_k\ra > 0$, and for the remaining terms
    \begin{align*}
        &- \frac{\eta}{n} v_{j,r,2}^{(\tilde t-1)} \|\bxi_k\|_2^2 {\ell'}_{k}^{(\tilde t-1)} -\frac{j \eta}{n} v_{j,r,2}^{(\tilde t-1)} \sum_{i=1,i\neq k}^n y_i {\ell'}_{i}^{(\tilde t-1)} \la \bxi_k,\bxi_i \ra \mathbb{I}(\la\wb_{j,r}^{(\tilde t-1)},\bxi_i\ra >0)\\
        \ge&  - \frac{\eta}{n} v_{j,r,2}^{(\tilde t-1)} \|\bxi_k\|_2^2 {\ell'}_{k}^{(\tilde t-1)} - \frac{\eta}{n} v_{j,r,2}^{(\tilde t-1)} \sum_{i=1,i\neq k}^n |{\ell'}_{i}^{(\tilde t-1)} \la \bxi_k,\bxi_i \ra|\\
        \ge& \frac{\eta}{n} v_{j,r,2}^{(\tilde t-1)} \|\bxi_k\|_2^2 |{\ell'}_{k}^{(\tilde t-1)}| - \frac{\eta}{n} v_{j,r,2}^{(\tilde t-1)} C_7 |\ell'^{(\tilde t-1)}_k | \cdot 2\sigma_p^2 \sqrt{d \log(4n^2/\delta)}\\
        =&\frac{\eta}{n}v_{j,r,2}^{(\tilde t-1)}|{\ell'}_{k}^{(\tilde t-1)}|(\|\bxi_k\|_2^2 - 2 C_7 \sigma_p^2 \sqrt{d \log(4n^2/\delta)})\\
        >& 0.
    \end{align*}
    The first inequality is from triangle inequality; The second inequality is from the first induction hypothesis and Lemma \ref{lemma: noise bound}; The last inequality is from Condition \ref{condition: main condition} about the definition of \(d\).

    So $\la\wb_{j,r}^{(t)},\bxi_k\ra > 0$ for $t = \tilde t$, then it follows that
    \begin{align*}
        S_k^{(\tilde t-1)} \subseteq S_k^{(\tilde t)}, S_{j,r}^{(\tilde t-1)} \subseteq S_{j,r}^{(\tilde t)}.
    \end{align*}
    And we have
    \begin{align*}
        S_k^{(0)} \subseteq S_k^{(\tilde t)}, S_{j,r}^{(0)} \subseteq S_{j,r}^{(\tilde t)}.
    \end{align*}

    Then we prove conclusion 4. Recall the update equation of $v_{y_k, r, 1}^{(t)}$ and $\la \wb_{y_k, r}^{(t)}, y_k \bmu \ra$:
    \begin{align*}
    \la\wb_{y_k,r}^{(t)},y_k \bmu\ra &= \la\wb_{j,r}^{(t-1)},y_k \bmu\ra - \frac{\eta}{n} v_{y_k,r,1}^{(t-1)} \|\bmu\|_2^2 \sum_{i=1}^n\ell'^{(t-1)}_i \mathbb{I}(\la\wb_{y_k,r}^{(t-1)},y_i \cdot \bmu\ra >0),\\
    v_{y_k,r,1}^{(t)} & = v_{y_k,r,1}^{(t-1)} - y_k \frac{\eta}{n}\sum_{i=1}^{n}\ell'^{(t-1)}_i \la \wb_{y_k,r}^{(t-1)},\bmu \ra \mathbb{I}(\la\wb_{y_k,r}^{(t-1)},y_i \cdot \bmu\ra >0).
    \end{align*}
    So we have the update equation of their ratio
    \begin{align*}
        \frac{v_{y_k,r,1}^{(t)}}{\la\wb_{y_k,r}^{(t)},y_k \bmu\ra} &= \frac{v_{y_k,r,1}^{(t-1)} - y_k \frac{\eta}{n}\sum_{i=1}^{n}\ell'^{(t-1)}_i \la \wb_{y_k,r}^{(t-1)},\bmu \ra \mathbb{I}(\la\wb_{y_k,r}^{(t-1)},y_i \cdot \bmu\ra >0)}{\la\wb_{y_k,r}^{(t-1)},y_k \bmu\ra - \frac{\eta}{n} v_{y_k,r,1}^{(t-1)} \|\bmu\|_2^2 \sum_{i=1}^n\ell'^{(t-1)}_i \mathbb{I}(\la\wb_{y_k,r}^{(t-1)},y_i \cdot \bmu\ra >0)}\\
        &= \frac{v_{y_k,r,1}^{(t-1)} - \frac{\eta}{n} \la \wb_{y_k, r}^{(t-1)}, y_k \bmu \ra \sum_{i=1}^n \ell'^{(t-1)}_i \mathbb{I}(\la\wb_{y_k,r}^{(t-1)},y_i \cdot \bmu\ra >0)}{\la\wb_{y_k,r}^{(t-1)},y_k \bmu\ra - \frac{\eta}{n} v_{y_k,r,1}^{(t-1)} \|\bmu\|_2^2 \sum_{i=1}^n\ell'^{(t-1)}_i \mathbb{I}(\la\wb_{y_k,r}^{(t-1)},y_i \cdot \bmu\ra >0)}\\
        &= \frac{\frac{v_{y_k,r,1}^{(t-1)}}{\la\wb_{y_k,r}^{(t-1)},y_k \bmu\ra} - \frac{\eta}{n} \sum_{i=1}^n \ell'^{(t-1)}_i \mathbb{I}(\la\wb_{y_k,r}^{(t-1)},y_i \cdot \bmu\ra >0)}{1 - \frac{\eta}{n}\frac{v_{y_k,r,1}^{(t-1)}}{\la\wb_{y_k,r}^{(t-1)},y_k \bmu\ra} \|\bmu\|_2^2 \sum_{i=1}^n\ell'^{(t-1)}_i \mathbb{I}(\la\wb_{y_k,r}^{(t-1)},y_i \cdot \bmu\ra >0)}\\
        &= \frac{v_{y_k,r,1}^{(t-1)}}{\la\wb_{y_k,r}^{(t-1)},y_k \bmu\ra} \cdot \frac{1 - \frac{\eta}{n} \frac{\la\wb_{y_k,r}^{(t-1)},y_k \bmu\ra}{v_{y_k,r,1}^{(t-1)}} \sum_{i=1}^n \ell'^{(t-1)}_i \mathbb{I}(\la\wb_{y_k,r}^{(t-1)},y_i \cdot \bmu\ra >0)}{1 - \frac{\eta}{n}\frac{v_{y_k,r,1}^{(t-1)}}{\la\wb_{y_k,r}^{(t-1)},y_k \bmu\ra} \|\bmu\|_2^2 \sum_{i=1}^n\ell'^{(t-1)}_i \mathbb{I}(\la\wb_{y_k,r}^{(t-1)},y_i \cdot \bmu\ra >0)}.
    \end{align*}
    From the induction hypothesis, the conclusion holds at time t-1, so there exist $k_3, k_4 = \Theta(1)$ that $\frac{k_3}{\|\bmu\|_2} \le \frac{v_{y_k, r, 1}^{(t-1)}}{\la \wb_{y_k, r}^{(t-1)}, y_k\bmu \ra} \le \frac{k_4 m v_0^2 \sigma_p^2 d}{n \|\bmu\|_2^2}$ for all $k \in [n], r \in [m]$. 
    
    First we could upper bound the ratio as
    \begin{align*}
        \frac{v_{j,r,1}^{(t)}}{\la\wb_{y_k,r}^{(t)},y_k \bmu\ra} &=  \frac{v_{y_k, r, 1}^{(t-1)}}{\la \wb_{y_k, r}^{(t-1)}, y_k\bmu \ra} \cdot \frac{1 - \frac{\la\wb_{j,r}^{(t-1)},y_k \bmu\ra}{v_{y_k,r,1}^{(t-1)}} \frac{\eta}{n} \sum_{i=1}^n \ell'^{(t-1)}_i \mathbb{I}(\la\wb_{y_k,r}^{(t-1)},y_i \cdot \bmu\ra >0)}{1 - \frac{\eta}{n}\frac{v_{y_k,r,1}^{(t-1)}}{\la\wb_{j,r}^{(t-1)},y_k \bmu\ra} \|\bmu\|_2^2 \sum_{i=1}^n\ell'^{(t-1)}_i \mathbb{I}(\la\wb_{y_k,r}^{(t-1)},y_i \cdot \bmu\ra >0)}\\
        &= \frac{v_{y_k, r, 1}^{(t-1)}}{\la \wb_{y_k, r}^{(t-1)}, y_k\bmu \ra} \cdot \frac{1 - \frac{\la\wb_{j,r}^{(t-1)},y_k \bmu\ra^2}{v_{y_k,r,1}^{(t-1)2} \|\bmu\|_2^2}   \frac{\eta}{n} \frac{v_{y_k,r,1}^{(t-1)}}{\la\wb_{j,r}^{(t-1)},y_k \bmu\ra} \|\bmu\|_2^2 \sum_{i=1}^n \ell'^{(t-1)}_i \mathbb{I}(\la\wb_{y_k,r}^{(t-1)},y_i \cdot \bmu\ra >0)}{1 - 
 \frac{\eta}{n} \frac{v_{y_k,r,1}^{(t-1)}}{\la\wb_{j,r}^{(t-1)},y_k \bmu\ra} \|\bmu\|_2^2 \sum_{i=1}^n \ell'^{(t-1)}_i \mathbb{I}(\la\wb_{y_k,r}^{(t-1)},y_i \cdot \bmu\ra >0)}\\
        &\le \frac{v_{y_k, r, 1}^{(t-1)}}{\la \wb_{y_k, r}^{(t-1)}, y_k\bmu \ra} \le \frac{k_4 m v_0^2 \sigma_p^2 d}{n \|\bmu\|_2^2}.
    \end{align*}
    The first inequality is from the induction hypothesis that $\frac{k_3}{\|\bmu\|_2} \le \frac{v_{y_k, r, 1}^{(t-1)}}{\la \wb_{y_k, r}^{(t-1)}, y_k\bmu \ra}$.

    Then to lower bound the ratio of two layers, note that $\frac{v_{y_k,r,1}^{(t)}}{\la\wb_{y_k,r}^{(t)},y_k \bmu\ra}$ will decrease only if 
    $$
    \frac{\la\wb_{y_k,r}^{(t-1)},y_k \bmu\ra}{v_{y_k,r,1}^{(t-1)}} \le \frac{v_{y_k,r,1}^{(t-1)}}{\la\wb_{y_k,r}^{(t-1)},y_k \bmu\ra} \|\bmu\|_2^2,
    $$
    which suffices to
    $$
    \frac{v_{y_k,r,1}^{(t-1)}}{\la\wb_{y_k,r}^{(t-1)},y_k \bmu\ra} \ge \frac{1}{\|\bmu\|_2^2}.
    $$
    Therefore, when $\frac{v_{y_k,r,1}^{(t-1)}}{\la\wb_{y_k,r}^{(t-1)},y_k \bmu\ra}$ decreases to the level of $\frac{1}{\|\bmu\|_2^2}$, the next iteration $\frac{v_{y_k,r,1}^{(t)}}{\la\wb_{y_k,r}^{(t)},y_k \bmu\ra}$ will not decrease and remain the same from then on, which implies $\frac{v_{y_k,r,1}^{(t)}}{\la\wb_{y_k,r}^{(t)},y_k \bmu\ra} \ge \frac{1}{\|\bmu\|_2^2}$. Therefore, conclusion 4 holds at time t.

    Finally we prove the last conclusion. Recall the update equation of $v_{y_k, r, 2}^{(t)}$ and $\la \wb_{y_k, r}^{(t)}, y_k \bmu \ra$:
    \begin{align*}
 \la\wb_{y_k,r}^{(t)},\bxi_k\ra &= \la\wb_{j,r}^{(t-1)},\bxi_k\ra - \frac{\eta}{n} v_{j,r,2}^{(t-1)} \|\bxi_k\|_2^2 \ell'^{(t-1)}_{k} -\frac{y_k \eta}{n} v_{j,r,2}^{(t-1)} \sum_{i=1,i\neq k}^n y_i \ell'^{(t-1)}_i \la \bxi_k,\bxi_i \ra \mathbb{I}(\la\wb_{j,r}^{(t-1)},\bxi_i\ra >0),\\ 
    v_{y_k,r,2}^{(t)} & = v_{y_k,r,2}^{(t-1)} - \frac{y_k \eta}{n}\sum_{i=1}^{n}\ell'^{(t-1)}_iy_i\la \wb_{y_k,r}^{(t-1)},\bxi_i \ra \mathbb{I}(\la\wb_{y_k,r}^{(t-1)},\bxi_i\ra >0).
\end{align*}

So we have
\begin{align*}
    \frac{v_{y_k, r, 2}^{(t)}}{\la \wb_{y_k, r}^{(t)}, \bxi_k \ra} &= \frac{v_{y_k,r,2}^{(t-1)} - \frac{y_k \eta}{n}\sum_{i=1}^{n}\ell'^{(t-1)}_iy_i\la \wb_{y_k,r}^{(t-1)},\bxi_i \ra \mathbb{I}(\la\wb_{y_k,r}^{(t-1)},\bxi_i\ra >0)}{\la\wb_{y_k,r}^{(t-1)},\bxi_k\ra - \frac{\eta}{n} v_{y_k,r,2}^{(t-1)} \|\bxi_k\|_2^2 \ell'^{(t-1)}_{k} -\frac{y_k \eta}{n} v_{y_k,r,2}^{(t-1)} \sum_{i=1,i\neq k}^n y_i \ell'^{(t-1)}_i \la \bxi_k,\bxi_i \ra \mathbb{I}(\la\wb_{y_k,r}^{(t-1)},\bxi_i\ra >0)}\\
    &= \frac{\frac{v_{y_k, r, 2}^{(t-1)}}{\la \wb_{y_k, r}^{(t-1)}, \bxi_k \ra} - \frac{y_k \eta}{n}\sum_{i=1}^n \ell'^{(t-1)} y_i \frac{\la \wb_{y_k,r}^{(t-1)},\bxi_i \ra}{\la \wb_{y_k,r}^{(t-1)},\bxi_k \ra} \mathbb{I}(\la\wb_{y_k,r}^{(t-1)},\bxi_i\ra >0)}{1 - \frac{\eta}{n} \frac{v_{y_k, r, 2}^{(t-1)}}{\la \wb_{y_k, r}^{(t-1)}, \bxi_k \ra} \|\bxi_k\|^2 \ell'^{(t-1)}_k - \frac{y_k \eta}{n} \frac{v_{y_k, r, 2}^{(t-1)}}{\la \wb_{y_k, r}^{(t-1)}, \bxi_k \ra} \sum_{i=1,i\neq k}^n y_i \ell'^{(t-1)}_i \la \bxi_k,\bxi_i \ra \mathbb{I}(\la\wb_{y_k,r}^{(t-1)},\bxi_i\ra >0)}\\
    &= \frac{v_{y_k, r, 2}^{(t-1)}}{\la \wb_{y_k, r}^{(t-1)}, \bxi_k \ra} \cdot \frac{1 - \frac{\la \wb_{y_k, r}^{(t-1)}, \bxi_k \ra}{v_{y_k, r, 2}^{(t-1)}} \cdot \frac{y_k \eta}{n}\sum_{i=1}^n \ell'^{(t-1)} y_i \frac{\la \wb_{y_k,r}^{(t-1)},\bxi_i \ra}{\la \wb_{y_k,r}^{(t-1)},\bxi_k \ra} \mathbb{I}(\la\wb_{y_k,r}^{(t-1)},\bxi_i\ra >0)}{1 - \frac{\eta}{n} \frac{v_{y_k, r, 2}^{(t-1)}}{\la \wb_{y_k, r}^{(t-1)}, \bxi_k \ra} \|\bxi_k\|^2 \ell'^{(t-1)}_k - \frac{y_k \eta}{n} \frac{v_{y_k, r, 2}^{(t-1)}}{\la \wb_{y_k, r}^{(t-1)}, \bxi_k \ra} \sum_{i=1,i\neq k}^n y_i \ell'^{(t-1)}_i \la \bxi_k,\bxi_i \ra \mathbb{I}(\la\wb_{y_k,r}^{(t-1)},\bxi_i\ra >0)}.
\end{align*}

From the induction hypothesis, the conclusion holds at time $t-1$, so there exist $k_1, k_2 = \Theta(1)$ that $\frac{k_1\sqrt{n}}{\sigma_p \sqrt{d}}\le \frac{v_{y_k,r,2}^{(t-1)}}{\la \wb_{y_k,r}^{(t-1)}, \bxi_k \ra} \le k_2 mv_0^2$ for all $k \in [n], r \in [m]$. 

We first upper bound the ratio as
\begin{align*}
     \frac{v_{y_k, r, 2}^{(t)}}{\la \wb_{y_k, r}^{(t)}, \bxi_k \ra} &= \frac{v_{y_k, r, 2}^{(t-1)}}{\la \wb_{y_k, r}^{(t-1)}, \bxi_k \ra} \cdot \frac{1 - \frac{\la \wb_{y_k, r}^{(t-1)}, \bxi_k \ra}{v_{y_k, r, 2}^{(t-1)}} \cdot \frac{y_k \eta}{n}\sum_{i=1}^n \ell'^{(t-1)} y_i \frac{\la \wb_{y_k,r}^{(t-1)},\bxi_i \ra}{\la \wb_{y_k,r}^{(t-1)},\bxi_k \ra} \mathbb{I}(\la\wb_{y_k,r}^{(t-1)},\bxi_i\ra >0)}{1 - \frac{\eta}{n} \frac{v_{y_k, r, 2}^{(t-1)}}{\la \wb_{y_k, r}^{(t-1)}, \bxi_k \ra} \|\bxi_k\|^2 \ell'^{(t-1)}_k - \frac{y_k \eta}{n} \frac{v_{y_k, r, 2}^{(t-1)}}{\la \wb_{y_k, r}^{(t-1)}, \bxi_k \ra} \sum_{i=1,i\neq k}^n y_i \ell'^{(t-1)}_i \la \bxi_k,\bxi_i \ra \mathbb{I}(\la\wb_{y_k,r}^{(t-1)},\bxi_i\ra >0)}\\
     &\le \frac{v_{y_k, r, 2}^{(t-1)}}{\la \wb_{y_k, r}^{(t-1)}, \bxi_k \ra} \cdot \frac{1 - \frac{\eta}{n} \sum_{i=1}^n \ell'^{(t-1)}_i \frac{\la \wb_{y_k, r}^{(t-1)}, \bxi_i \ra}{v_{y_k, r, 2}^{(t-1)}} \mathbb{I}(\la\wb_{y_k,r}^{(t-1)},\bxi_i\ra >0)}{1 - \frac{\eta}{2n} \frac{v_{y_k, r, 2}^{(t-1)}}{\la \wb_{y_k, r}^{(t-1)}, \bxi_k \ra} \|\bxi_k\|^2 \ell'^{(t-1)}_k}\\
     &=\frac{v_{y_k, r, 2}^{(t-1)}}{\la \wb_{y_k, r}^{(t-1)}, \bxi_k \ra} \cdot \frac{1 - \frac{\eta}{2n} \frac{v_{y_k, r, 2}^{(t-1)}}{\la \wb_{y_k, r}^{(t-1)}, \bxi_k \ra} \|\bxi_k\|^2 \ell'^{(t-1)}_k \cdot \sum_{i=1}^n \frac{2 \ell'^{(t-1)}_i}{\ell'^{(t-1)}_k \|\bxi_k\|^2} \frac{\la \wb_{y_k, r}^{(t-1)}, \bxi_i \ra \la \wb_{y_k, r}^{(t-1)}, \bxi_k \ra}{v_{y_k, r, 2}^{(t-1)2}} \mathbb{I}(\la\wb_{y_k,r}^{(t-1)},\bxi_i\ra >0)}{1 - \frac{\eta}{2n} \frac{v_{y_k, r, 2}^{(t-1)}}{\la \wb_{y_k, r}^{(t-1)}, \bxi_k \ra} \|\bxi_k\|^2 \ell'^{(t-1)}_k}\\
     &\le \frac{v_{y_k, r, 2}^{(t-1)}}{\la \wb_{y_k, r}^{(t-1)}, \bxi_k \ra} \le  k_2 mv_0^2.
\end{align*}
The first inequality is from Condition \ref{condition: main condition} and the second inequality is from induction hypothesis that $\frac{k_1\sqrt{n}}{\sigma_p \sqrt{d}}\le \frac{v_{y_k,r,2}^{(t-1)}}{\la \wb_{y_k,r}^{(t-1)}, \bxi_k \ra}$.

Then to lower bound the ratio of two layers, we have
\begin{align*}
    \frac{v_{y_k, r, 2}^{(t)}}{\la \wb_{y_k, r}^{(t)}, \bxi_k \ra} &= \frac{v_{y_k, r, 2}^{(t-1)}}{\la \wb_{y_k, r}^{(t-1)}, \bxi_k \ra} \cdot \frac{1 - \frac{\eta}{n} \sum_{i=1}^n \ell'^{(t-1)}_i \frac{\la \wb_{y_k, r}^{(t-1)}, \bxi_i \ra}{v_{y_k, r, 2}^{(t-1)}} \mathbb{I}(\la\wb_{y_k,r}^{(t-1)},\bxi_i\ra >0)}{1 - \Theta(1) \frac{\eta}{n} \frac{v_{y_k, r, 2}^{(t-1)}}{\la \wb_{y_k, r}^{(t-1)}, \bxi_k \ra} \|\bxi_k\|^2 \ell'^{(t-1)}_k},
\end{align*}
So $\frac{v_{y_k, r, 2}^{(t)}}{\la \wb_{y_k, r}^{(t)}, \bxi_k \ra}$ will decrease only if 
\begin{align*}
    \frac{\eta}{n} \sum_{i=1}^n \ell'^{(t-1)}_i \frac{\la \wb_{y_k, r}^{(t-1)}, \bxi_i \ra}{v_{y_k, r, 2}^{(t-1)}} \mathbb{I}(\la\wb_{y_k,r}^{(t-1)},\bxi_i\ra >0) \le \Theta(1) \frac{\eta}{n} \frac{v_{y_k, r, 2}^{(t-1)}}{\la \wb_{y_k, r}^{(t-1)}, \bxi_k \ra} \|\bxi_k\|^2 \ell'^{(t-1)}_k,
\end{align*}
which suffices to
\begin{align*}
    \frac{v_{y_k, r, 2}^{(t)}}{\la \wb_{y_k, r}^{(t)}, \bxi_k \ra} \ge \Theta\bigg(\frac{\sqrt{n}}{\sigma_p \sqrt{d}}\bigg).
\end{align*}
Therefore, when $\frac{v_{y_k, r, 2}^{(t-1)}}{\la \wb_{y_k, r}^{(t-1)}, \bxi_k \ra}$ decreases to the level of $\frac{\sqrt{n}}{\sigma_p \sqrt{d}}$, in the next iteration $\frac{v_{y_k, r, 2}^{(t)}}{\la \wb_{y_k, r}^{(t)}, \bxi_k \ra}$ will be non-decreasing and remain the same level from then on. So there exists $k_1 = \Theta(1)$ that $\frac{v_{y_k, r, 2}^{(t)}}{\la \wb_{y_k, r}^{(t)}, \bxi_k \ra} \ge \frac{k_1 \sqrt{n}}{\sigma_p \sqrt{d}}$. Therefore, all conclusions hold at time t.
\end{proof}

\begin{proof}[Proof of Proposition \ref{prop: stage2_analysis_c1}]
    Now we could prove Proposition \ref{prop: stage2_analysis_c1}. The results are obvious at $t=T_1$ from Proposition \ref{prop: stage1_l}. Suppose that there exists $\tilde T \le T^*$ such that all the results in Proposition \ref{prop: stage2_analysis_c1} hold for all $T_1 \le t \le \tilde T-1$. We could prove that these also hold at $t=\tilde T$.

We prove (\ref{eq: neg_noise_bound_c1}) first. We consider two cases. When $\underline{\rho}_{j,r,k}^{(t-1)} \le -2\sqrt{\log(8mn/\delta)}\cdot \sigma_0 \sigma_p\sqrt{d} - 4 \sqrt{\frac{\log(4n^2/\delta)}{d}} n\alpha$, we have
\begin{align*}
    \la \wb_{j,r}^{(\tilde T-1)}, \bxi_k \ra \le \la \wb_{j,r}^{(0)}, \bxi_k \ra + \underline{\rho}_{j,r,k}^{(\tilde T-1)} + 4 \sqrt{\frac{\log(4n^2/\delta)}{d}} n\alpha \le 0.
\end{align*}
Therefore,
\begin{align*}
    \underline{\rho}_{j,r,k}^{(\tilde T)} &= \underline{\rho}_{j,r,k}^{(\tilde T-1)} + \frac{\eta}{nm} \cdot \ell'^{(\tilde T-1)}_k v_{j,r,2}^{(\tilde T-1)} \cdot \mathbb{I}(\la \wb_{j,r}^{(\tilde T-1)}, \bxi_k \ra \ge 0) \cdot \mathbb{I}(y_k = -j) \|\bxi_k\|_2^2\\
    &= \underline{\rho}_{j,r,k}^{(\tilde T-1)}\\
    &\ge -2\sqrt{\log(8mn/\delta)}\cdot \sigma_0 \sigma_p\sqrt{d} - 8 \sqrt{\frac{\log(4n^2/\delta)}{d}} n\alpha.
\end{align*}
The last inequity is from induction hypothesis. On the other hand, when $\underline{\rho}_{j,r,k}^{(t-1)} > -2\sqrt{\log(8mn/\delta)}\cdot \sigma_0 \sigma_p\sqrt{d} - 4 \sqrt{\frac{\log(4n^2/\delta)}{d}} n\alpha$, we first prove that $v_{j,r,2}^{(\tilde T-1)}$ is upper bounded. From Lemma \ref{lemma: key_lemma_stage2_c1} we know that $v^{(t)}_{j,r,2} /\la\wb_{j,r}^{(t)},\bxi_k\ra \le \Theta(m v_{j, r, 2}^{(0)2})$ for $t \le t-1$. Assume that $\frac{v^{(t)}_{j,r,2}}{\la\wb_{j,r}^{(t)},\bxi_k\ra} \le k_2 m v_0^2$ for all $j \in \{\pm 1\}$ and $k \in [n]$, and thus
\begin{align}
    v_{j,r,2}^{(\tilde T-1)} &\le k_2 m v_0^2 \cdot \la\wb_{j,r}^{(\tilde T-1)},\bxi_k\ra\nonumber\\
    &\le k_2 m v_0^2 \cdot \bigg[\la\wb_{j,r}^{(0)},\bxi_k\ra + \rho_{j,r,k}^{(\tilde T-1)} + 4 \sqrt{\frac{\log(4n^2/\delta)}{d}} n\alpha \bigg]\nonumber\\
    &\le 2 \alpha k_2 m v_0^2. \label{eq: v_bound_c1}
\end{align}

So we have
\begin{align*}
     \underline{\rho}_{j,r,k}^{(\tilde T)} &= \underline{\rho}_{j,r,k}^{(\tilde T-1)} + \frac{\eta}{nm} \cdot \ell'^{(\tilde T-1)}_k v_{j,r,2}^{(\tilde T-1)} \cdot \mathbb{I}(\la \wb_{j,r}^{(\tilde T-1)}, \bxi_k \ra \ge 0) \cdot \mathbb{I}(y_k = j) \|\bxi_k\|_2^2\\
     &\ge -2\sqrt{\log(8mn/\delta)}\cdot \sigma_0 \sigma_p\sqrt{d} - 4 \sqrt{\frac{\log(4n^2/\delta)}{d}} n\alpha - \frac{2 \alpha \eta k_2 v_0^2\sigma_p^2 d}{n}\\
     &\ge -2\sqrt{\log(8mn/\delta)}\cdot \sigma_0 \sigma_p\sqrt{d} - 8 \sqrt{\frac{\log(4n^2/\delta)}{d}} n\alpha,
\end{align*}
where the first inequity is from Lemma \ref{lemma: noise bound} and the upper bound of $v_{j,r,2}^{(\tilde T-1)}$. The last inequity is due to the definition of $\eta$ in Condition \ref{condition: main condition}.

Next we prove (\ref{eq: pos_noise_bound_c1}) holds for $t = \tilde T$. Consider the loss derivative
\begin{align}
    |\ell'^{(t)}_k| &= \frac{1}{1 + \exp\{y_k \cdot [(F_{+1}(\wb^{(t)}_{+1}, \bV^{(t)}_{+1}; \bx_k) - F_{-1}(\wb^{(t)}_{-1}, \bV^{(t)}_{-1}; \bx_k)]\}}\nonumber\\
    &\le \exp\{-y_k \cdot [(F_{+1}(\wb^{(t)}_{+1}, \bV^{(t)}_{+1}; \bx_k) - F_{-1}(\wb^{(t)}_{-1}, \bV^{(t)}_{-1}; \bx_k)]\}\nonumber\\
    &\le \exp\{-F_{y_k}(\wb^{(t)}_{y_k}, \bV^{(t)}_{y_k}; \bx_k) + 0.5\}.\label{eq: loss_bound_c1}
\end{align}
The last inequity is from Lemma \ref{lemma: output_leading_term_c1}. Then recall the update equation of $\overline{\rho}_{j,r,k}^{(t)}$:
\begin{align*}
    \overline{\rho}_{j,r,k}^{(t+1)} = \overline{\rho}_{j,r,k}^{(t)} - \frac{\eta}{nm} \cdot \ell'^{(t)}_k v_{j,r,2}^{(t)} \cdot \mathbb{I}(\la \wb_{j,r}^{(t)}, \bxi_k \ra \ge 0) \cdot \mathbb{I}(y_k = j) \|\bxi_k\|_2^2.
\end{align*}
Now assume $t_{j,r,k}$ is the last time $t \le T^*$ when $\overline{\rho}_{j,r,k}^{(t)} \le 0.5\alpha$. Then we have
\begin{align*}
    \overline{\rho}_{j,r,k}^{(\tilde T)} =& \overline{\rho}_{j,r,k}^{(t_{j,r,k})} - \underbrace{\frac{\eta}{nm} \cdot \ell'^{(t_{j,r,k})}_k v_{j,r,2}^{(t_{j,r,k})} \cdot \mathbb{I}(\la \wb_{j,r}^{(t_{j,r,k})}, \bxi_k \ra \ge 0) \cdot \mathbb{I}(y_k = j) \|\bxi_k\|_2^2}_{I_1}\\
    &-\underbrace{ \sum\limits_{t_{j,r,k} < t <\tilde T} \frac{\eta}{nm} \cdot \ell'^{(t)}_k v_{j,r,2}^{(t)} \cdot \mathbb{I}(\la \wb_{j,r}^{(t)}, \bxi_k \ra \ge 0) \cdot \mathbb{I}(y_k = j) \|\bxi_k\|_2^2}_{I_2}.
\end{align*}
The idea is that we will upper bound $|I_1| \le 0.25\alpha$ and $|I_2| \le 0.25\alpha$. So $\overline{\rho}_{j,r,k}^{(\tilde T)} \le \alpha$ holds. We first bound $I_1$:
\begin{align*}
    |I_1| &\le \frac{\eta}{nm} \cdot v_{j,r,2}^{(t_{j,r,k})} \|\bxi_k\|_2^2\\
    &\le \frac{\eta}{nm} \cdot 2 \alpha k_2 m v_0^2 \cdot \frac{3\sigma_p^2d}{2}\\
    &\le 0.25\alpha,
\end{align*}
where the second inequity is from Lemma \ref{lemma: noise bound} and (\ref{eq: v_bound_c1}). The last inequity is by the definition of $\eta$ in Condition \ref{condition: main condition}.

Then we bound $I_2$. For $t_{j,r,k} < t <\tilde T$ and $y_k = j$, we can lower bound $\la \wb_{j,r}^{(t)}, \bxi_k \ra$ as follows:
\begin{align*}
    \la \wb_{j,r}^{(t)}, \bxi_k \ra &\ge \la \wb_{j,r}^{(0)}, \bxi_k \ra + \overline{\rho}_{j,r,k}^{(t)} - 4 \sqrt{\frac{\log(4n^2/\delta)}{d}} n\alpha\\
    &\ge -2\sqrt{\log(8mn/\delta)}\cdot \sigma_0 \sigma_p\sqrt{d} + 0.5\alpha - 4 \sqrt{\frac{\log(4n^2/\delta)}{d}} n\alpha\\
    &\ge 0.25\alpha,
\end{align*}
where the second inequity is due to $\overline{\rho}_{j,r,k}^{(t)} > 0.5\alpha$ for $t > t_{j,r,k}$. From Lemma \ref{lemma: key_lemma_stage2_c1} we have $v_{j,r,2}^{(t)} \ge C_8 v_0$ for $C_8 = \Theta(1)$. Therefore, we have
\begin{align*}
    |I_2| &\le \sum\limits_{t_{j,r,k} < t <\tilde T} \frac{\eta}{nm} \cdot \exp\bigg(-\sum_{r=1}^m v_{j,r,2}^{(t)} \sigma(\la \wb_{j,r}^{(t)}, \bxi_k \ra ) + 0.5\bigg) \cdot v_{j,r,2}^{(t)} \|\bxi_k\|_2^2\\
    &\le \frac{2 \eta (\tilde T - t_{j,r,k} + 1)}{nm} \cdot \exp\bigg(-\frac{k_1 m v_0 \alpha}{4} \bigg) \cdot 2 \alpha k_2 m v_0^2 \cdot \frac{3\sigma_p^2 d}{2}\\
    &\le \frac{2\eta \tilde T}{nm} \cdot \exp(-\log(T^*)) \cdot 2 \alpha k_2 m v_0^2 \cdot \frac{3\sigma_p^2 d}{2}\\
    &= \frac{2\eta}{nm} \cdot 2 \alpha k_2 m v_0^2 \cdot \frac{3\sigma_p^2 d}{2}\\
    &= \frac{6\eta k_2 \sigma_p^2 d \alpha v_0^2}{n}\\
    &\le 0.25\alpha,
\end{align*}
where the first inequity is from (\ref{eq: loss_bound_c1}); the second inequity is from the lower bound of $\la \wb_{j,r}^{(t)}, \bxi_k \ra$ and $v_{j,r,2}^{(t)}$; the third inequity is by $\alpha = \frac{4\log(T^*)}{k_1 m v_0}$ and the last inequity is from the definition of $\eta$ in Condition \ref{condition: main condition}.

Combining the bound of $I_1$ and $I_2$, we have $\overline{\rho}_{j,r,k}^{(t)} \le \alpha$ at $t = \tilde T$.

Then we prove (\ref{eq: signal_output_bound_c1}) holds for $t = \tilde T$. Recall the update equation of $\gamma_{j,r}^{(t)}$
\begin{align*}
    \gamma_{j,r}^{(t+1)} = \gamma_{j,r}^{(t)} - \frac{\eta}{nm} \sum_{i=1}^n \ell'^{(t)}_i v_{j,r,1}^{(t)} \|\bmu\|_2^2 \mathbb{I}(\la \wb_{j,r}^{(t)}, y_i \bmu \ra).
\end{align*}
We first prove that $\gamma_{j,r}^{(\tilde T)} \ge \gamma_{j,r}^{(\tilde T-1)}$, and hence $\gamma_{j,r}^{(\tilde T)} \ge \gamma_{j,r}^{(0)} = 0$ for any $j \in \{\pm 1\}, r \in [m]$. From Proposition \ref{prop: stage1_l}, we know that $v_{j,r,1}^{(t)}$ will remain at its initial order throughout stage 1. And recall the update equation of $v_{j,r,1}^{(t)}$:
\begin{align*}
    v_{j,r,1}^{(t+1)} & = v_{j,r,1}^{(t)} - j \frac{\eta}{n}\sum_{i=1}^{n}\ell'^{(t)}_i \la \wb_{j,r}^{(t)},\bmu \ra \mathbb{I}(\la\wb_{j,r}^{(t)},y_i \cdot \bmu\ra >0).
\end{align*}
So $v_{j,r,1}^{(t)}$ will keep increasing when $\gamma_{j,r}^{(t-1)} = \la \wb_{j,r}^{(t-1)}, j \cdot \bmu \ra \ge 0$. This holds due to the inductive hypothesis. Therefore, we have $v_{j,r,1}^{(t)} \ge 0$, so it follows that $\gamma_{j,r}^{(\tilde T)} \ge \gamma_{j,r}^{(\tilde T-1)}$.

For the other hand of (\ref{eq: signal_output_bound_c1}), we prove a strengthened hypothesis that there exists a $i^* \in [n]$ with $y_{i^*} = j$ such that for $1 \le t \le T^*$ we have that
\begin{align*}
    \frac{\gamma_{j,r}^{(t)}}{\overline{\rho}_{j,r,i^*}^{(t)}} \le \frac{C_9 n \|\bmu\|_2^2}{m v_0\sigma_p^2 d}.
\end{align*}
And $i^*$ can be taken as any sample in the training set.

Recall the update equation of $\gamma_{j,r}^{(t)}$ and $\overline{\rho}_{j,r,i}^{(t)}$:
\begin{align*}
    \gamma_{j,r}^{(\tilde T)} &= \gamma_{j,r}^{(\tilde T-1)} - \frac{\eta}{nm} \sum_{i=1}^n \ell'^{(\tilde T-1)}_i v_{j,r,1}^{(\tilde T-1)} \|\bmu\|_2^2 \mathbb{I}(\la \wb_{j,r}^{(\tilde T-1)}, y_i \bmu \ra \ge 0),\\
    \overline{\rho}_{j,r,k}^{(\tilde T)} &= \overline{\rho}_{j,r,k}^{(\tilde T-1)} - \frac{\eta}{nm} \cdot \ell'^{(\tilde T-1)}_k v_{j,r,2}^{(\tilde T-1)} \cdot \mathbb{I}(\la \wb_{j,r}^{(\tilde T-1)}, \bxi_k \ra \ge 0) \cdot \mathbb{I}(y_k = j) \|\bxi_k\|_2^2.
\end{align*}
From Lemma \ref{lemma: key_lemma_stage2_c1}, for any $i^* \in S_{j,r}^{(0)}$, it holds that $i^* \in S_{j,r}^{(t)}$ for any $1 \le t \le \tilde T-1$. So we have
\begin{align*}
    \overline{\rho}_{j,r,i^*}^{(\tilde T)} &= \overline{\rho}_{j,r,i^*}^{(\tilde T-1)} - \frac{\eta}{nm} \cdot \ell'^{(\tilde T-1)}_{i^*} \cdot v_{j,r,2}^{(\tilde T-1)} \cdot \|\bxi_{i^*}\|^2  
    \ge \overline{\rho}_{j,r,i^*}^{(\tilde T-1)} - \frac{\eta}{nm} \cdot \ell'^{(\tilde T-1)}_{i^*} \cdot v_{j,r,2}^{(\tilde T-1)} \cdot \sigma_p^2 d/2.
\end{align*}
For $\gamma_{j,r}^{(\tilde T)}$ we have $y_{i^*} = j$ that
\begin{align*}
    \gamma_{j,r}^{(\tilde T)} &\le \gamma_{j,r}^{(\tilde T-1)} - \frac{\eta}{nm} \sum_{i=1}^n |\ell'^{(\tilde T-1)}_i| v_{j,r,1}^{(\tilde T-1)} \|\bmu\|_2^2 \cdot \mathbb{I}(y_i = j) \le \gamma_{j,r}^{(\tilde T-1)} - \frac{\eta}{m} C_7 \ell'^{(\tilde T-1)}_{i^*} v_{j,r,1}^{(\tilde T-1)} \|\bmu\|_2^2,
\end{align*}
where the first inequality is from triangle inequality and the second is from Lemma \ref{lemma: key_lemma_stage2_c1}. 
Then we have
\begin{align*}
    \frac{\gamma_{j,r}^{(\tilde T)}}{\overline{\rho}_{j,r,i^*}^{(\tilde T)}} &\le \max\Bigg\{\frac{\gamma_{j,r}^{(\tilde T-1)}}{\overline{\rho}_{j,r,i^*}^{(\tilde T-1)}}, \frac{ C_7 n |\ell'^{(\tilde T-1)}_{i^*}| v_{j,r,1}^{(\tilde T-1)} \|\bmu\|_2^2}{|\ell'^{(\tilde T-1)}_{i^*}| v_{j,r,2}^{(\tilde T-1)} \sigma_p^2 d/2} \Bigg\}\\ 
    &\le \max\Bigg\{\frac{\gamma_{j,r}^{(\tilde T-1)}}{\overline{\rho}_{j,r,i^*}^{(\tilde T-1)}}, \frac{ C_7 n  C_4 v_0 \|\bmu\|_2^2}{C_8 v_0 \sigma_p^2 d/2} \Bigg\}\\
    &= \max\Bigg\{\frac{\gamma_{j,r}^{(\tilde T-1)}}{\overline{\rho}_{j,r,i^*}^{(\tilde T-1)}}, \frac{C_7 C_4 n \|\bmu\|_2^2}{C_8 \sigma_p^2 d/2} \Bigg\}\\
    &\le \frac{C_9 n \|\bmu\|_2^2}{m v_0\sigma_p^2 d}.
\end{align*}
The second inequality is from the induction hypothesis (\ref{eq: v1_bound_c1}) and the second statement in \ref{lemma: key_lemma_stage2_c1}.

Then we prove (\ref{eq: v1_bound_c1}). From the third conclusion of Lemma \ref{lemma: key_lemma_stage2_c1}, we have $\Theta \Big(\frac{1}{\|\bmu\|_2}\Big) \le \frac{v_{j, r, 1}^{(t)}}{\la \wb_{j, r}^{(t)}, j\bmu \ra} \le \Theta\Big(\frac{m v_0^2 \sigma_p^2 d}{n \|\bmu\|_2^2}\Big)$ for all $j \in \{\pm 1\}, r \in [m]$. Meanwhile, we can upper bound $\la \wb_{j, r}^{(t)}, j\bmu \ra$ as
\begin{align*}
    \la \wb_{j, r}^{(t)}, j\bmu \ra &= \la \wb_{j, r}^{(0)}, j\bmu \ra + \gamma_{j, r}^{(t)}\\
    &\le \sqrt{2\log(8m/\delta)} \cdot \sigma_0\|\bmu\|_2 + \gamma_{j, r}^{(t)}\\
    &\le \frac{C_6 n \|\bmu\|_2^2 \alpha}{m v_0\sigma_p^2 d}.
\end{align*}
The first inequality is from Lemma \ref{lemma: noise bound} and the second inequality is from the upper bound of $\gamma_{j,r}^{(t)}$ we just derived. So we could upper bound $v_{j,r,1}^{(t)}$ as 
\begin{align*}
    v_{j,r,1}^{(t)} &\le \la \wb_{j, r}^{(t)}, j\bmu \ra \cdot  \Theta\bigg(\frac{m v_0^2 \sigma_p^2 d}{n \|\bmu\|_2^2}\bigg)\\
    &\le \frac{C_6 n \|\bmu\|_2^2 \alpha}{m v_0\sigma_p^2 d} \cdot \Theta\Big(\frac{m v_0^2 \sigma_p^2 d}{n \|\bmu\|_2^2}\Big)\\
    &\le C_4 v_0 \alpha.
\end{align*}
The first inequality is from Lemma \ref{lemma: key_lemma_stage2_c1} and the last inequality is due to our definition $C_4, C_6 = \Theta(1)$.

Last we prove (\ref{eq: v2_bound_c1}). From the last conclusion of Lemma \ref{lemma: key_lemma_stage2_c1}, we have $\Theta \Big(\frac{\sqrt{n}}{\sigma_p \sqrt{d}}\Big) \le \frac{v_{y_k,r,2}^{(t)}}{\la \wb_{y_k,r}^{(t)}, \bxi_k \ra} \le \Theta (mv_0^2)$ for all $k \in [n], r \in [m]$. And we can bound $\la \wb_{j,r}^{(t)}, \bxi_k \ra$ as
\begin{align*}
    \la \wb_{j,r}^{(t)}, \bxi_k \ra &= \la \wb_{j,r}^{(0)}, \bxi_k \ra + \overline{\rho}_{j,r,k}^{(t)} + \sum_{i=1, i \neq k}^n \rho_{j,r,i}^{(t)} \frac{\la \bxi_i, \bxi_k \ra}{\|\bxi_i\|_2^2}\\
    &\le 2\sqrt{\log(8mn/\delta)}\cdot \sigma_0 \sigma_p\sqrt{d} + \overline{\rho}_{j,r,k}^{(t)} + \sum_{i=1, i \neq k}^n \overline{\rho}_{j,r,i}^{(t)} \cdot2 \sigma_p^2 \sqrt{\log(4n^2/\delta)/d}\\
    &\le k \alpha.
\end{align*}
Here $k \ge 1$ is a constant. The first inequality is from Lemma \ref{lemma: noise bound} and Lemma \ref{lemma: preliminary_initial inner product}; The second inequality is from the definition of $d$ in Condition \ref{condition: main condition}. Then we can bound $v_{j,r,2}^{(t)}$ as
\begin{align*}
    v_{j,r,2}^{(t)} &\le \la \wb_{j,r}^{(t)}, \bxi_k \ra \cdot \Theta (mv_0^2)\\
    &\le k \alpha \cdot \Theta (mv_0^2)\\
    &\le C_4 mv_0^2 \alpha.
\end{align*}
Therefore, we finish the proof for all the conclusions.
\end{proof}

Then we prove that the training loss can be arbitrarily small in a long time. We have the following lemma to lower bound the output of training sample:
\begin{lemma}
\label{lemma: output_lower_c1}
    Under Condition \ref{condition: main condition}, for $T^{*, 1} \le t \le T^*$, we have
    \begin{align*}
        \sum_{r=1}^m v_{y_i, r, 2}^{(\tilde t)} \sigma(\la \wb_{y_i, r}^{(\tilde t)}, \bxi_i \ra) \ge \log(M t)
    \end{align*}
    for all $i \in [n]$. Here $M = \frac{1}{2} e^{-0.25} \eta (\frac{k_1 \|\bxi_k\|^2}{\sqrt{n}\sigma_p \sqrt{d}} + \frac{1}{C_7 k_2 m v_0} \cdot (|S_{y_k, r}^{(\tilde t-1)}|/n)) = \Theta(\frac{\eta \sigma_p \sqrt{d}}{\sqrt{n}})$.
\end{lemma}

\begin{proof}[Proof of Lemma \ref{lemma: output_lower_c1}]
    We use induction to prove this lemma. First from the analysis in stage 1, the conclusion holds at time $T^{*, 1}$. Then assume that the conclusion holds at time $\tilde t-1$, we consider the time $\tilde t$.

    We can lower bound the loss derivative as
\begin{align}
    |\ell'^{(\tilde t-1)}_k| &= \frac{1}{1 + \exp\{y_k \cdot [(F_{+1}(\bW^{(\tilde t-1)}_{+1}, \bV^{(\tilde t-1)}_{+1}; \bx_k) - F_{-1}(\bW^{(t)}_{-1}, \bV^{(t)}_{-1}; \bx_k)]\}}\nonumber\\
    &\ge \frac{1}{2} \exp(-F_{y_k}(\bW^{(\tilde t-1)}_{y_k}, \bV^{(\tilde t-1)}_{y_k}; \bx_k))\nonumber\\
    &\ge \frac{1}{2} e^{-0.25} \exp\bigg(-\sum_{r=1}^m v_{y_k, r, 2}^{(\tilde t-1)} \sigma\big(\la \wb_{y_k, r}^{(\tilde t-1)}, \bxi_k \ra\big)\bigg) .\label{eq: loss_lower_bound_c1}
\end{align}
The first inequality is from the fact that $F_{y_k}(\bW^{(t)}_{y_k}, \bV^{(t)}_{y_k}; \bx_k) \ge F_{-y_k}(\bW^{(t)}_{-y_k}, \bV^{(t)}_{-y_k}; \bx_k)$ and the second inequality is from Lemma \ref{lemma: output_leading_term_c1}.

Recall the update equation of $v_{y_k, r, 2}^{(t)} \sigma(\la \wb_{y_k, r}^{(t)}, \bxi_k \ra)$:
    \begin{align*}
        &v_{y_k, r, 2}^{(\tilde t)} \sigma(\la \wb_{y_k, r}^{(\tilde t)}, \bxi_k \ra) \\
        &\qquad= v_{y_k, r, 2}^{(\tilde t-1)} \sigma(\la \wb_{y_k, r}^{(\tilde t-1)}, \bxi_k \ra) - \frac{\eta}{n} v_{y_k, r, 2}^{(\tilde t-1)2} \|\bxi_k\|_2^2 \ell'^{(\tilde t-1)}_k- \frac{y_k \eta}{n} \la \wb_{y_k, r}^{(\tilde t-1)}, \bxi_k \ra \sum_{i=1}^n \ell'^{(\tilde t-1)}_i y_i \la \wb_{y_k, r}^{(\tilde t-1)}, \bxi_i \ra \mathbb{I}(\la \wb_{y_k, r}^{(\tilde t-1)}, \bxi_i \ra > 0)\\
        &\qquad\quad+ \frac{\eta^2}{n^2} v_{y_k, r, 2}^{(\tilde t-1)} \sum_{i=1, i\neq k}^n y_i \ell'^{(\tilde t-1)}_i \la \bxi_k, \bxi_i \ra \mathbb{I}(\la \wb_{y_k, r}^{(\tilde t-1)}, \bxi_i \ra \ge 0) \sum_{i=1}^n \ell'^{(\tilde t-1)}_i y_i \la \wb_{y_k, r}^{(\tilde t-1)}, \bxi_i \ra \mathbb{I}(\la \wb_{y_k, r}^{(\tilde t-1)}, \bxi_i \ra \ge 0).
    \end{align*}
    
From the analysis in Lemma \ref{lemma: key_lemma_stage2_c1}, We could assume that $\frac{k_1 \sqrt{n}}{\sigma_p \sqrt{d}} \le  \frac{v^{(t)}_{j,r,2}}{\la\wb_{j,r}^{(t)},\bxi_k\ra} \le k_2 m v_0^2$ for all $j = y_k$, $k \in [n]$ and lower bound $v_{y_k, r, 2}^{(\tilde t)} \sigma(\la \wb_{y_k, r}^{(\tilde t)}, \bxi_k \ra)$ that
    \begin{align*}
        v_{y_k, r, 2}^{(\tilde t)} \sigma(\la \wb_{y_k, r}^{(\tilde t)}, \bxi_k \ra) \ge& v_{y_k, r, 2}^{(\tilde t-1)} \sigma(\la \wb_{y_k, r}^{(\tilde t-1)}, \bxi_k \ra) - 
\frac{k_1 \eta }{\sqrt{n} \sigma_p \sqrt{d}} v_{y_k, r, 2}^{(\tilde t-1)} \la \wb_{y_k, r}^{(\tilde t -1)}, \bxi_k \ra \|\bxi_k\|_2^2 \ell'^{(\tilde t-1)}_i\\
&- \frac{\eta}{C_7 k_2 m v_0^2} v_{y_k, r, 2}^{(\tilde t-1)} \la \wb_{y_k, r}^{(\tilde t -1)}, \bxi_k \ra \ell'^{(\tilde t-1)}_k \cdot (|S_{y_k, r}^{(\tilde t-1)}|/n))\\
=& v_{y_k, r, 2}^{(\tilde t-1)} \sigma(\la \wb_{y_k, r}^{(\tilde t-1)}, \bxi_k \ra) - v_{y_k, r, 2}^{(\tilde t-1)} \sigma(\la \wb_{y_k, r}^{(\tilde t-1)}, \bxi_k \ra) \cdot \eta \ell'^{(\tilde t-1)}_k (\frac{k_1 \|\bxi_k\|^2}{\sqrt{n}\sigma_p \sqrt{d}} + \frac{1}{C_7 k_2 m v_0} \cdot (|S_{y_k, r}^{(\tilde t-1)}|/n)).
    \end{align*}
    The inequality is from the definition of $k_1, k_2$ and $\ell'^{(t)}_k \le 0$. Sum up from $r=1$ to m, we have
    \begin{align*}
        \sum_{r=1}^m v_{y_k, r, 2}^{(\tilde t)} \sigma(\la \wb_{y_k, r}^{(\tilde t)}, \bxi_k \ra) &\ge \sum_{r=1}^m v_{y_k, r, 2}^{(\tilde t-1)} \sigma(\la \wb_{y_k, r}^{(\tilde t-1)}, \bxi_k \ra) - v_{y_k, r, 2}^{(\tilde t-1)} \sigma(\la \wb_{y_k, r}^{(\tilde t-1)}, \bxi_k \ra) \cdot \eta \ell'^{(\tilde t-1)}_k (\frac{k_1 \|\bxi_k\|^2}{\sqrt{n}\sigma_p \sqrt{d}} + \frac{1}{C_7 k_2 m v_0} \cdot (|S_{y_i, r}^{(\tilde t-1)}|/n))\\
        &\ge  \sum_{r=1}^m v_{y_k, r, 2}^{(\tilde t-1)} \sigma(\la \wb_{y_k, r}^{(\tilde t-1)}, \bxi_k \ra) + M v_{y_k, r, 2}^{(\tilde t-1)} \sigma(\la \wb_{y_k, r}^{(\tilde t-1)}, \bxi_k \ra) \cdot \exp(-\sum_{r=1}^m v_{y_k, r, 2}^{(\tilde t-1)} \sigma(\la \wb_{y_k, r}^{(\tilde t-1)}, \bxi_k \ra))\\ 
        &\ge \log(M(\tilde t-1)) + \frac{M \log(\tilde t - 1)}{\tilde t -1}\\
        &\ge \log(M \tilde t).
    \end{align*}
    In the second inequality we define $M = \frac{1}{2} e^{-0.25} \eta (\frac{k_1 \|\bxi_k\|^2}{\sqrt{n}\sigma_p \sqrt{d}} + \frac{1}{C_7 k_2 m v_0} \cdot (|S_{y_k, r}^{(\tilde t-1)}|/n))$; The third inequality is from our induction hypothesis $\sum_{r=1}^m v_{y_k, r, 2}^{(\tilde t-1)} \sigma(\la \wb_{y_k, r}^{(\tilde t-1)}, \bxi_k \ra) \ge \log(M(\tilde t-1))$ and $f(x) = \frac{x}{e^x}$ is monotonically decreasing when $x \ge 0$; The last inequality is because $f(x) = \frac{\log(x)}{x}$ is monotonically decreasing when $x \ge e$.
    
\end{proof}

With the lower bound of the output, we can prove that the training loss can be arbitrarily small as time grows.
\begin{lemma}
\label{lemma: small_train_loss_c1}
Under Condition \ref{condition: main condition}, there exists a time $t \le O(\frac{1}{M \epsilon})$ that 
\begin{align*}
    L_D(\Wb^{(t)}, \vb^{(t)}) \le \epsilon.
\end{align*}
Here $M$ is the same as the definition in Lemma \ref{lemma: output_lower_c1}.
\end{lemma}

\begin{proof}[Proof of Lemma \ref{lemma: small_train_loss_c1}]
    Combining the results of Lemma \ref{lemma: output_leading_term_c1} and Lemma \ref{lemma: output_lower_c1} we have
    \begin{align*}
        y_i f(\bW, \btheta; \bx_i) &\ge -0.25 + \sum_{r=1}^m v_{y_k, r, 2}^{(t)} \sigma(\la \wb_{y_k, r}^{(t)}, \bxi_k \ra)\\
        &\ge -0.25 + \log(M t).
    \end{align*}
    Therefore, we have
    \begin{align*}
        L_D(\Wb^{(t)}, \vb^{(t)}) &\le \log(1 + e^{0.25}/(M t)) \le \frac{e^{0.25}}{Mt}.
    \end{align*}
    The last inequality is from $\log(1 + x) \le x$ for $x \ge 0$. So when $t \ge \Omega(\frac{1}{M \epsilon})$, we have $L_D(\Wb^{(t)}, \vb^{(t)}) \le \epsilon$.
\end{proof}

\subsection{Test error analysis}
In this section, we estimate the test error and derive the last two conclusions in Theorem \ref{thm: single_phase}.
Assume a new data $(\bx, y), \bx = (\bx^{(1)}, \bx^{(2)})^{\top}$ where $\bx^{(1)} = y \cdot \bmu$. This is equivalent to estimating $\mathbb{P}(y \cdot f(\bW, \bV; \bx) > 0)$. The output of our trained model is
    \begin{align*}
        y \cdot f(\bW^{(t)}, \bV^{(t)}; \bx) = & \sum_{r=1}^{m} [v_{y,r,1}^{(t)} \sigma(\la \wb_{y,r}^{(t)}, y\bmu \ra) + v_{y,r,2}^{(t)}\sigma(\la \wb_{y,r}^{(t)}, \bxi \ra)]\\
        &- \sum_{r=1}^{m} [v_{-y,r,1}^{(t)} \sigma(\la \wb_{-y,r}^{(t)}, y\bmu \ra) + v_{-y,r,2}^{(t)}\sigma(\la \wb_{-y,r}^{(t)}, \bxi \ra)].
    \end{align*}

From previous analysis, when $t = \Omega\Big(\frac{n}{\eta \sigma_p^2 d m v_0^2}\Big)$, the output of signal part is lower bounded,
\begin{align}
\label{eq: signal_output_lower_c1}
    \sum_{r=1}^m v_{j,r,1}^{(t)} \la\wb_{j,r}^{(t)},y_k \bmu\ra = \Theta(t \eta m v_0^2 \|\bmu\|_2^2) \ge \Theta\Bigg(\frac{n \|\bmu\|_2^2 }{\sigma_p^2 d}\Bigg).
\end{align}

And from the update equation of $\la \wb_{j,r}^{(t)}, \bxi_k\ra$ we know that for all $j \neq y_k$
    \begin{align*}
        \la\wb_{j,r}^{(t+1)},y_k \bmu\ra &= \la\wb_{j,r}^{(t)},y_k \bmu\ra - j y_k \frac{\eta}{n} v_{j,r,1}^{(t)} \|\bmu\|_2^2 \sum_{i=1}^n\ell'^{(t)}_i \mathbb{I}(\la\wb_{j,r}^{(t)},y_i \cdot \bmu\ra >0)\\
        &= \la\wb_{j,r}^{(t)},y_k \bmu\ra + \frac{\eta}{n} v_{j,r,1}^{(t)} \|\bmu\|_2^2 \sum_{i=1}^n\ell'^{(t)}_i \mathbb{I}(\la\wb_{j,r}^{(t)},y_i \cdot \bmu\ra >0)\\
        &\le \la\wb_{j,r}^{(t)},y_k \bmu\ra \\
        & \le \la\wb_{j,r}^{(0)},y_k \bmu\ra
    \end{align*}
    The second equity is from Lemma \ref{lemma: positive_v} and Lemma \ref{lemma: key_lemma_stage2_c1} that the output layer is positive all the time. So $\la \wb_{-y, r}^{(t)}, y\bmu \ra$ will be non-increasing and its order is neglectable compared with $\la \wb_{y,r}^{(t)}, y\bmu \ra$ which is non-decreasing. Therefore, we have
    \begin{align*}
        \sum_{r=1}^{m} v_{-y,r,1}^{(t)} \sigma(\la \wb_{-y,r}^{(t)}, y\bmu \ra) \le \sum_{r=1}^{m} v_{y,r,1}^{(t)} \sigma(\la \wb_{y,r}^{(t)}, y\bmu \ra).
    \end{align*}

So we have
\begin{align}
\label{eq: output_lower_c1}
    y \cdot f(\bW^{(t)}, \bV^{(t)}; \bx) &\ge \sum_{r=1}^m v_{j,r,1}^{(t)} \la\wb_{j,r}^{(t)},y_k \bmu\ra - \sum_{r=1}^m v_{-y,r,2}^{(t)} \sigma(\la \wb_{-y,r}^{(t)}, \bxi \ra)\nonumber\\
    &\ge \frac{c n \|\bmu\|_2^2 }{\sigma_p^2 d} - \sum_{r=1}^m v_{-y,r,2}^{(t)} \sigma(\la \wb_{-y,r}^{(t)}, \bxi \ra).
\end{align}

Then we denote $g(\bxi) = \sum_{r=1}^{m} v_{-y,r,2}^{(t)} \sigma(\la \wb_{-y,r}^{(t)}, \bxi \ra)$. According to Theorem 5.2.2 in \cite{vershynin2018introduction}, we know that for any $x \ge 0$ it holds that
    \begin{equation}\label{eq: test_error_p}
        \mathbb{P}(g(\bxi) - \mathbb{E}[g(\bxi)] \ge x) \le \exp \Big(- \frac{cx^2}{\sigma_p^2\|g\|_{Lip}^2}\Big),
    \end{equation}
    where c is a constant.

We compute the expectation of $g(\bxi)$ as
\begin{align*}
    \mathbb{E}(\sum_{r=1}^m v_{y,r,2}^{(t)} \sigma(\la \wb_{y,r}^{(t)}, \bxi \ra)) &= \sum_{r=1}^{m} v_{-y,r,2}^{(t)} \mathbb{E}[\sigma(\la \wb_{-y,r}^{(t)}, \bxi \ra)]\\
        &= \sum_{r=1}^{m} v_{-y,r,2}^{(t)} \frac{\|\wb_{-y,r}^{(t)}\|_2 \sigma_p}{\sqrt{2\pi}}.
\end{align*}
And to calculate the Lipschitz norm, we have
    \begin{align*}
        |g(\bxi) - g(\bxi')| &= \Bigg|\sum_{r=1}^{m} v_{-y,r,2}^{(t)} \sigma(\la \wb_{-y,r}^{(t)}, \bxi \ra) - \sum_{r=1}^{m} v_{-y,r,2}^{(t)} \sigma(\la \wb_{-y,r}^{(t)}, \bxi' \ra)\Bigg|\\
        &\le \sum_{r=1}^m \Big|v_{-y,r,2}^{(t)} (\sigma(\la \wb_{-y,r}^{(t)}, \bxi \ra) - \sigma(\la \wb_{-y,r}^{(t)}, \bxi' \ra)) \Big|\\
        &\le \sum_{r=1}^m v_{-y,r,2}^{(t)} \Big|\la \wb_{-y,r}^{(t)}, \bxi - \bxi' \ra \Big|\\
        &\le \sum_{r=1}^m v_{-y,r,2}^{(t)} \|\wb^{(t)}_{-y,r}\|_2(\bxi - \bxi'),
    \end{align*}
    where the first inequality is by triangle inequality; the second inequality is by the property of ReLU and Lemma \ref{lemma: positive_v} that $v_{-y,r,2}^{(t)} > 0$; the last inequality is by Cauchy-Schwartz inequality. Therefore, we have
    \begin{equation}\label{eq: lip_norm}
        \|g\|_{Lip} \le \sum_{r=1}^m v_{-y,r,2}^{(t)} \|\wb^{(t)}_{-y,r}\|_2.
    \end{equation}   

Next we could upper bound $\|\wb_{j,r}^{(t)}\|$ as
    \begin{align*}
        \|\wb_{j,r}^{(t)}\| &\le  \|\wb_{j,r}^{(0)}\| + |\gamma_{j,r}^{(t)}\cdot \|\bmu\|_2^{-2} \cdot \bmu|_2 + \bigg|\sum_{i=1}^n \rho_{j,r,i}^{(t)} \cdot \|\bxi_i\|_2^{-2} \cdot \bxi_i \bigg|\\ 
        &\le \|\wb_{j,r}^{(0)}\| + |\gamma_{j,r}^{(t)}\cdot \|\bmu\|_2^{-2} \cdot \bmu| + \frac{1}{\sqrt{n} \sigma_p \sqrt{d}} \cdot \bigg|\sum_{k=1}^n \rho_{j,r,k}^{(t)}\bigg|\\        
        &\le \frac{3}{2} \sigma_0^2 d + \tilde \Theta \bigg(\frac{n \|\bmu\|_2}{m^2 v_0^2 \sigma_p^2 d }\bigg) + \tilde \Theta \bigg(\frac{\sqrt{n}}{m v_0 \sigma_p \sqrt{d}} \bigg)\\
        &= \tilde \Theta \bigg(\frac{\sqrt{n}}{m v_0 \sigma_p \sqrt{d}} \bigg),
    \end{align*}
    where the first inequality is from triangle inequality and the next two inequalities are from our estimate of parameters $\gamma$ and $\rho$ in Proposition \ref{prop: stage2_analysis_c1}.

Meanwhile, from Proposition \ref{prop: stage2_analysis_c1} the output layer $v_{j,r,2}^{(t)} = \tilde \Theta(v_0)$.
 So we have
    \begin{align}
    \label{eq: expect_upper_c1}
        \mathbb{E}(\sum_{r=1}^m v_{y,r,2}^{(t)} \sigma(\la \wb_{y,r}^{(t)}, \bxi \ra)) \le \tilde \Theta\bigg(\sqrt{\frac{n}{d}}\bigg).
    \end{align}

     Now using the method in (\ref{eq: test_error_p}) with results above, we can obtain
    \begin{align*}
        \mathbb{P}_{(\bx, y) \sim \textit{D}}(y \cdot f(\bW^{(t)}, \bV^{(t)}; \bx) < 0) &\le \mathbb{P}_{(\bx, y) \sim \textit{D}}\bigg(\sum_{r=1}^m  v_{-y,r,2}^{(t)}\sigma(\la \wb_{-y,r}^{(t)}, \bxi \ra) > \sum_{r=1}^{m} v_{y,r,1}^{(t)} \sigma(\la \wb_{y,r}^{(t)}, y\bmu \ra)\bigg)\\
        =& \mathbb{P}_{(\bx, y) \sim \textit{D}}\bigg(g(\bxi) - \mathbb{E}[g(\bxi)] > \sum_{r=1}^{m} v_{y,r,1}^{(t)} \sigma(\la \wb_{y,r}^{(t)}, y\bmu \ra)\\
        & - \sum_{r=1}^{m} v_{-y,r,2}^{(t)} \frac{\|\wb_{-y,r}^{(t)}\|_2 \sigma_p}{\sqrt{2\pi}}\bigg)\\
        \le& \exp{\Bigg[-\frac{c (\sum_{r=1}^{m} v_{y,r,1}^{(t)} \sigma(\la \wb_{y,r}^{(t)}, y\bmu \ra - \sum_{r=1}^{m} v_{-y,r,2}^{(t)} \frac{\|\wb_{-y,r}^{(t)}\|_2 \sigma_p}{\sqrt{2\pi}})^2}{\sigma_p^2 (c_1 \sum_{r=1}^m v_{-y,r,2}^{(t)} \|\wb^{(t)}_{-y,r}\|_2)^2}\bigg]}\\
        =& \exp{\Bigg[-c\bigg(\frac{\sum_{r=1}^{m} v_{y,r,1}^{(t)} \sigma(\la \wb_{y,r}^{(t)}, y\bmu \ra)}{c_1 \sigma_p \sum_{r=1}^m v_{-y,r,2}^{(t)} \|\wb^{(t)}_{-y,r}\|_2}-\frac{1}{\sqrt{2\pi}c_1}\bigg)^2\Bigg]}\\
        \le& \exp{(1/c_1^2)}\exp{\Bigg(-0.5c\bigg(\frac{\sum_{r=1}^{m} v_{y,r,1}^{(t)} \sigma(\la \wb_{y,r}^{(t)}, y\bmu \ra)}{c_1 \sigma_p \sum_{r=1}^m v_{-y,r,2}^{(t)} \|\wb^{(t)}_{-y,r}\|_2}\bigg)^2\Bigg)}\\
        \le&\exp{\bigg(\frac{1}{c_1^2} - \frac{\tilde \Theta(\frac{n \|\bmu\|_2^2 }{\sigma_p^2 d})^2}{\tilde \Theta(\sqrt{\frac{n}{d}})^2} \bigg)}\\
        =& \exp(- c_2 n\|\bmu\|_2^4 \sigma_p^{-4} d^{-1}))\\
        =& o(1).
    \end{align*}
    Here $c_1, c_2 = \tilde \Theta(1)$. The first inequality is from (\ref{eq: output_lower_c1}); The second inequality is from (\ref{eq: test_error_p}); The third inequality is from the fact that $(s-t)^2 \ge s^2/2 - t^2$ for all $s, t \ge 0$; The last inequality is from (\ref{eq: signal_output_lower_c1}) and (\ref{eq: expect_upper_c1}). This completes the proof for the second conclusion in Theorem \ref{thm: single_phase}.

    Then we compute the test error lower bound. This is equal to provide a lower bound for $\mathbb{P}_{(\bx, y) \sim \textit{D}}\big(y \cdot f(\bW^{(t)}, \bV^{(t)}; \bx) < 0\big)$. To estimate this, we first introduce an important lemma:
    
    \begin{lemma}
\label{lemma: large_noise_change_c1}
    When $t = T_1$, denote $g(\bxi) = \sum_{j,r}jv_{j,r,2}^{(t)} \sigma(\la \wb_{j,r}^{(t)}, \bxi \ra )$. There exists a fixed vector $\btheta$ with $\|\btheta\|_2 \le 0.04 \sigma_p$ such that,
    \begin{align*}
        \sum_{j'\in \{\pm 1\}} [g(j'\bxi + \btheta) - g(j'\bxi)] \ge 4 c_3 \max \limits_{j'\in \{\pm 1\}} \Big\{\sum_r v_{j,r,1}^{(t)} \sigma(\la \wb_{j,r}^{(t)}, j\bmu \ra)\Big\},
    \end{align*}
    for all $\bxi \in \mathbb{R}^d$.
\end{lemma}

\begin{proof}[Proof of Lemma \ref{lemma: large_noise_change_c1}]
    Without loss of generality, let $\max \limits_{j'\in \{\pm 1\}} \{\sum_r v_{j,r,1}^{(t)} \sigma(\la \wb_{j,r}^{(t)}, j\bmu \ra)\} = \sum_r v_{1,r,1}^{(t)} \sigma(\la \wb_{1,r}^{(t)}, \bmu \ra)$.  We construct $\btheta$ as
    \begin{align*}
        \btheta = \lambda \cdot \sum_i \mathbb{I}(y_i = 1) \bxi_i,
    \end{align*}
    where $\lambda = \frac{c_4 \|\bmu\|_2^2}{\sigma_p^2 d}$ and $c_4$ is a sufficiently large constant. Then we have
    \begin{align*}
        &\sigma(\la \wb_{1,r}^{(t)}, \bxi+\btheta \ra) - \sigma(\la \wb_{1,r}^{(t)}, \bxi \ra) + \sigma(\la \wb_{1,r}^{(t)}, -\bxi+\btheta \ra) - \sigma(\la \wb_{1,r}^{(t)}, -\bxi \ra)\\
        &\ge \sigma'(\la \wb_{1,r}^{(t)}, \bxi \ra) \la \wb_{1,r}^{(t)}, \btheta \ra + \sigma'(\la \wb_{1,r}^{(t)}, -\bxi \ra) \la \wb_{1,r}^{(t)}, \btheta \ra\\
        &= \la \wb_{1,r}^{(t)}, \btheta \ra\\
        &\ge \lambda \cdot \sum_{y_i=1} \la \wb_{1,r}^{(t)}, \bxi_i \ra
    \end{align*}
    The first inequity is because ReLU is a Liptchitz.
    Sum up all neurons, we have
    \begin{align*}
        &g(\bxi+\btheta) - g(\bxi) + g(-\bxi + \btheta) - g(-\bxi)\\
        &\ge \lambda \bigg[\sum_{r=1}^m v_{j,r,2}^{(t)} \sum_{y_i=1} \sigma(\la \wb_{j,r,2}^{(t)}, \bxi_i \ra )\bigg]\\
        &\ge \lambda \sum_{r=1}^m \frac{\sum_{y_i=1} v_{j,r,2}^{(t)} \sigma(\la \wb_{j,r,2}^{(t)}, \bxi_i \ra)}{v_{j,r,1}^{(t)} \sigma(\la \wb_{j,r}^{(t)}, \bmu \ra)} \cdot (v_{j,r,1}^{(t)} \sigma(\la \wb_{j,r}^{(t)}, \bmu \ra))\\
        &\ge \lambda \cdot \sum_{r=1}^m \tilde \Theta(\frac{n/m}{n\|\bmu\|_2^2 m^{-1}\sigma_p^{-2}d^{-1}}) \cdot (v_{j,r,1}^{(t)} \sigma(\la \wb_{j,r}^{(t)}, \bmu \ra))\\
         &\ge \lambda \cdot \tilde \Theta\Bigg(\frac{\sigma_p^2 d}{\|\bmu\|_2^2} \Bigg) \cdot \sum_{r=1}^m v_{j,r,1}^{(t)} \sigma(\la \wb_{j,r}^{(t)}, \bmu \ra)\\
        &\ge 4 c_3 \cdot \sum_{r=1}^m v_{j,r,1}^{(t)} \sigma(\la \wb_{j,r}^{(t)}, \bmu \ra).
    \end{align*}
    The third inequity is from Proposition \ref{prop: stage2_analysis_c1} which estimates the outputs of signal and noise, and the last inequity is due to our definition of $\lambda$ and from Condition \ref{condition: main condition} that $\sigma_p \sqrt{d} \ge O(n/m)$. 
    
    Finally, the norm of $\btheta$ is small that
    \begin{align*}
        \|\btheta\|_2 = \bigg\|\lambda \cdot \sum_{y_i=1}\bxi_i \bigg\|_2 = \Theta(\sqrt{n} \|\bmu\|_2^2 \sigma_p^{-1} \sqrt{d}^{-1}) \le 0.04 \sigma_p,
    \end{align*}
    where the last inequity is by condition $n \|\bmu\|_2^4 = O(\sigma_p^4 d)$.
\end{proof}
    
    Then we can prove the last conclusion in Theorem \ref{thm: single_phase}. We can lower bound the test error as
    \begin{align*}
        &\mathbb{P}_{(\bx, y) \sim \textit{D}}\big(y \cdot f(\bW^{(t)}, \bV^{(t)}; \bx) < 0\big)\\
        &= \mathbb{P}_{(\bx, y) \sim \textit{D}}\bigg(\sum_r v_{-y,r,2}^{(t)} \sigma(\la \wb_{-y,r}^{(t)}, \bxi \ra) - \sum_r v_{y,r,2}^{(t)} \sigma(\la \wb_{y,r}^{(t)}, \bxi \ra) 
        \ge \sum_r v_{y,r,1}^{(t)} \sigma(\la \wb_{j,r}^{(t)}, y \bmu \ra) - \sum_r v_{-y,r,1}^{(t)} \sigma(\la \wb_{-y,r}^{(t)}, y \bmu \ra)\bigg)\\
        &\ge 0.5 \mathbb{P}_{(\bx, y) \sim \textit{D}}\bigg(\bigg|\sum_r v_{-y,r,2}^{(t)} \sigma(\la \wb_{-y,r}^{(t)}, \bxi \ra) - \sum_r v_{y,r,2}^{(t)} \sigma(\la \wb_{y,r}^{(t)}, \bxi \ra)\bigg| \ge c_4 \max \limits_{j'\in \{\pm 1\}} \Big\{\sum_r v_{j,r,1}^{(t)} \sigma(\la \wb_{j,r}^{(t)}, j\bmu \ra)\Big\}\bigg).
    \end{align*}
     If the value of left side is too large, the randomness of noise will have a greater effect on the invariant signal part, which will lead to wrong prediction. Let $g(\bxi) = \sum_r v_{-y,r,2}^{(t)} \sigma(\la \wb_{-y,r}^{(t)}, \bxi \ra) - \sum_r v_{y,r,2}^{(t)} \sigma(\la \wb_{y,r}^{(t)}, \bxi \ra)$ and the set
     \begin{align*}
         \Omega:=\Big\{\bxi \Big| |g(\bxi)| \ge c_4 \max \limits_{j'\in \{\pm 1\}} \Big\{\sum_r v_{j,r,1}^{(t)} \sigma(\la \wb_{j,r}^{(t)}, j\bmu \ra)\Big\}\Big\}.
     \end{align*}
     So we have
     \begin{align*}
          \mathbb{P}_{(\bx, y) \sim \textit{D}}\big(&y \cdot f(\bW^{(t)}, \bV^{(t)}; \bx) < 0\big) \ge 0.5 \mathbb{P}(\Omega).
     \end{align*}
     Then our goal is to lower bound $\mathbb{P}(\Omega)$. From Lemma \ref{lemma: large_noise_change_c1}, we know that $\sum_{j'\in \{\pm 1\}} [g(j'\bxi + \btheta) - g(j'\bxi)] \ge 4 c_4 \max \limits_{j'\in \{\pm 1\}} \{\sum_r v_{j,r,1}^{(t)} \sigma(\la \wb_{j,r}^{(t)}, j\bmu \ra)\}$. Therefore, by pigeon’s hole principle, there must exist one of the $\bxi, \bxi+\btheta, -\bxi, -\bxi+\btheta$ belongs $\Omega$. So we have 
     \begin{align*}
         \Omega \cup -\Omega \cup \Omega - \{\btheta\} \cup -\Omega - \{\btheta\} = \mathbb{R}^d.
     \end{align*}
     This indicates that at least one of $\mathbb{P}(\Omega), \mathbb{P}(-\Omega), \mathbb{P}(\Omega-{\btheta}), \mathbb{P}(-\Omega-{\btheta})$ is greater than 0.25. Also note that $\|\btheta\|_2 \le 0.04 \sigma_p$, so we have
     \begin{align*}
         \big|\mathbb{P}(\bxi) - \mathbb{P}(\bxi-\btheta)\big| &= \big|\mathbb{P}_{\bxi \in \textit{N}(0,\sigma_p^2 I_d)}(\bxi \in \Omega) - \mathbb{P}_{\bxi \in \textit{N}(\btheta,\sigma_p^2 I_d)}(\bxi \in \Omega)\big|\\
         &\le TV(\textit{N}(0,\sigma_p^2 I_d), \textit{N}(\btheta,\sigma_p^2 I_d))\\
         &\le \frac{\|\btheta\|_2}{\sigma_p}\\
         &\le 0.02
     \end{align*}
     Here the first inequality is by Proposition 2.1 in \citet{devroye2018total} and the second inequality is from Lemma \ref{lemma: large_noise_change_c1} that $\|\btheta\|_2 \le 0.04 \sigma_p$. Meanwhile, we have $\mathbb{P}(\Omega) = \mathbb{P}(-\Omega)$. Combine with the results above, we can conclude that $\mathbb{P}(\Omega) \ge 0.23$.

     Therefore, we can lower bound the wrong prediction probability
     \begin{align*}
         \mathbb{P}_{(\bx, y) \sim \textit{D}}(&y \cdot f(\bW^{(t)}, \bV^{(t)}; \bx) < 0)\\
         &\ge 0.5 \mathbb{P}(\Omega)\\
         &\ge 0.1.
     \end{align*}
This completes the proof of Theorem \ref{thm: single_phase}.

\section{Proof of Small Initialization}
Next we consider the case when the output layer initialization $v_0  \le \Theta(n^{1/4}\sigma_p^{-1/2}d^{-1/4}m^{-1/2})$. In this case, as the output layer has a relatively small initialization, it will simultaneously increase with the first layer so the dynamic is more complicated. Same as above, we decouple the whole process into two stages.

\subsection{First stage}
Same as above, we define stage 1 as the training period $0 \le t \le T^{*,1}$, and $T^{*,1}$ is
\begin{equation}
\label{eq: define_T1_s}
    T^{*,1}= \max\limits_{k,j,r}\bigg\{t: \sum_{r=1}^m v^{(t)}_{j,r,2}\cdot\sigma(\la\wb_{j,r}^{(t)},\bxi_k\ra) \le 0.1\bigg\}.
\end{equation}

We present the main conclusion in this stage first:

\begin{proposition}
\label{prop: stage1_s}
 Under Condition~\ref{condition: main condition}, with probability at least $1-\delta$ there exists $T_1 = \tilde O \Big(\frac{\sqrt{n}}{\eta \sigma_p \sqrt{d}}\Big)$ such that for every sample $k \in [n]$ with label $y_k = j$:
 \begin{enumerate}
     \item Noise memorization reaches constant level: $v^{(T_1)}_{y_k,r,2}\cdot\sigma(\la\wb_{y_k,r}^{(T_1)},\bxi_k\ra)= 0.05/m$.
     \item Two layers of noise part converge: $\la\wb_{y_k,r}^{(T_1)},\bxi_k\ra / v^{(T_1)}_{y_k,r,2} = \Theta(\sigma_p \sqrt{d} / \sqrt{n})$ .
     \item Signal learning reaches level that: $v^{(T_1)}_{y_k,r,1}\cdot\sigma(\la\wb_{y_k,r}^{(T_1)},y_k \bmu\ra)=\tilde \Theta\Big(\frac{\sqrt{n} \|\bmu\|_2^2 v_0^2}{\sigma_p \sqrt{d}}\Big)$.
     \item Two layers of signal part have not converged: $ \la\wb_{y_k,r}^{(T_1)},y_k \bmu\ra / v^{(T_1)}_{y_k,r,1} = o(\|\bmu\|_2)$.
 \end{enumerate}
\end{proposition}

Different from the previous section, we use a Two-Phase Analysis to better characterize the dynamic in Stage 1.

\subsubsection{Linear phase}
    We define the first phase as all iterations $0 \le t \le T_0$, where $T_0 = \Theta\Big(\frac{\sqrt{n}}{\eta} \sigma_p\sqrt{d}\Big)$. The selection of $T_0$ is explained in the following section. We present the main conclusion in this phase:

    \begin{lemma}[Noise Memorization becomes balancing]
\label{lemma: noise_balanced}
Under Condition \ref{condition: main condition}, if $|\ell'^{(t)}_i|\in [0.4, 0.6]$, there exists a time $T_0 = \Theta(\frac{\sqrt{n}}{\eta \sigma_p \sqrt{d}}) \le T^{*,1}$ that for all $y_k=j$:
\begin{itemize}
\item $v^{(T_0)}_{j,r,2}/\la\wb_{j,r}^{(T_0)},\bxi_k\ra=\Theta(\sqrt{n}/\sigma_p \sqrt{d})$;
\item $v^{(T_0)}_{j,r,2}\cdot\sigma(\la\wb_{j,r}^{(T_0)},\bxi_k\ra)=\Theta\Big(\frac{\sigma_p \sqrt{d} \cdot v_0^2}{\sqrt{n}}\Big)$;
\item $v^{(T_0)}_{j,r,1}\cdot\sigma(\la\wb_{j,r}^{(T_0)},y_k\bmu\ra)=\Theta\Big(\frac{\sqrt{n} \|\bmu\|_2^2 v_0^2}{\sigma_p \sqrt{d}}\Big)$.
\end{itemize}
\end{lemma}
To prove the main conclusion we also need several technical lemmas. First as the proof of Lemma \ref{lemma: positive_v} and Lemma \ref{lemma: inner_product_balance} have no constraint on the output layer initialization, these two lemmas still hold in this case.

Next we can show that the neurons activated at initialization will remain active throughout the process
\begin{lemma}
\label{lemma: active neuron}
Define $S_i^{(t)} := \{r \in [m]: \la w_{y_i,r},\bxi_k \ra >0\}$, and $S_{j,r}^{(t)}:=\{k\in [n]:y_k=j,\la w_{j,r}^{(t)}, \bxi_k \ra > 0\}$. For all $k \in [n], r \in [m]$ and $j \in \{\pm 1\}$, $S_k^{(0)} \subseteq S_k^{(t)}, S_{j,r}^{(0)} \subseteq S_{j,r}^{(t)}$
\end{lemma}

\begin{proof}[Proof of Lemma \ref{lemma: active neuron}]
First we prove that $S_k^{(0)} \subseteq S_k^{(t)}$, and it is similar for $S_{j,r}^{(t)}$. Recall the update equation of $\la w_{j,r}^{(t)}, \bxi_k \ra$
\begin{align*}
\la\wb_{j,r}^{(t)},\bxi_k\ra &= \la\wb_{j,r}^{( t-1)},\bxi_k\ra - \frac{j y_k \eta}{n} v_{j,r,2}^{( t-1)} \|\bxi_k\|_2^2 {\ell'}_{k}^{( t-1)} -\frac{j \eta}{n} v_{j,r,2}^{( t-1)} \sum_{i=1,i\neq k}^n y_i {\ell'}_{i}^{( t-1)} \la \bxi_k,\bxi_i \ra \mathbb{I}(\la\wb_{j,r}^{( t-1)},\bxi_i\ra >0)\\ 
&\ge \la\wb_{j,r}^{( t-1)},\bxi_k\ra + \frac{\eta}{n} v_{j,r,2}^{( t-1)} \cdot 0.4\cdot \sigma_p^2d/2 - \frac{\eta}{n} v_{j,r,2}^{( t-1)} \cdot 0.6 n \cdot 2 \sigma_p^2 \cdot \sqrt{d \log(4n^2/\delta)}\\
&\ge \la\wb_{j,r}^{( t-1)},\bxi_k\ra \ge 0,
\end{align*}
where the first inequity is from Lemma \ref{lemma: noise bound} and $|\ell'^{(t)}_i \in [0.4, 0.6]$ for all $i \in [n]$. The second inequity is due to $d=\Omega(n^2\log(4n^2/\delta))$.
So this indicates that
\begin{align*}
|S_i^{(0)}| \subseteq |S_i^{( t-1)}| \subseteq |S_i^{(t)}|.
\end{align*}
\end{proof}

Then we can prove for the main conclusion in Linear Phase.
\begin{proof}[Proof of Lemma \ref{lemma: noise_balanced}]
We prove for the noise part first. Recall the update of $\la\wb_{j,r}^{(t)},\bxi_k\ra$ and $v_{j,r,2}^{(t)}$:
\begin{align*}
 \la\wb_{j,r}^{(t+1)},\bxi_k\ra &= \la\wb_{j,r}^{(t)},\bxi_k\ra - \frac{j y_k \eta}{n} v_{j,r,2}^{(t)} \|\bxi_k\|_2^2 \ell'^{(t)}_{k} -\frac{j \eta}{n} v_{j,r,2}^{(t)} \sum_{i=1,i\neq k}^n y_i \ell'^{(t)}_i \la \bxi_k,\bxi_i \ra \mathbb{I}(\la\wb_{j,r}^{(t)},\bxi_i\ra >0),\\ 
    v_{j,r,2}^{(t+1)} & = v_{j,r,2}^{(t)} - \frac{j \eta}{n}\sum_{i=1}^{n}\ell'^{(t)}_iy_i\la \wb_{j,r}^{(t)},\bxi_i \ra \mathbb{I}(\la\wb_{j,r}^{(t)},\bxi_i\ra >0).
\end{align*}
From lemma \ref{lemma: inner_product_balance}, we know that  for all $y_i = y_k = j$, $\frac{\la\wb_{j,r}^{(t)},\bxi_i\ra}{\la\wb_{j,r}^{(t)},\bxi_k\ra} \ge \frac{1}{9}$. Meanwhile, from Lemma \ref{lemma: preliminary_S_jr_intialization} and Lemma \ref{lemma: active neuron}, the samples activate the specific neuron will remain active later, so $\sum_{i=1}^{n} \mathbb{I}(\la\wb_{j,r}^{(t)},\bxi_i\ra >0) \ge n/8$.

Similar to the previous case, we use the technical lemma in Section \ref{sec: tech_lemmas} to analyze the dynamic of two layers. We restate the lemma for convenience.

\lemmaseq*

Take $a_0, b_0, A, B$ as $\la\wb_{j,r}^{(0)},\bxi_k\ra, v_0, 0.4 \frac{\eta}{n}  \|\bxi_k\|_2^2, \eta/180 $, from Condition \ref{condition: main condition} we know that these parameters satisfy the condition of Lemma \ref{lemma: interwined_sequence}. So we could get the time when the noise part becomes balanced:
\begin{align*}
&T_0 = \Theta(1 / \sqrt{AB}) = \frac{3\sqrt{20n}}{\eta \|\bxi_k\|} = \Theta\bigg(\frac{\sqrt{n}}{\eta \sigma_p \sqrt{d}}\bigg).
\end{align*}
Meanwhile, from lemma \ref{lemma: interwined_sequence} we also know that $v^{(T_0)}_{j,r,2} = \Theta(v_0)$,
 $\la\wb_{j,r}^{(T_0)},\bxi_k\ra = \Theta(\sqrt{A/B} \cdot v_0)$, and we could compute the output of noise part at time $T_0$:
\begin{align*}
&v^{(T_0)}_{j,r,2}\cdot\sigma(\la\wb_{j,r}^{(T_0)},\bxi_k\ra) = \Theta(\sqrt{A/B} \cdot b_0 ^2) = \Theta\Bigg(\frac{\|\bxi_k\| \cdot v_0^2}{\sqrt{n}}\Bigg) = \Theta\Bigg(\frac{\sigma_p \sqrt{d} \cdot v_0^2}{\sqrt{n}}\Bigg) \le 0.1/m,
\end{align*}
 where the last inequity is due to $v_0=O(n^{1/4} \sigma_p^{-1/2} d^{-1/4} m^{-1/2})$. So $T_0 \le T^{*,1}$. This completes the proof for the first two statements.

Last we consider the signal part within time $T_0$. Recall the update of $\la\wb_{j,r}^{(t)},y_k \bmu\ra$ and $v_{j,r,1}^{(t)}$:
\begin{align*}
\la\wb_{j,r}^{(t+1)},y_k \bmu\ra &= \la\wb_{j,r}^{(t)},y_k \bmu\ra - j y_k \frac{\eta}{n} v_{j,r,1}^{(t)} \|\bmu\|_2^2 \sum_{i=1}^n\ell'^{(t)}_i \mathbb{I}(\la\wb_{j,r}^{(t)},y_i \cdot \bmu\ra >0),\\
v_{j,r,1}^{(t+1)} & = v_{j,r,1}^{(t)} - j \frac{\eta}{n}\sum_{i=1}^{n}\ell'^{(t)}_i \la \wb_{j,r}^{(t)},\bmu \ra \mathbb{I}(\la\wb_{j,r}^{(t)},y_i \cdot \bmu\ra >0).
\end{align*}

We use the other technical lemma:

\lemmacompare*

We use lemma \ref{lemma: compare_sequence} by taking $\overline{a_0}=\la\wb_{j,r}^{(0)},\bxi_k\ra, \overline{b_0} = v_0$, $\overline{A} = 0.4 \frac{\eta}{n}  \|\bxi_k\|_2^2, \overline{B} = \eta/180$ and $\underline{a_0}=\la\wb_{j,r}^{(0)},\bmu\ra, \underline{b_0} = v_0$, $\underline{A} = 0.3 \eta  \|\bmu\|_2^2$, $\underline{B} = 0.3 \eta$. From the Condition \ref{condition: main condition}, these parameters satisfy the condition of lemma \ref{lemma: compare_sequence}. So we could get that there exists a constant c that $v_{j,r,1}^{(t)} \le \underline{b_t} = O(\overline{b_t}) \le O(v_{j,r,2}^{(t)}) = c \cdot v_0$. 

Meanwhile, It is obvious that $\la\wb_{j,r}^{(t)},y_k \bmu\ra \le \underline{a_t}$, $v_{j,r,1}^{(t)} \le \underline{b_t}$ and $\la\wb_{j,r}^{(t)},\bxi_k\ra \ge \overline{a_t}, v_{j,r,2}^{(t)} \ge \overline{b_t}$. So we have:
\begin{align*}
\la\wb_{j,r}^{(t)},y_k \bmu\ra &\le \underline{a_t} \le \underline{a_0} + t\cdot \underline{A} \cdot c \cdot v_0,\\
\la\wb_{j,r}^{(t)},y_k \bmu\ra &\ge \underline{a_0} + t/2 \cdot \underline{A} \cdot v_0,
\end{align*}
which implies that $\la\wb_{j,r}^{(t)},y_k \bmu\ra = \Theta(t \eta b_0 \|\bmu\|_2^2)$ and $v_{j,r,1}^{(t_1)} = \Theta(b_0)$. So by the time $T_0$ the balancing factor:
\begin{align*}
\la\wb_{j,r}^{(T_0)},y_k \bmu\ra / v_{j,r,1}^{(T_0)} = \eta T_0 \|\bmu\|_2^2 = \Theta\bigg(\frac{\sqrt{n}\|\bmu\|_2^2}{\sigma_p \sqrt{d}}\bigg) = o(\|\bmu\|_2).
\end{align*}
This suggests that the signal part is not balanced. Moreover, we could compute the signal learning at time $T_0$ that:
\begin{align*}
v_{j,r,1}^{(T_0)} \la\wb_{j,r}^{(T_0)},y_k \bmu\ra = \Theta(T_0 \eta b_0^2 \|\bmu\|_2^2) = \Theta\Bigg(\frac{\sqrt{n} \|\bmu\|_2^2 v_0^2}{\sigma_p \sqrt{d}}\Bigg).
\end{align*}
This completes the proof of the last statement.
\end{proof}

\subsubsection{Quadratic phase}

At the end of Linear Phase, we know that the two layers of noise part become balanced, so the growth rate of noise memorization will become more quadratic-like. We define \textbf{Phase 2} as all iterations $T_0 \le t \le T_1$, where $T_1 = \tilde O\Big(\frac{\sqrt{n}}{\eta \sigma_p \sqrt{d}}\Big)$.

\begin{lemma}[Noise Memorization becomes nearly constant]
\label{lemma: noise_output_constant}
Under Condition \ref{condition: main condition}, if $|\ell'^{(t)}_i|\in [0.4, 0.6]$, there exists a time $T_1$ that $T_0 < T_1= \tilde O\Big(\frac{\sqrt{n}}{\eta \sigma_p \sqrt{d}}\Big) \le T^{*,1}$, and for all $y_k=j$ we have:
\begin{itemize}
\item $v^{(T_1)}_{j,r,2}\cdot\sigma(\la\wb_{j,r}^{(T_1)},\bxi_k\ra)= 0.05/m$;
\item 
$\la\wb_{j,r}^{(T_1)},y_k \bmu\ra / v^{(T_1)}_{j,r,1} = o(\|\bmu\|_2)$;
\item $v^{(T_1)}_{j,r,1}\cdot\sigma(\la\wb_{j,r}^{(T_1)},y_k \bmu\ra)=\tilde O\Big(\frac{\sqrt{n} \|\bmu\|_2^2 v_0^2}{\sigma_p \sqrt{d}}\Big)$.
\end{itemize}
\end{lemma}

\begin{proof}[Proof of Lemma \ref{lemma: noise_output_constant}]
We consider the growth of the noise part. From lemma \ref{lemma: noise_balanced}, we know that the noise part reaches balancing stage (i.e. $v^{(t)}_{j,r,2}/\la\wb_{j,r}^{(t)},\bxi_k\ra=\Theta(\sqrt{n}/\|\bxi_k\|_2)$) by the time $T_0$ and will remain balanced after. So for every $T_0 \le t \le T^{*,1}$, we use lemma \ref{lemma: interwined_sequence} again and take $a_0, b_0, A, B$ as $\la\wb_{j,r}^{(0)},\bxi_k\ra, v_0, 0.4 \frac{\eta}{n}  \|\bxi_k\|_2^2, \eta/180 $, we have
\begin{align*}
a_{t+1} &= a_t + A \cdot b_t\\
&= a_t + A \cdot \Theta(\sqrt{n}/\|\bxi_k\|_2) \cdot a_t\\
&= a_t + \Theta\bigg(0.4 \frac{\eta}{n}  \|\bxi_k\|_2^2  \frac{\sqrt{n}}{\|\bxi_k\|}\bigg) \cdot a_t\\
&= \bigg(1 + \Theta\bigg(\frac{\eta \sigma_p \sqrt{d}}{\sqrt{n}}\bigg)\bigg) \cdot a_t,\\
b_{t+1} &= b_t + B \cdot a_t\\
&= b_t + B \cdot \Theta(\|\bxi_k\|/\sqrt{n}) \cdot b_t\\
&= b_t + \Theta(0.4 k_1 k_3 \eta \|\bxi_k\|/\sqrt{n}) \cdot b_t\\
&= \bigg(1+ \Theta\bigg(\frac{\eta \sigma_p \sqrt{d}}{\sqrt{n}}\bigg)\bigg) \cdot b_t.
\end{align*}
Combining the two results above, we could derive the output of noise at $t \ge T_0$
\begin{align*}
v^{(t)}_{j,r,2}\cdot\sigma(\la\wb_{j,r}^{(t)},\bxi_k\ra) &= \bigg(1+ \Theta\bigg(\frac{\eta \sigma_p \sqrt{d}}{\sqrt{n}}\bigg)\bigg)^2 v^{(t-1)}_{j,r,2}\cdot\sigma(\la\wb_{j,r}^{(t-1)},\bxi_k\ra)\\
&= \bigg(1+ \Theta\bigg(\frac{\eta \sigma_p \sqrt{d}}{\sqrt{n}}\bigg)\bigg)^{2(t-T_0)} v^{(T_0)}_{j,r,2}\cdot\sigma(\la\wb_{j,r}^{(T_0)},\bxi_k\ra)\\
&= \bigg(1+ \Theta\bigg(\frac{\eta \sigma_p \sqrt{d}}{\sqrt{n}}\bigg)\bigg)^{2(t-T_0)} \Theta\Bigg(\frac{\sigma_p \sqrt{d} \cdot v_0^2}{\sqrt{n}}\Bigg)\\
&\ge \exp\bigg(\Theta \bigg(\frac{\eta \sigma_p \sqrt{d}}{\sqrt{n}} \bigg)(t-T_0) \bigg) \cdot \Theta \Bigg(\frac{\sigma_p \sqrt{d} \cdot v_0^2}{\sqrt{n}}\Bigg),
\end{align*}
where the last inequity is due to $1+x > e^{x/2}$ for $x < 1$ and $\eta=O\Big(\frac{\sqrt{n}}{\sigma_p \sqrt{d}}\Big)$. On the other hand, we could upper bound the output that
\begin{align*}
v^{(t)}_{j,r,2}\cdot\sigma(\la\wb_{j,r}^{(t)},\bxi_k\ra)
&= \bigg(1+ \Theta\bigg(\frac{\eta \sigma_p \sqrt{d}}{\sqrt{n}}\bigg)\bigg)^{2(t-T_0)} \Theta\Bigg(\frac{\sigma_p \sqrt{d} \cdot v_0^2}{\sqrt{n}}\Bigg)\\
&\le \exp\bigg(\Theta\bigg(\frac{\eta \sigma_p \sqrt{d}}{\sqrt{n}}\bigg) 2(t-T_0)\bigg) \cdot \Theta\Bigg(\frac{\sigma_p \sqrt{d} \cdot v_0^2}{\sqrt{n}}\Bigg),
\end{align*}
where the inequity is due to $1+x < e^x$ for $x > 0$. So we could estimate the output of noise part as $\exp\Big(\Theta\Big(\frac{\eta \sigma_p \sqrt{d}}{\sqrt{n}}\Big)(t-T_0)\Big) \cdot \Theta\Big(\frac{\sigma_p \sqrt{d} \cdot v_0^2}{\sqrt{n}}\Big)$. When $t= T_1 = \Theta\Bigg(\frac{\sqrt{n} \cdot \log(\frac{\sqrt{n}}{\sigma_p \sqrt{d} v_0^2})}{\eta m \sigma_p \sqrt{d}}\Bigg) + T_0 = \tilde O\Big(\frac{\sqrt{n}}{\eta \sigma_p \sqrt{d}}\Big)$, the output of noise is $v^{(T_1)}_{j,r,2}\cdot\sigma(\la\wb_{j,r}^{(T_1)},\bxi_k\ra)= 0.05/m$. Meanwhile, as $0.05/m \le 0.1/m$, we also have $T_1 \le T^{*,1}$.\\
\\
Last we consider signal learning. We use Lemma \ref{lemma: interwined_sequence} again and take $a_0, b_0, A, B$ as $\la\wb_{j,r}^{(0)},\bmu\ra, v_0,\\$ $ 0.2 \eta \|\bmu\|_2^2, 0.2 \eta$. From the Condition \ref{condition: main condition} we know that these parameters satisfy the condition of Lemma \ref{lemma: interwined_sequence}. So we could get the time when signal learning becomes balancing:
\begin{align*}
    T_2 = \Theta(1/\sqrt{AB}) = \frac{1}{0.2 \eta \|\bmu\|_2} = \Theta\bigg(\frac{1}{\eta \|\bmu\|_2}\bigg).
\end{align*}
 Meanwhile, from Lemma \ref{lemma: interwined_sequence} we could derive that $v^{(T_1)}_{j,r,1} = \Theta(v_0) $, so there exists a constant $c'$ that for all $t \le T_2$, $v^{(t)}_{j,r,1}\le c' \cdot v_0$. 

But from Condition \ref{condition: main condition} that $\frac{\|\bmu\|_2}{\sigma_p\sqrt{d}} = O\big(\frac{1}{\sqrt{n}}\big)$, we have $T_1 \le T_2$, so at the time the output of noise becomes nearly constant, the signal part has not reached balancing stage. We could further derive that
\begin{align*}
\la\wb_{j,r}^{(T_1)},y_i \bmu\ra &\le \la\wb_{j,r}^{(0)},y_i \bmu\ra + T_1 \cdot (0.5+\kappa') \|\bmu\|_2^2 \cdot c'v_0,\\
\la\wb_{j,r}^{(T_1)},y_i \bmu\ra &\ge \la\wb_{j,r}^{(0)},y_i \bmu\ra + T_1/2 \cdot (0.5-\kappa') \|\bmu\|_2^2 \cdot c'v_0,
\end{align*}
which implies that $\la\wb_{j,r}^{(T_1)},y_i \bmu\ra = \Theta(T_1 \eta b_0 \|\bmu\|_2^2)$ and $v_{j,r,1}^{(T_1)} = \Theta(v_0)$. So when $SNR = o\big(\frac{1}{\sqrt{n}}\big)$, the balancing factor for the signal part is
\begin{align*}
\la\wb_{j,r}^{(T_1)},y_i \bmu\ra / v_{j,r,1}^{(T_1)} = \eta T_3 \|\bmu\|_2^2 = \tilde \Theta\bigg(\frac{\sqrt{n} \|\bmu\|_2^2}{\sigma_p \sqrt{d}}\bigg) = o(\|\bmu\|_2).
\end{align*}
Meanwhile, we could estimate the output of the signal part: 
\begin{align*}
v_{j,r,1}^{(T_1)} \la\wb_{j,r}^{(T_1)},y_i \bmu\ra = \Theta(T_1 \eta b_0^2 \|\bmu\|_2^2) = \tilde \Theta \Bigg(\frac{\sqrt{n} \|\bmu\|_2^2 v_0^2}{\sigma_p \sqrt{d}}\Bigg).
\end{align*}
\end{proof}

\begin{proof}[\textbf{\underline{Proof of Proposition~\ref{prop: stage1_s}}}]
Now we prove the main proposition \ref{prop: stage1_s} with all lemmas above. Note that all the lemmas hold when $\ell'^{(t)}_i \in [0.4, 0.6]$ for any $i \in [n]$, we will prove that within time $T^{*,1}$, $\ell'^{(t)}_i \in [0.4, 0.6]$ for any $i \in [n]$ holds.

Recall that we define $T^{*,1} = \max\{t:\max\limits_k \{v^{(t)}_{j,r,2}\cdot\sigma(\la\wb_{j,r}^{(t)},\bxi_k\ra)\} \le 0.1/m\}$ in (\ref{eq: define_T1_s}). We could bound the value of $|\ell'^{(t)}_i|$ within $T^{*,1}$ for all $i \in [n]$:
\begin{align*}
    |0.5 - |\ell'^{(t)}_i||&= \bigg|0.5 - \frac{1}{1+\exp\{y_i \cdot[F_{+1}(\bW_{+1}^{(t)},V_{+1},x_i)-F_{-1}(\bW_{-1}^{(t)},V_{-1},x_i)]\}}\bigg|\\
    &\le \bigg|\frac{\exp\{y_i \cdot[F_{+1}(\bW_{+1}^{(t)},V_{+1},x_i)-F_{-1}(\bW_{-1}^{(t)},V_{-1},x_i)]\}}{4}\bigg|,
\end{align*}
where the last inequity is due to $|\frac{1}{2} - \frac{1}{1+e^x}|\le |\frac{x}{4}|$ for any $x$. Then we have
\begin{align*}
    |\ell'^{(t)}_i| &\ge 0.5 - \frac{\exp\{y_i \cdot[F_{+1}(\bW_{+1}^{(t)},V_{+1},x_i)-F_{-1}(\bW_{-1}^{(t)},V_{-1},x_i)]\}}{4}\\
    &= 0.5 - \Theta(\exp\{y_i \cdot[F_{+1}(\bW_{+1}^{(t)},V_{+1},x_i)-F_{-1}(\bW_{-1}^{(t)},V_{-1},x_i)]\})\\
    &= 0.5 - \Theta(m \cdot \max\{v^{(t)}_{j,r,2}\cdot\sigma(\la\wb_{j,r}^{(t)},\bxi_k\ra), v^{(t)}_{j,r,1}\cdot\sigma(\la\wb_{j,r}^{(t)},\bmu\ra)\})\\
    &= 0.5 - \Theta(m \cdot \max \{v^{(t)}_{j,r,2}\cdot\sigma(\la\wb_{j,r}^{(t)},\bxi_k\ra)\})\\
    &= 0.5 - 0.1\\
    &= 0.4.
\end{align*}
 Similarly we could derive $|\ell'^{(t)}_i| \le 0.6$. So we have $|\ell'^{(t)}_i|\in [0.4, 0.6]$ for all $t \le T^{*,1}$.

Then we estimate the scale of $T^{*,1}$. From the analysis in Lemma \ref{lemma: noise_output_constant}, we know that
\begin{align*}
v^{(t)}_{j,r,2}\cdot\sigma(\la\wb_{j,r}^{(t)},\bxi_k\ra) &\ge \exp\bigg(\Theta \bigg(\frac{\eta \sigma_p \sqrt{d}}{\sqrt{n}} \bigg)(t-T_1) \bigg) \cdot \Theta \Bigg(\frac{\sigma_p \sqrt{d} \cdot v_0^2}{\sqrt{n}}\Bigg),\\
v^{(t)}_{j,r,2}\cdot\sigma(\la\wb_{j,r}^{(t)},\bxi_k\ra) &\le \exp\bigg(\Theta\bigg(\frac{\eta \sigma_p \sqrt{d}}{\sqrt{n}}\bigg) 2(t-T_0)\bigg) \cdot \Theta\Bigg(\frac{,\sigma_p \sqrt{d} \cdot v_0^2}{\sqrt{n}}\Bigg).
\end{align*}
So when $v^{(t)}_{j,r,2}\cdot\sigma(\la\wb_{j,r}^{(t)},\bxi_k\ra) = 0.1/m$, we have $t= T^{*,1} = \Theta\Bigg(\frac{\sqrt{n} \cdot \log(\frac{\sqrt{n}}{\sigma_p \sqrt{d} v_0^2})}{\eta m \sigma_p \sqrt{d}}\Bigg) + T_0 = \tilde O\Big(\frac{\sqrt{n}}{\eta \sigma_p \sqrt{d}}\Big)$, which is the same order as $T_1$. But as mentioned before, we have $T_1 \le T^{*,1}$ for $v^{(T_1)}_{j,r,2}\cdot\sigma(\la\wb_{j,r}^{(T_1)},\bxi_k\ra) \le 0.05/m \le 0.1/m$.
Therefore, all the lemmas above hold within time $T_1$. So we could derive the first statement of Proposition \ref{prop: stage1_s} from lemma \ref{lemma: noise_balanced} and the last three statements from lemma \ref{lemma: noise_output_constant}, which completes the proof of Proposition~\ref{prop: stage1_s}.
\end{proof}

\subsection{Second stage}
From the previous analysis, the noise part will go over two phases in the first stage and become balanced while the signal part remains unbalanced. We give the main proposition to estimate the signal and noise output in this case. Here $T^* = \eta^{-1} poly(d, n, m,\varepsilon^{-1})$ is the maximum admissible iterations as above.

\begin{proposition}
\label{prop: stage2_analysis_c2}
    For any $T_1 \le t \le T^*$, it holds that,
    \begin{align}
        &0 \le v_{j,r,1}^{(t)} \le C_4 v_0 \alpha, \label{eq: v1_bound_c2}\\
        &0 \le \gamma_{j,r}^{(t)} \le \frac{ C_5 \sqrt{n}  v_0 \|\bmu\|_2^2 \alpha}{\sigma_p \sqrt{d}},\label{eq: signal_output_bound_c2}\\
        &0 \le \overline{\rho}_{j,r,k}^{(t)} \le \alpha,\label{eq: pos_noise_bound_c2}\\
        &0 \ge \underline{\rho}_{j,r,k}^{(t)} \ge -2\sqrt{\log(8mn/\delta)}\cdot \sigma_0 \sigma_p\sqrt{d} - 8 \sqrt{\frac{\log(4n^2/\delta)}{d}} n\alpha \ge -\alpha \label{eq: neg_noise_bound_c2}
    \end{align}
    for all $j \in \{\pm 1\}, r \in [m]$ and $k \in [n]$. Here $\alpha = 4\sqrt{\frac{\sigma_p \sqrt{d} \log(T^*)}{k_1 m \sqrt{n}}}$ where $k_1 = \min\limits_{k \in [n], t \le T_1} \Big\{\frac{v_{j,r,2}^{(t)} \sigma_p \sqrt{d}}{\sqrt{n}\la \wb_{j,r}^{(t)}, \bxi_k \ra}\Big\} = \Theta(1)$ and $C_4, C_5=\Theta(1)$.
\end{proposition}

We also need some preliminary lemmas to prove this proposition. First note that the proof of Lemma \ref{lemma: inner_rho_diff_c1} and Lemma \ref{lemma: output_leading_term_c1} do not depend on the lower bound of $v_0$ and $v_0$, so these two lemmas still hold in this case. We restate them for convenience:

\begin{lemma}
\label{lemma: inner_rho_diff_c2}
    Under Condition \ref{condition: main condition}, suppose (\ref{eq: v1_bound_c2}) to (\ref{eq: neg_noise_bound_c2}) hold at iteration t. Then for all $k \in [n]$ it holds that
    \begin{align}
        &| \la \wb_{-y_k, r}^{(t)} - \wb_{-y_k, r}^{(0)}, \bxi_k \ra - \underline{\rho}_{-y_k, r, k}^{(t)}| \le 4n \alpha \sigma_p^2 \cdot \sqrt{\log(4n^2/\delta)/d}\label{eq: under_rho_diff_c1_s}\\
        &| \la \wb_{y_k, r}^{(t)} - \wb_{y_k, r}^{(0)}, \bxi_k \ra - \overline{\rho}_{y_k, r, k}^{(t)}| \le 4n \alpha \sigma_p^2 \cdot \sqrt{\log(4n^2/\delta)/d} \label{eq: over_rho_diff_c1_s}
    \end{align}
\end{lemma}

\begin{lemma}
\label{lemma: output_leading_term_c2}
    Under Condition \ref{condition: main condition}, suppose (\ref{eq: v1_bound_c2}) to (\ref{eq: neg_noise_bound_c2}) hold at iteration t. Then for all $k \in [n]$ it holds that
    \begin{align*}
        &F_{-y_k}(\bW_{-y_k}^{(t)}, \bV_{-y_k}^{(t)}; \bx_k) \le 0.25,\\
        &-0.25 + \sum_{r=1}^m v_{y_k, r, 2}^{(t)} \sigma(\la \wb_{y_k, r}^{(t)}, \bxi_k \ra) \le F_{y_k}(\bW_{y_k}^{(t)}, \bV_{y_k}^{(t)}; \bx_k) \le 0.25 + \sum_{r=1}^m v_{y_k, r, 2}^{(t)} \sigma(\la \wb_{y_k, r}^{(t)}, \bxi_k \ra).
    \end{align*}

    Further, as $y_k f(\bW^{(t)}, \bV^{(t)}; \bx_k) = F_{y_k}(\bW_{y_k}^{(t)}, \bV_{y_k}^{(t)}; \bx_k) - F_{-y_k}(\bW_{-y_k}^{(t)}, \bV_{-y_k}^{(t)}; \bx_k)$, we have
    \begin{equation}
    \label{eq: output_leading_term_c2}
        -0.5 + \sum_{r=1}^m v_{y_k, r, 2}^{(t)} \sigma(\la \wb_{y_k, r}^{(t)}, \bxi_k \ra) \le y_k \cdot f(\bW^{(t)}, \bV^{(t)}; \bx_k) \le 0.5 + \sum_{r=1}^m v_{y_k, r, 2}^{(t)} \sigma(\la \wb_{y_k, r}^{(t)}, \bxi_k \ra).
    \end{equation}
\end{lemma}

Then we introduce the key lemma for the main proof.
\begin{lemma}
\label{lemma: key_lemma_stage2_c2}
    Under Condition \ref{condition: main condition}, suppose (\ref{eq: v1_bound_c2}) to (\ref{eq: neg_noise_bound_c2}) hold at iteration t. Then it holds that:
    \begin{enumerate}
        \item $\sum_{r=1}^m v_{y_k, r, 2}^{(t)} \sigma(\la \wb_{y_k, r}^{(t)}, \bxi_k \ra) - v_{y_i, r, 2}^{(t)} \sigma(\la \wb_{y_i, r}^{(t)}, \bxi_i \ra) \le \kappa$ and $\ell'^{(t)}_i / \ell'^{(t)}_k \le C_6$ for all $i, k \in [n]$ and $y_i = y_k$.
        \item $v_{j,r,2}^{(t)} \ge C_7 v_0$ for all $j \in \{\pm 1\}, r \in [m]$.
        \item Define $S_i^{(t)} := \{r \in [m]: \la w_{y_i,r},\bxi_k \ra >0\}$, and $S_{j,r}^{(t)}:=\{k\in [n]:y_k=j,\la w_{j,r}^{(t)}, \bxi_k \ra > 0\}$. For all $k \in [n], r \in [m]$ and $j \in \{\pm 1\}$, $S_k^{(0)} \subseteq S_k^{(t)}, S_{j,r}^{(0)} \subseteq S_{j,r}^{(t)}$.
        \item $\Theta \Big(\frac{1}{\|\bmu\|_2}\Big) \le \frac{v_{y_k, r, 1}^{(t)}}{\la \wb_{y_k, r}^{(t)}, y_k\bmu \ra} \le \Theta\Big(\frac{\sigma_p \sqrt{d}}{\sqrt{n} \|\bmu\|_2^2}\Big)$ for all $k \in [n], r \in [m]$.
        \item $\frac{v_{y_k,r,2}^{(t)}}{\la \wb_{y_k,r}^{(t)}, \bxi_k \ra} = \Theta\Big(\frac{\sqrt{n}}{\sigma_p \sqrt{d}}\Big)$ for all $k \in [n], r \in [m]$.
    \end{enumerate}
    Here, $\kappa, C_6, C_7$ can be taken as $1, \exp(1.5)$ and $0.5$.
\end{lemma}

\begin{proof}[Proof of Lemma \ref{lemma: key_lemma_stage2_c2}]
    We use induction to prove this lemma. We first prove that all conclusions hold when $t = T_1$. From Proposition \ref{prop: stage1_s} we have
    \begin{align*}
        v^{(T_1)}_{y_k,r,2}\cdot\sigma(\la\wb_{y_k,r}^{(T_1)},\bxi_k\ra) \le 0.1/m.
    \end{align*}
    Therefore, 
    \begin{align*}
        \sum_{r=1}^m v_{y_k, r, 2}^{(t)} \sigma(\la \wb_{y_k, r}^{(t)}, \bxi_k \ra) - v_{y_i, r, 2}^{(t)} \sigma(\la \wb_{y_i, r}^{(t)}, \bxi_i \ra) &\le \sum_{r=1}^m |v_{y_k, r, 2}^{(t)} \sigma(\la \wb_{y_k, r}^{(t)}, \bxi_k \ra)| + \sum_{r=1}^m |v_{y_i, r, 2}^{(t)} \sigma(\la \wb_{y_i, r}^{(t)}, \bxi_i \ra)|\\
        &= 0.1+0.1\\
        &= 0.2\\
        &\le \kappa.
    \end{align*}
    Meanwhile, as $\ell'^{(t)}_k \in [0.4, 0.6]$ for all $k \in [n]$, we have $\ell'^{(t)}_i / \ell'^{(t)}_k \le 1.5 \le C_6$. So the first conclusion holds.

    Then from Proposition \ref{prop: stage1_s} and Lemma \ref{lemma: active neuron}, we could directly derive the last three conclusions.

    Now, suppose that there exists $\tilde t \le T^*$ such that five conditions hold for any $0 \le t \le \tilde t-1$, we prove that these conditions also hold for $t = \tilde t$.

    We first prove conclusion 1. From (\ref{eq: output_leading_term_c2}) we have
    \begin{align}
        \bigg|y_k \cdot f(\bW^{(t)}, \bV^{(t)}; \bx_k) - y_i \cdot f(\bW^{(t)}, \bV^{(t)}; \bx_i) - \sum_{r=1}^m \Big[v_{y_k, r, 2}^{(t)} \sigma(\la \wb_{y_k, r}^{(t)}, \bxi_k \ra) - v_{y_i, r, 2}^{(t)} \sigma(\la \wb_{y_i, r}^{(t)}, \bxi_i \ra)\Big]\bigg| \le 0.5.\label{eq: output_diff}
    \end{align}

    Recall the update equation of $v_{y_k, r, 2}^{(t)} \sigma(\la \wb_{y_k, r}^{(t)}, \bxi_k \ra)$:
    \begin{align*}
        &v_{y_k, r, 2}^{(\tilde t)} \sigma(\la \wb_{y_k, r}^{(\tilde t)}, \bxi_k \ra) \\
        &\quad= v_{y_k, r, 2}^{(\tilde t-1)} \sigma(\la \wb_{y_k, r}^{(\tilde t-1)}, \bxi_k \ra) - \frac{\eta}{n} v_{y_k, r, 2}^{(\tilde t-1)2} \|\bxi_k\|_2^2 \ell'^{(\tilde t-1)}_k
        - \frac{y_k \eta}{n} \la \wb_{y_k, r}^{(\tilde t-1)}, \bxi_k \ra \sum_{i=1}^n \ell'^{(\tilde t-1)}_i y_i \la \wb_{y_k, r}^{(\tilde t-1)}, \bxi_i \ra \mathbb{I}(\la \wb_{y_k, r}^{(\tilde t-1)}, \bxi_i \ra > 0)\\
        &\quad\quad+ \frac{\eta^2}{n^2} v_{y_k, r, 2}^{(\tilde t-1)} \sum_{i=1, i\neq k}^n y_i \ell'^{(\tilde t-1)}_i \la \bxi_k, \bxi_i \ra \mathbb{I}(\la \wb_{y_k, r}^{(\tilde t-1)}, \bxi_i \ra \ge 0) \sum_{i=1}^n \ell'^{(\tilde t-1)}_i y_i \la \wb_{y_k, r}^{(\tilde t-1)}, \bxi_i \ra \mathbb{I}(\la \wb_{y_k, r}^{(\tilde t-1)}, \bxi_i \ra \ge 0).
    \end{align*}
    From the last induction hypothesis, we have $\frac{v_{j,r,2}^{(t)}}{\la \wb_{j,r}^{(t)}, \bxi_k \ra} = \Theta(\frac{\sqrt{n}}{\sigma_p \sqrt{d}})$, and we could assume there exist $k_1, k_2 = \Theta(1)$ that $\frac{k_1 \sqrt{n}}{\sigma_p \sqrt{d}} \le  \frac{v^{(t)}_{j,r,2}}{\la\wb_{j,r}^{(t)},\bxi_k\ra} \le \frac{k_2 \sqrt{n}}{\sigma_p \sqrt{d}}$ for all $j \in \{\pm 1\}$ and $k \in [n]$, thus,
    \begin{align*}
        v_{y_k, r, 2}^{(\tilde t)} \sigma(\la \wb_{y_k, r}^{(\tilde t)}, \bxi_k \ra) \le& v_{y_k, r, 2}^{(\tilde t-1)} \sigma(\la \wb_{y_k, r}^{(\tilde t-1)}, \bxi_k \ra) - \frac{\eta}{n} v_{y_k, r, 2}^{(\tilde t-1)2} \|\bxi_k\|_2^2 \ell'^{(\tilde t-1)}_k\\
        &- \frac{\eta}{n} \sum_{i=1}^n \ell'^{(\tilde t-1)}_i \cdot \frac{\sigma_p^2 d v_{y_k,r,2}^{(\tilde t-1)2}}{k_1^2 n} \cdot \mathbb{I}(\la \wb_{y_k, r}^{(\tilde t-1)}, \bxi_i \ra > 0)\\
        \le& v_{y_k, r, 2}^{(\tilde t-1)} \sigma(\la \wb_{y_k, r}^{(\tilde t-1)}, \bxi_k \ra) - \frac{\eta}{n} v_{y_k, r, 2}^{(\tilde t-1)2} \ell'^{(\tilde t-1)}_k \bigg(\|\bxi_k\|_2^2 + (C_6/k_1^2) \sigma_p^2 d \cdot (|S_{y_k, r}^{(\tilde t-1)}|/n)\bigg),
    \end{align*}
    where the last inequality is from the first induction hypothesis.

    We then lower bound $v_{y_k, r, 2}^{(\tilde t)} \sigma(\la \wb_{y_k, r}^{(\tilde t)}, \bxi_i \ra)$ that
    \begin{align*}
        v_{y_k, r, 2}^{(\tilde t)} \sigma(\la \wb_{y_k, r}^{(\tilde t)}, \bxi_i \ra) \ge& v_{y_k, r, 2}^{(\tilde t-1)} \sigma(\la \wb_{y_k, r}^{(\tilde t-1)}, \bxi_i \ra) - \frac{\eta}{n} v_{y_k, r, 2}^{(\tilde t-1)2} \|\bxi_i\|_2^2 \ell'^{(\tilde t-1)}_i\\
        &- \frac{\eta}{n} \sum_{i'=1}^n \ell'^{(\tilde t-1)}_{i'} \cdot \frac{\sigma_p^2 d v_{y_k,r,2}^{(\tilde t-1)2}}{k_2^2 n} \cdot \mathbb{I}(\la \wb_{y_k, r}^{(\tilde t-1)}, \bxi_{i'} \ra > 0)\\
        \ge& v_{y_k, r, 2}^{(\tilde t-1)} \sigma(\la \wb_{y_k, r}^{(\tilde t-1)}, \bxi_i \ra) - \frac{\eta}{n} v_{y_k, r, 2}^{(\tilde t-1)2} \ell'^{(\tilde t-1)}_i \bigg(\|\bxi_i\|_2^2 + (C_6 k_2^2)^{-1} \sigma_p^2 d \cdot (|S_{y_k, r}^{(\tilde t-1)}|/n)\bigg).
    \end{align*}
    The last inequality is also from the first induction hypothesis.
    Combining the above two inequalities, we have
    \begin{align*}
        &\sum_{r=1}^m v_{y_k, r, 2}^{(\tilde t)} \sigma(\la \wb_{y_k, r}^{(\tilde t)}, \bxi_k \ra) - v_{y_k, r, 2}^{(\tilde t)} \sigma(\la \wb_{y_k, r}^{(\tilde t)}, \bxi_i \ra) \\
        \le& \sum_{r=1}^m v_{y_k, r, 2}^{(\tilde t-1)} \sigma(\la \wb_{y_k, r}^{(\tilde t-1)}, \bxi_k \ra) - v_{y_k, r, 2}^{(\tilde t-1)} \sigma(\la \wb_{y_k, r}^{(\tilde t-1)}, \bxi_i \ra)\\
        -& \bigg[\frac{\eta}{n} v_{y_k, r, 2}^{(\tilde t-1)2} \ell'^{(\tilde t-1)}_k \bigg(\|\bxi_k\|_2^2 + (C_6/k_1^2) \sigma_p^2 d \cdot (|S_{y_k, r}^{(\tilde t-1)}|/n)\bigg) - \frac{\eta}{n} v_{y_k, r, 2}^{(\tilde t-1)2} \ell'^{(\tilde t-1)}_i \bigg(\|\bxi_i\|_2^2 + (C_6 k_2^2)^{-1} \sigma_p^2 d \cdot (|S_{y_k, r}^{(\tilde t-1)}|/n)\bigg)\bigg].
    \end{align*}
    We consider two cases: $\sum_{r=1}^m v_{y_k, r, 2}^{(\tilde t-1)} \sigma(\la \wb_{y_k, r}^{(\tilde t-1)}, \bxi_k \ra) - v_{y_k, r, 2}^{(\tilde t-1)} \sigma(\la \wb_{y_k, r}^{(\tilde t-1)}, \bxi_i \ra) \le 0.9\kappa$ and $\sum_{r=1}^m v_{y_k, r, 2}^{(\tilde t-1)} \sigma(\la \wb_{y_k, r}^{(\tilde t-1)}, \bxi_k \ra) - v_{y_k, r, 2}^{(\tilde t-1)} \sigma(\la \wb_{y_k, r}^{(\tilde t-1)}, \bxi_i \ra) > 0.9\kappa$. When $\sum_{r=1}^m v_{y_k, r, 2}^{(\tilde t-1)} \sigma(\la \wb_{y_k, r}^{(\tilde t-1)}, \bxi_k \ra) - v_{y_k, r, 2}^{(\tilde t-1)} \sigma(\la \wb_{y_k, r}^{(\tilde t-1)}, \bxi_i \ra) \le 0.9\kappa$, we can first upper bound $v_{y_k, r, 2}^{(\tilde t-1)}$ as
    \begin{align*}
        v_{y_k, r, 2}^{(\tilde t-1)} &\le \frac{k_2 \sqrt{n}}{\sigma_p \sqrt{d}}\cdot \la \wb_{y_k, r}^{(\tilde t-1)}, \bxi_k \ra\\
        &\le \frac{k_2 \sqrt{n}}{\sigma_p \sqrt{d}} \cdot \bigg[\la\wb_{j,r}^{(0)},\bxi_k\ra + \rho_{j,r,k}^{(\tilde t-1)} + 4 \sqrt{\frac{\log(4n^2/\delta)}{d}} n\alpha \bigg]\\
        &\le \frac{2\alpha k_2 \sqrt{n}}{\sigma_p \sqrt{d}}.
    \end{align*}
     The first inequality is from our last induction hypothesis; The second inequality is from Lemma \ref{lemma: inner_rho_diff_c2} and the last inequality is from (\ref{eq: pos_noise_bound_c2}).
     
    So we have
    \begin{align*}
        \sum_{r=1}^m v_{y_k, r, 2}^{(\tilde t)} \sigma(\la \wb_{y_k, r}^{(\tilde t)}, \bxi_k \ra) - v_{y_k, r, 2}^{(\tilde t)} \sigma(\la \wb_{y_k, r}^{(\tilde t)}, \bxi_i \ra) &\le 0.9\kappa - \frac{\eta}{n} \sum_{r=1}^m v_{y_k, r, 2}^{(\tilde t-1)2} \ell'^{(\tilde t-1)}_k \bigg(\|\bxi_k\|_2^2 + (C_6/k_1^2) \sigma_p^2 d \cdot (|S_{y_k, r}^{(\tilde t-1)}|/n)\bigg)\\
        &\le 0.9\kappa + \frac{\eta}{n} \sum_{r=1}^m v_{y_k, r, 2}^{(\tilde t-1)2} (\|\bxi_k\|_2^2 + (C_6/k_1^2) \sigma_p^2 d)\\
        &\le 0.9 \kappa + \frac{\eta m}{n} \Big(\frac{2\alpha k_2 \sqrt{n}}{\sigma_p \sqrt{d}}\Big)^2 (3/2 + C_6/k_1^2) \sigma_p^2 d\\
        &\le 0.9 \kappa + 0.1 \kappa\\
        &\le \kappa.
    \end{align*}
    The first inequality is due to $\ell'^{(\tilde t-1)}_i \le 0$; The second inequality is due to $|S_{y_k, r}^{(\tilde t -1)}| \le n$ and the last inequality is from the definition of $\eta$.

    On the other hand, when  $v_{y_k, r, 2}^{(\tilde t-1)} \sigma(\la \wb_{y_k, r}^{(\tilde t-1)}, \bxi_k \ra) - v_{y_k, r, 2}^{(\tilde t-1)} \sigma(\la \wb_{y_k, r}^{(\tilde t-1)}, \bxi_i \ra) > 0.9\kappa$, from (\ref{eq: output_diff}) we have
    \begin{align*}
        y_k \cdot f(\bW^{(\tilde t-1)}, \bV^{(\tilde t-1)}; \bx_k) - y_k \cdot f(\bW^{(\tilde t-1)}, \bV^{(\tilde t-1)}; \bx_i) &\ge \sum_{r=1}^m \Big[v_{y_k, r, 2}^{(\tilde t-1)} \sigma(\la \wb_{y_k, r}^{(\tilde t-1)}, \bxi_k \ra) - v_{y_k, r, 2}^{(\tilde t-1)} \sigma(\la \wb_{y_k, r}^{(\tilde t-1)}, \bxi_i \ra)\Big] - 0.5\\
        &\ge 0.9 \kappa - 0.5 \kappa\\
        &\ge 0.4 \kappa.
    \end{align*}
    The second inequality is from $\kappa = 1$. So we have
    \begin{align*}
        \frac{\ell'^{(\tilde t-1)}_i}{\ell'^{(\tilde t-1)}_k} \le \exp(y_k \cdot f(\bW^{(\tilde t-1)}, \bV^{(\tilde t-1)}; \bx_k) - y_k \cdot f(\bW^{(\tilde t-1)}, \bV^{(\tilde t-1)}; \bx_i)) \le \exp(-0.4 \kappa).
    \end{align*}
    The first inequality is due to $\frac{1 + \exp(b)}{1 + \exp(a)} \le \exp(b-a)$ for $b \ge a \ge 0$.
    
    Therefore, for all $r \in [m]$ we have
    \begin{align*}
        &\frac{v_{y_k, r, 2}^{(\tilde t-1)} \ell'^{(\tilde t-1)}_i (\|\bxi_i\|_2^2 + (C_6 k_2^2)^{-1} \sigma_p^2 d \cdot (|S_{y_i, r}^{(\tilde t-1)}|/n))}{v_{y_k, r, 2}^{(\tilde t)} \ell'^{(\tilde t-1)}_k (\|\bxi_k\|_2^2 + (C_6/k_1^2) \sigma_p^2 d \cdot (|S_{y_k, r}^{(\tilde t-1)}|/n))}\\
        =& \frac{\ell'^{(\tilde t-1)}_i (\|\bxi_i\|_2^2 + (C_6 k_2^2)^{-1} \sigma_p^2 d \cdot (|S_{y_i, r}^{(\tilde t-1)}|/n))}{\ell'^{(\tilde t-1)}_k (\|\bxi_k\|_2^2 + (C_6/k_1^2) \sigma_p^2 d \cdot (|S_{y_k, r}^{(\tilde t-1)}|/n))}\\
        \le& \frac{\ell'^{(\tilde t-1)}_i}{\ell'^{(\tilde t-1)}_k} \cdot \max\bigg\{\frac{\|\bxi_i\|_2^2}{\|\bxi_k\|_2^2}, \frac{k_1^2}{C_6^2 k_2^2}\bigg\}\\        
        \le& \frac{\ell'^{(\tilde t-1)}_i \|\bxi_i\|_2^2}{\ell'^{(\tilde t-1)}_k \|\bxi_k\|_2^2} \\
        \le& \exp(-0.4 \kappa)\\
        \le& 1.
    \end{align*}
    The first equality is from $y_i = y_k$, so $v_{y_i, r, 2}^{(\tilde t-1)} = v_{y_k, r, 2}^{(\tilde t-1)}$; The first inequality is from $\frac{a+b}{c+d} \le \max\{\frac{a}{c}, \frac{b}{d}\}$ and $|S_{y_i, r}^{(\tilde t-1)}| = |S_{y_k, r}^{(\tilde t-1)}|$; The second inequality is due to $k_1 \le k_2$ and $C_6 = \exp(1.5)$, so $\frac{k_1^2}{C_6^2 k_2^2} \le 1$; The third inequality is from Lemma \ref{lemma: noise bound} that
    \begin{align*}
         \big|\|\bxi_k\|_2^2-\sigma_p^2d \big| = O\big(\sigma_p^2 \cdot \sqrt{d\log(4n/\delta)}\big).
    \end{align*}
    So it follows that
    \begin{align*}
        |v_{y_i, r, 2}^{(\tilde t-1)} \ell'^{(\tilde t-1)}_i (\|\bxi_i\|_2^2 + (C_6 k_2^2)^{-1} \sigma_p^2 d \cdot (|S_{y_i, r}^{(\tilde t-1)}|/n))| \le |v_{y_k, r, 2}^{(\tilde t)} \ell'^{(\tilde t-1)}_k (\|\bxi_k\|_2^2 + (C_6/k_1^2) \sigma_p^2 d \cdot (|S_{y_k, r}^{(\tilde t-1)}|/n))|.
    \end{align*}
    Therefore,
    \begin{align*}
        v_{y_k, r, 2}^{(\tilde t)} \sigma(\la \wb_{y_k, r}^{(\tilde t)}, \bxi_k \ra) - v_{y_i, r, 2}^{(\tilde t)} \sigma(\la \wb_{y_i, r}^{(\tilde t)}, \bxi_i \ra)
        \le v_{y_k, r, 2}^{(\tilde t-1)} \sigma(\la \wb_{y_k, r}^{(\tilde t-1)}, \bxi_k \ra) - v_{y_i, r, 2}^{(\tilde t-1)} \sigma(\la \wb_{y_i, r}^{(\tilde t-1)}, \bxi_i \ra) \le \kappa.
    \end{align*}
    Meanwhile, from (\ref{eq: output_diff}) we have 
    \begin{align*}
    &y_k \cdot f(\bW^{(\tilde t)}, \bV^{(\tilde t)}; \bx_k) - y_i \cdot f(\bW^{(\tilde t)}, \bV^{(\tilde t)}; \bx_i) \le 0.5 + \kappa, \\
        &\frac{\ell'^{(\tilde t)}_i}{\ell'^{(\tilde t)}_k} \le \exp(y_k \cdot f(\bW^{(\tilde t)}, \bV^{(\tilde t)}; \bx_k) - y_i \cdot f(\bW^{(\tilde t)}, \bV^{(\tilde t)}; \bx_i)) \le \exp(0.5 + \kappa) = C_6.
    \end{align*}

Then we prove conclusion 2. Denote $\ell'^{(t)}_{min} = -\min\limits_{i \in [n]} |\ell'^{(t)}_i|$ and $\ell'^{(t)}_{max} = -\max\limits_{i \in [n]} |\ell'^{(t)}_i|$ . We could lower bound $v_{j,r,2}^{(t)}$ as
\begin{align*}
    v_{j,r,2}^{(t)}  =& v_{j,r,2}^{(t-1)} - \frac{j \eta}{n}\sum_{i=1}^{n}{\ell'^{(t-1)}_i}y_i\la \wb_{j,r}^{(t-1)},\bxi_i \ra \mathbb{I}(\la\wb_{j,r}^{(t-1)},\bxi_i\ra >0)\\
    =& v_{j,r,2}^{(t-1)} +\frac{\eta}{n} \sum_{k=1,y_k\neq j}^n {\ell'^{(t-1)}_k} \la \wb_{j,r}^{(t-1)},\bxi_k \ra \mathbb{I}(\la\wb_{j,r}^{(t-1)},\bxi_k\ra >0) \\
    &- \frac{\eta}{n} \sum_{k=1,y_k=j}^n {\ell'^{(t-1)}_k} \la \wb_{j,r}^{(t-1)},\bxi_k \ra \mathbb{I}(\la\wb_{j,r}^{(t-1)},\bxi_k\ra >0)\\
    \ge& v_{j,r,2}^{(t-1)} + \frac{\eta}{n} \sum_{k=1, y_k \neq j}^n \ell'^{(t)}_{max} \bigg(\la \wb_{j,r}^{(0)}, \bxi_k \ra + \underline{\rho}_{j,r,k}^{(t-1)} +\sum_{i=1, i\neq k}^n \rho_{j,r,i}^{(t-1)} \frac{\la \bxi_i, \bxi_k \ra}{\|\bxi_i\|_2^2}\bigg)\\
    &- \frac{\eta}{n} \sum_{k=1, y_k = j}^n \ell'^{(t)}_{min} \bigg(\la \wb_{j,r}^{(0)}, \bxi_k \ra + \overline{\rho}_{j,r,k}^{(t-1)} +\sum_{i=1, i\neq k}^n \rho_{j,r,i}^{(t-1)} \frac{\la \bxi_i, \bxi_k \ra}{\|\bxi_i\|_2^2}\bigg)\\
    \ge& v_{j,r,2}^{(t-1)} + \frac{\eta}{n} \sum_{k=1, y_k \neq j}^n \ell'^{(t)}_{max} \bigg(\la \wb_{j,r}^{(0)}, \bxi_k \ra + \underline{\rho}_{j,r,k}^{(t-1)} +\sum_{i=1, i\neq k}^n \overline{\rho}_{j,r,i}^{(t-1)} \frac{\la \bxi_i, \bxi_k \ra}{\|\bxi_i\|_2^2}\bigg)\\
    &- \frac{\eta}{n} \sum_{k=1, y_k = j}^n \ell'^{(t)}_{min} \bigg(\la \wb_{j,r}^{(0)}, \bxi_k \ra + \overline{\rho}_{j,r,k}^{(t-1)} +\sum_{i=1, i\neq k}^n \underline{\rho}_{j,r,i}^{(t-1)} \frac{\la \bxi_i, \bxi_k \ra}{\|\bxi_i\|_2^2}\bigg)\\
    =& v_{j,r,2}^{(t-1)} + \frac{\eta}{n} \ell'^{(t)}_{max} \sum_{k=1, y_k \neq j}^n \la \wb_{j,r}^{(0)}, \bxi_k \ra + \underline{\rho}_{j,r,k}^{(t)}\bigg(1 - \frac{\ell'^{(t)}_{min}}{\ell'^{(t)}_{max}} \sum_{i=1, i \neq k}^n \frac{|\la \bxi_i, \bxi_k \ra|}{\|\bxi_i\|_2^2}\bigg)\\
    &-\frac{\eta}{n} \ell'^{(t)}_{min}
    \sum_{k=1, y_k = j}^n  \la \wb_{j,r}^{(0)}, \bxi_k \ra + \overline{\rho}_{j,r,k}^{(t)}\bigg(1  - \frac{\ell'^{(t)}_{max}}{\ell'^{(t)}_{min}} \sum_{i=1, i \neq k}^n \frac{|\la \bxi_i, \bxi_k \ra|}{\|\bxi_i\|_2^2}\bigg)\\
    \ge& v_{j,r,2}^{(t-1)} + \frac{\eta}{n} {\ell'^{(t)}_{max}} \sum_{k=1, y_k \neq j}^n 2 \sqrt{\log(8mn/\delta)}\cdot \sigma_0 \sigma_p\sqrt{d} + \underline{\rho}_{j,r,k}^{(t)}(1 - C_6 \sum_{i=1}^n 4 \sigma_p^2 \cdot \sqrt{\log(4n^2/\delta)/d})\\
    &- \frac{\eta}{n} {\ell'^{(t)}_{min}}\sum_{k=1, y_k = j}^n -2 \sqrt{\log(8mn/\delta)}\cdot \sigma_0 \sigma_p\sqrt{d} + \underline{\rho}_{j,r,k}^{(t)}(1 - \frac{1}{C_6} \sum_{i=1}^n 4 \sigma_p^2 \cdot \sqrt{\log(4n^2/\delta)/d})\\
    \ge& v_{j,r,2}^{(t-1)} - \eta \sqrt{\log(8mn/\delta)}\cdot \sigma_0 \sigma_p\sqrt{d} - \frac{\eta}{n}\sum_{k=1, y_k \neq j}^n 0.5 \underline{\rho}_{j,r,k}^{(t)} + \frac{\eta}{n} \sum_{k=1, y_k = j}^n 0.5 \overline{\rho}_{j,r,k}^{(t)}\\
    \ge& v_{j,r,2}^{(t-1)}.
\end{align*}
The first inequality is from the definition of $\ell'^{(t)}_{min}$ and $\ell'^{(t)}_{max}$; The second inequality is from the monotonicity of $\rho$ so that $\overline{\rho} \ge 0$ and $\underline{\rho} \le 0$; The third inequality is from Lemma \ref{lemma: noise bound} and Lemma \ref{lemma: preliminary_initial inner product} as well as the first induction hypothesis; The fourth inequality is from Condition \ref{condition: main condition} the definition of d; The last inequality is from the monotonicity of $\rho$ so the negative term $\eta \sqrt{\log(8mn/\delta)}\cdot \sigma_0 \sigma_p\sqrt{d}$ is canceled out.

    Next we prove conclusion 3. Recall the update equation of $\la \wb_{j,r}^{(t)}, \bxi_k \ra$ when $y_k = j$:
    \begin{align*}
    \la\wb_{j,r}^{(t)},\bxi_k\ra &= \la\wb_{j,r}^{( t-1)},\bxi_k\ra - \frac{\eta}{n} v_{j,r,2}^{( t-1)} \|\bxi_k\|_2^2 {\ell'}_{k}^{( t-1)} -\frac{j \eta}{n} v_{j,r,2}^{( t-1)} \sum_{i=1,i\neq k}^n y_i {\ell'}_{i}^{( t-1)} \la \bxi_k,\bxi_i \ra \mathbb{I}(\la\wb_{j,r}^{( t-1)},\bxi_i\ra >0).
    \end{align*}
    Note that $S_k^{(0)} \subseteq S_k^{(\tilde t)}, S_{j,r}^{(0)} \subseteq S_{j,r}^{(\tilde t)}$ both equal to $\la\wb_{j,r}^{(\tilde t)},\bxi_k\ra > 0$ so we only need to prove this. From induction hypothesis, we have $\la\wb_{j,r}^{(\tilde t-1)},\bxi_k\ra > 0$, and we have
    \begin{align*}
        &- \frac{\eta}{n} v_{j,r,2}^{(\tilde t-1)} \|\bxi_k\|_2^2 {\ell'}_{k}^{(\tilde t-1)} -\frac{j \eta}{n} v_{j,r,2}^{(\tilde t-1)} \sum_{i=1,i\neq k}^n y_i {\ell'}_{i}^{(\tilde t-1)} \la \bxi_k,\bxi_i \ra \mathbb{I}(\la\wb_{j,r}^{(\tilde t-1)},\bxi_i\ra >0)\\
        \ge&  - \frac{\eta}{n} v_{j,r,2}^{(\tilde t-1)} \|\bxi_k\|_2^2 {\ell'}_{k}^{(\tilde t-1)} - \frac{\eta}{n} v_{j,r,2}^{(\tilde t-1)} \sum_{i=1,i\neq k}^n |{\ell'}_{i}^{(\tilde t-1)} \la \bxi_k,\bxi_i \ra|\\
        \ge& \frac{\eta}{n} v_{j,r,2}^{(\tilde t-1)} \|\bxi_k\|_2^2 |{\ell'}_{k}^{(\tilde t-1)}| - \frac{\eta}{n} v_{j,r,2}^{(\tilde t-1)} C_6 |\ell'^{(\tilde t-1)}_k | \cdot 2\sigma_p^2 \sqrt{d \log(4n^2/\delta)}\\
        =&\frac{\eta}{n}v_{j,r,2}^{(\tilde t-1)}|{\ell'}_{k}^{(\tilde t-1)}|(\|\bxi_k\|_2^2 - 2 C_6 \sigma_p^2 \sqrt{d \log(4n^2/\delta)})\\
        >& 0.
    \end{align*}
    The first inequality is from triangle inequality; The second inequality is from the first induction hypothesis and Lemma \ref{lemma: noise bound}; The last inequality is from the second induction hypothesis $v_{j,r,2}^{(\tilde t-1)} \ge 0$ and Condition \ref{condition: main condition} about the definition of \(d\).

    So $\la\wb_{j,r}^{(t)},\bxi_k\ra > 0$ for $t = \tilde t$, then it follows that
    \begin{align*}
        S_k^{(\tilde t-1)} \subseteq S_k^{(\tilde t)}, S_{j,r}^{(\tilde t-1)} \subseteq S_{j,r}^{(\tilde t)}.
    \end{align*}
    And we have
    \begin{align*}
        S_k^{(0)} \subseteq S_k^{(\tilde t)}, S_{j,r}^{(0)} \subseteq S_{j,r}^{(\tilde t)}.
    \end{align*}
    
    Then we prove conclusion 4. Recall the update equation of $v_{y_k, r, 1}^{(t)}$ and $\la \wb_{y_k, r}^{(t)}, y_k \bmu \ra$:
    \begin{align*}
    \la\wb_{y_k,r}^{(t)},y_k \bmu\ra &= \la\wb_{j,r}^{(t-1)},y_k \bmu\ra - \frac{\eta}{n} v_{y_k,r,1}^{(t-1)} \|\bmu\|_2^2 \sum_{i=1}^n\ell'^{(t-1)}_i \mathbb{I}(\la\wb_{y_k,r}^{(t-1)},y_i \cdot \bmu\ra >0),\\
    v_{y_k,r,1}^{(t)} & = v_{y_k,r,1}^{(t-1)} - y_k \frac{\eta}{n}\sum_{i=1}^{n}\ell'^{(t-1)}_i \la \wb_{y_k,r}^{(t-1)},\bmu \ra \mathbb{I}(\la\wb_{y_k,r}^{(t-1)},y_i \cdot \bmu\ra >0).
    \end{align*}
    So we have the update equation of their ratio
    \begin{align*}
        \frac{v_{j,r,1}^{(t)}}{\la\wb_{y_k,r}^{(t)},y_k \bmu\ra} &= \frac{v_{y_k,r,1}^{(t-1)} - y_k \frac{\eta}{n}\sum_{i=1}^{n}\ell'^{(t-1)}_i \la \wb_{y_k,r}^{(t-1)},\bmu \ra \mathbb{I}(\la\wb_{y_k,r}^{(t-1)},y_i \cdot \bmu\ra >0)}{\la\wb_{j,r}^{(t-1)},y_k \bmu\ra - \frac{\eta}{n} v_{y_k,r,1}^{(t-1)} \|\bmu\|_2^2 \sum_{i=1}^n\ell'^{(t-1)}_i \mathbb{I}(\la\wb_{y_k,r}^{(t-1)},y_i \cdot \bmu\ra >0)}\\
        &= \frac{v_{y_k,r,1}^{(t-1)} - \frac{\eta}{n} \la \wb_{y_k, r}^{(t-1)}, y_k \bmu \ra \sum_{i=1}^n \ell'^{(t-1)}_i \mathbb{I}(\la\wb_{y_k,r}^{(t-1)},y_i \cdot \bmu\ra >0)}{\la\wb_{j,r}^{(t-1)},y_k \bmu\ra - \frac{\eta}{n} v_{y_k,r,1}^{(t-1)} \|\bmu\|_2^2 \sum_{i=1}^n\ell'^{(t-1)}_i \mathbb{I}(\la\wb_{y_k,r}^{(t-1)},y_i \cdot \bmu\ra >0)}\\
        &= \frac{\frac{v_{y_k,r,1}^{(t-1)}}{\la\wb_{j,r}^{(t-1)},y_k \bmu\ra} - \frac{\eta}{n} \sum_{i=1}^n \ell'^{(t-1)}_i \mathbb{I}(\la\wb_{y_k,r}^{(t-1)},y_i \cdot \bmu\ra >0)}{1 - \frac{\eta}{n}\frac{v_{y_k,r,1}^{(t-1)}}{\la\wb_{j,r}^{(t-1)},y_k \bmu\ra} \|\bmu\|_2^2 \sum_{i=1}^n\ell'^{(t-1)}_i \mathbb{I}(\la\wb_{y_k,r}^{(t-1)},y_i \cdot \bmu\ra >0)}\\
        &=  \frac{v_{y_k, r, 1}^{(t-1)}}{\la \wb_{y_k, r}^{(t-1)}, y_k\bmu \ra} \cdot \frac{1 - \frac{\la\wb_{j,r}^{(t-1)},y_k \bmu\ra}{v_{y_k,r,1}^{(t-1)}} \frac{\eta}{n} \sum_{i=1}^n \ell'^{(t-1)}_i \mathbb{I}(\la\wb_{y_k,r}^{(t-1)},y_i \cdot \bmu\ra >0)}{1 - \frac{\eta}{n}\frac{v_{y_k,r,1}^{(t-1)}}{\la\wb_{j,r}^{(t-1)},y_k \bmu\ra} \|\bmu\|_2^2 \sum_{i=1}^n\ell'^{(t-1)}_i \mathbb{I}(\la\wb_{y_k,r}^{(t-1)},y_i \cdot \bmu\ra >0)}.
    \end{align*}
    From the induction hypothesis, the conclusion holds at time t-1, so there exist $k_3, k_4 = \Theta(1)$ that $\frac{k_3}{\|\bmu\|_2} \le \frac{v_{y_k, r, 1}^{(t-1)}}{\la \wb_{y_k, r}^{(t-1)}, y_k\bmu \ra} \le \frac{k_4 \sigma_p \sqrt{d}}{\sqrt{n} \|\bmu\|_2^2}$ for all $k \in [n], r \in [m]$.
    
    First we could upper bound the ratio as
    \begin{align*}
        \frac{v_{j,r,1}^{(t)}}{\la\wb_{y_k,r}^{(t)},y_k \bmu\ra} &=  \frac{v_{y_k, r, 1}^{(t-1)}}{\la \wb_{y_k, r}^{(t-1)}, y_k\bmu \ra} \cdot \frac{1 - \frac{\la\wb_{j,r}^{(t-1)},y_k \bmu\ra}{v_{y_k,r,1}^{(t-1)}} \frac{\eta}{n} \sum_{i=1}^n \ell'^{(t-1)}_i \mathbb{I}(\la\wb_{y_k,r}^{(t-1)},y_i \cdot \bmu\ra >0)}{1 - \frac{\eta}{n}\frac{v_{y_k,r,1}^{(t-1)}}{\la\wb_{j,r}^{(t-1)},y_k \bmu\ra} \|\bmu\|_2^2 \sum_{i=1}^n\ell'^{(t-1)}_i \mathbb{I}(\la\wb_{y_k,r}^{(t-1)},y_i \cdot \bmu\ra >0)}\\
        &= \frac{v_{y_k, r, 1}^{(t-1)}}{\la \wb_{y_k, r}^{(t-1)}, y_k\bmu \ra} \cdot \frac{1 - \frac{\la\wb_{j,r}^{(t-1)},y_k \bmu\ra^2}{v_{y_k,r,1}^{(t-1)2} \|\bmu\|_2^2}   \frac{\eta}{n} \frac{v_{y_k,r,1}^{(t-1)}}{\la\wb_{j,r}^{(t-1)},y_k \bmu\ra} \|\bmu\|_2^2 \sum_{i=1}^n \ell'^{(t-1)}_i \mathbb{I}(\la\wb_{y_k,r}^{(t-1)},y_i \cdot \bmu\ra >0)}{1 - 
 \frac{\eta}{n} \frac{v_{y_k,r,1}^{(t-1)}}{\la\wb_{j,r}^{(t-1)},y_k \bmu\ra} \|\bmu\|_2^2 \sum_{i=1}^n \ell'^{(t-1)}_i \mathbb{I}(\la\wb_{y_k,r}^{(t-1)},y_i \cdot \bmu\ra >0)}\\
        &\le \frac{v_{y_k, r, 1}^{(t-1)}}{\la \wb_{y_k, r}^{(t-1)}, y_k\bmu \ra} \le \frac{k_4 \sigma_p \sqrt{d}}{\sqrt{n} \|\bmu\|_2^2}.
    \end{align*}
    The first inequality is from the induction hypothesis that $\frac{k_3}{\|\bmu\|_2} \le \frac{v_{y_k, r, 1}^{(t-1)}}{\la \wb_{y_k, r}^{(t-1)}, y_k\bmu \ra}$.
    
    Then to lower bound the ratio of two layers, note that $\frac{v_{y_k,r,1}^{(t)}}{\la\wb_{y_k,r}^{(t)},y_k \bmu\ra}$ will decrease only if 
    $$
    \frac{\la\wb_{y_k,r}^{(t-1)},y_k \bmu\ra}{v_{y_k,r,1}^{(t-1)}} \le \frac{v_{y_k,r,1}^{(t-1)}}{\la\wb_{y_k,r}^{(t-1)},y_k \bmu\ra} \|\bmu\|_2^2,
    $$
    which suffices to
    $$
    \frac{v_{y_k,r,1}^{(t-1)}}{\la\wb_{y_k,r}^{(t-1)},y_k \bmu\ra} \ge \frac{1}{\|\bmu\|_2^2}.
    $$
    Therefore, when $\frac{v_{y_k,r,1}^{(t-1)}}{\la\wb_{y_k,r}^{(t-1)},y_k \bmu\ra}$ decreases to the level of $\frac{1}{\|\bmu\|_2^2}$, the next iteration $\frac{v_{y_k,r,1}^{(t)}}{\la\wb_{y_k,r}^{(t)},y_k \bmu\ra}$ will not decrease and remain the same from then on, which implies $\frac{v_{y_k,r,1}^{(t)}}{\la\wb_{y_k,r}^{(t)},y_k \bmu\ra} \ge \frac{1}{\|\bmu\|_2^2}$.
    Therefore, conclusion 4 holds at time t.

    Finally we prove the last conclusion. Recall the update equation of $v_{y_k, r, 2}^{(t)}$ and $\la \wb_{y_k, r}^{(t)}, \bxi_k \ra$:
    \begin{align*}
 \la\wb_{y_k,r}^{(t)},\bxi_k\ra &= \la\wb_{j,r}^{(t-1)},\bxi_k\ra - \frac{\eta}{n} v_{j,r,2}^{(t-1)} \|\bxi_k\|_2^2 \ell'^{(t-1)}_{k} -\frac{y_k \eta}{n} v_{j,r,2}^{(t-1)} \sum_{i=1,i\neq k}^n y_i \ell'^{(t-1)}_i \la \bxi_k,\bxi_i \ra \mathbb{I}(\la\wb_{j,r}^{(t-1)},\bxi_i\ra >0),\\ 
    v_{y_k,r,2}^{(t)} & = v_{y_k,r,2}^{(t-1)} - \frac{y_k \eta}{n}\sum_{i=1}^{n}\ell'^{(t-1)}_iy_i\la \wb_{y_k,r}^{(t-1)},\bxi_i \ra \mathbb{I}(\la\wb_{y_k,r}^{(t-1)},\bxi_i\ra >0).
\end{align*}

So we have
\begin{align*}
    \frac{v_{y_k, r, 2}^{(t)}}{\la \wb_{y_k, r}^{(t)}, \bxi_k \ra} &= \frac{v_{y_k,r,2}^{(t-1)} - \frac{y_k \eta}{n}\sum_{i=1}^{n}\ell'^{(t-1)}_iy_i\la \wb_{y_k,r}^{(t-1)},\bxi_i \ra \mathbb{I}(\la\wb_{y_k,r}^{(t-1)},\bxi_i\ra >0)}{\la\wb_{y_k,r}^{(t-1)},\bxi_k\ra - \frac{\eta}{n} v_{y_k,r,2}^{(t-1)} \|\bxi_k\|_2^2 \ell'^{(t-1)}_{k} -\frac{y_k \eta}{n} v_{y_k,r,2}^{(t-1)} \sum_{i=1,i\neq k}^n y_i \ell'^{(t-1)}_i \la \bxi_k,\bxi_i \ra \mathbb{I}(\la\wb_{y_k,r}^{(t-1)},\bxi_i\ra >0)}\\
    &= \frac{\frac{v_{y_k, r, 2}^{(t-1)}}{\la \wb_{y_k, r}^{(t-1)}, \bxi_k \ra} - \frac{y_k \eta}{n}\sum_{i=1}^n \ell'^{(t-1)} y_i \frac{\la \wb_{y_k,r}^{(t-1)},\bxi_i \ra}{\la \wb_{y_k,r}^{(t-1)},\bxi_k \ra} \mathbb{I}(\la\wb_{y_k,r}^{(t-1)},\bxi_i\ra >0)}{1 - \frac{\eta}{n} \frac{v_{y_k, r, 2}^{(t-1)}}{\la \wb_{y_k, r}^{(t-1)}, \bxi_k \ra} \|\bxi_k\|^2 \ell'^{(t-1)}_k - \frac{y_k \eta}{n} \frac{v_{y_k, r, 2}^{(t-1)}}{\la \wb_{y_k, r}^{(t-1)}, \bxi_k \ra} \sum_{i=1,i\neq k}^n y_i \ell'^{(t-1)}_i \la \bxi_k,\bxi_i \ra \mathbb{I}(\la\wb_{y_k,r}^{(t-1)},\bxi_i\ra >0)}\\
    &= \frac{v_{y_k, r, 2}^{(t-1)}}{\la \wb_{y_k, r}^{(t-1)}, \bxi_k \ra} \cdot \frac{1 - \frac{\la \wb_{y_k, r}^{(t-1)}, \bxi_k \ra}{v_{y_k, r, 2}^{(t-1)}} \cdot \frac{y_k \eta}{n}\sum_{i=1}^n \ell'^{(t-1)} y_i \frac{\la \wb_{y_k,r}^{(t-1)},\bxi_i \ra}{\la \wb_{y_k,r}^{(t-1)},\bxi_k \ra} \mathbb{I}(\la\wb_{y_k,r}^{(t-1)},\bxi_i\ra >0)}{1 - \frac{\eta}{n} \frac{v_{y_k, r, 2}^{(t-1)}}{\la \wb_{y_k, r}^{(t-1)}, \bxi_k \ra} \|\bxi_k\|^2 \ell'^{(t-1)}_k - \frac{y_k \eta}{n} \frac{v_{y_k, r, 2}^{(t-1)}}{\la \wb_{y_k, r}^{(t-1)}, \bxi_k \ra} \sum_{i=1,i\neq k}^n y_i \ell'^{(t-1)}_i \la \bxi_k,\bxi_i \ra \mathbb{I}(\la\wb_{y_k,r}^{(t-1)},\bxi_i\ra >0)}.
\end{align*}

From the induction hypothesis, the conclusion holds at time $t-1$, so there exist $k_1, k_2 = \Theta(1)$ that $\frac{k_1\sqrt{n}}{\sigma_p \sqrt{d}}\le \frac{v_{y_k,r,2}^{(t-1)}}{\la \wb_{y_k,r}^{(t-1)}, \bxi_k \ra} \le \frac{k_2\sqrt{n}}{\sigma_p \sqrt{d}}$ for all $k \in [n], r \in [m]$. 

We first upper bound the ratio as
\begin{align*}
     \frac{v_{y_k, r, 2}^{(t)}}{\la \wb_{y_k, r}^{(t)}, \bxi_k \ra} &= \frac{v_{y_k, r, 2}^{(t-1)}}{\la \wb_{y_k, r}^{(t-1)}, \bxi_k \ra} \cdot \frac{1 - \frac{\la \wb_{y_k, r}^{(t-1)}, \bxi_k \ra}{v_{y_k, r, 2}^{(t-1)}} \cdot \frac{y_k \eta}{n}\sum_{i=1}^n \ell'^{(t-1)} y_i \frac{\la \wb_{y_k,r}^{(t-1)},\bxi_i \ra}{\la \wb_{y_k,r}^{(t-1)},\bxi_k \ra} \mathbb{I}(\la\wb_{y_k,r}^{(t-1)},\bxi_i\ra >0)}{1 - \frac{\eta}{n} \frac{v_{y_k, r, 2}^{(t-1)}}{\la \wb_{y_k, r}^{(t-1)}, \bxi_k \ra} \|\bxi_k\|^2 \ell'^{(t-1)}_k - \frac{y_k \eta}{n} \frac{v_{y_k, r, 2}^{(t-1)}}{\la \wb_{y_k, r}^{(t-1)}, \bxi_k \ra} \sum_{i=1,i\neq k}^n y_i \ell'^{(t-1)}_i \la \bxi_k,\bxi_i \ra \mathbb{I}(\la\wb_{y_k,r}^{(t-1)},\bxi_i\ra >0)}\\
     &\le \frac{v_{y_k, r, 2}^{(t-1)}}{\la \wb_{y_k, r}^{(t-1)}, \bxi_k \ra} \cdot \frac{1 - \frac{\eta}{n} \sum_{i=1}^n \ell'^{(t-1)}_i \frac{\la \wb_{y_k, r}^{(t-1)}, \bxi_i \ra}{v_{y_k, r, 2}^{(t-1)}} \mathbb{I}(\la\wb_{y_k,r}^{(t-1)},\bxi_i\ra >0)}{1 - \frac{\eta}{2n} \frac{v_{y_k, r, 2}^{(t-1)}}{\la \wb_{y_k, r}^{(t-1)}, \bxi_k \ra} \|\bxi_k\|^2 \ell'^{(t-1)}_k}\\
     &=\frac{v_{y_k, r, 2}^{(t-1)}}{\la \wb_{y_k, r}^{(t-1)}, \bxi_k \ra} \cdot \frac{1 - \frac{\eta}{2n} \frac{v_{y_k, r, 2}^{(t-1)}}{\la \wb_{y_k, r}^{(t-1)}, \bxi_k \ra} \|\bxi_k\|^2 \ell'^{(t-1)}_k \cdot \sum_{i=1}^n \frac{2 \ell'^{(t-1)}_i}{\ell'^{(t-1)}_k \|\bxi_k\|^2} \frac{\la \wb_{y_k, r}^{(t-1)}, \bxi_i \ra \la \wb_{y_k, r}^{(t-1)}, \bxi_k \ra}{v_{y_k, r, 2}^{(t-1)2}} \mathbb{I}(\la\wb_{y_k,r}^{(t-1)},\bxi_i\ra >0)}{1 - \frac{\eta}{2n} \frac{v_{y_k, r, 2}^{(t-1)}}{\la \wb_{y_k, r}^{(t-1)}, \bxi_k \ra} \|\bxi_k\|^2 \ell'^{(t-1)}_k}\\
     &\le \frac{v_{y_k, r, 2}^{(t-1)}}{\la \wb_{y_k, r}^{(t-1)}, \bxi_k \ra} \le  \frac{k_2\sqrt{n}}{\sigma_p \sqrt{d}}.
\end{align*}
The first inequality is from Condition \ref{condition: main condition} and the second inequality is from induction hypothesis that $\frac{k_1\sqrt{n}}{\sigma_p \sqrt{d}}\le \frac{v_{y_k,r,2}^{(t-1)}}{\la \wb_{y_k,r}^{(t-1)}, \bxi_k \ra}$.

Then to lower bound the ratio of two layers, we have
\begin{align*}
    \frac{v_{y_k, r, 2}^{(t)}}{\la \wb_{y_k, r}^{(t)}, \bxi_k \ra} &= \frac{v_{y_k, r, 2}^{(t-1)}}{\la \wb_{y_k, r}^{(t-1)}, \bxi_k \ra} \cdot \frac{1 - \frac{\eta}{n} \sum_{i=1}^n \ell'^{(t-1)}_i \frac{\la \wb_{y_k, r}^{(t-1)}, \bxi_i \ra}{v_{y_k, r, 2}^{(t-1)}} \mathbb{I}(\la\wb_{y_k,r}^{(t-1)},\bxi_i\ra >0)}{1 - \Theta(1) \frac{\eta}{n} \frac{v_{y_k, r, 2}^{(t-1)}}{\la \wb_{y_k, r}^{(t-1)}, \bxi_k \ra} \|\bxi_k\|^2 \ell'^{(t-1)}_k},
\end{align*}
So $\frac{v_{y_k, r, 2}^{(t)}}{\la \wb_{y_k, r}^{(t)}, \bxi_k \ra}$ will decrease only if 
\begin{align*}
    \frac{\eta}{n} \sum_{i=1}^n \ell'^{(t-1)}_i \frac{\la \wb_{y_k, r}^{(t-1)}, \bxi_i \ra}{v_{y_k, r, 2}^{(t-1)}} \mathbb{I}(\la\wb_{y_k,r}^{(t-1)},\bxi_i\ra >0) \le \Theta(1) \frac{\eta}{n} \frac{v_{y_k, r, 2}^{(t-1)}}{\la \wb_{y_k, r}^{(t-1)}, \bxi_k \ra} \|\bxi_k\|^2 \ell'^{(t-1)}_k,
\end{align*}
which suffices to
\begin{align*}
    \frac{v_{y_k, r, 2}^{(t)}}{\la \wb_{y_k, r}^{(t)}, \bxi_k \ra} \ge \Theta\bigg(\frac{\sqrt{n}}{\sigma_p \sqrt{d}}\bigg).
\end{align*}
Therefore, when $\frac{v_{y_k, r, 2}^{(t-1)}}{\la \wb_{y_k, r}^{(t-1)}, \bxi_k \ra}$ decreases to the level of $\frac{\sqrt{n}}{\sigma_p \sqrt{d}}$, in the next iteration $\frac{v_{y_k, r, 2}^{(t)}}{\la \wb_{y_k, r}^{(t)}, \bxi_k \ra}$ will be non-decreasing and remain the same level from then on. So there exists $k_1 = \Theta(1)$ that $\frac{v_{y_k, r, 2}^{(t)}}{\la \wb_{y_k, r}^{(t)}, \bxi_k \ra} \ge \frac{k_1 \sqrt{n}}{\sigma_p \sqrt{d}}$. Therefore, all conclusions hold at time t.
\end{proof}

\begin{proof}[Proof of Proposition \ref{prop: stage2_analysis_c2}]

Now we could prove Proposition \ref{prop: stage2_analysis_c2}. The results are obvious at $t=T_1$ from Proposition \ref{prop: stage1_s}. Suppose that there exists $\tilde T \le T^*$ such that all the results in Proposition \ref{prop: stage2_analysis_c2} hold for all $T_1 \le t \le \tilde T-1$. We could prove that these also hold at $t=\tilde T$.

We prove (\ref{eq: neg_noise_bound_c2}) first. We consider two cases. When $\underline{\rho}_{j,r,k}^{(t-1)} \le -2\sqrt{\log(8mn/\delta)}\cdot \sigma_0 \sigma_p\sqrt{d} - 4 \sqrt{\frac{\log(4n^2/\delta)}{d}} n\alpha$, we have
\begin{align*}
    \la \wb_{j,r}^{(\tilde T-1)}, \bxi_k \ra \le \la \wb_{j,r}^{(0)}, \bxi_k \ra + \underline{\rho}_{j,r,k}^{(\tilde T-1)} + 4 \sqrt{\frac{\log(4n^2/\delta)}{d}} n\alpha \le 0.
\end{align*}
Therefore,
\begin{align*}
    \underline{\rho}_{j,r,k}^{(\tilde T)} &= \underline{\rho}_{j,r,k}^{(\tilde T-1)} + \frac{\eta}{nm} \cdot \ell'^{(\tilde T-1)}_k v_{j,r,2}^{(\tilde T-1)} \cdot \mathbb{I}(\la \wb_{j,r}^{(\tilde T-1)}, \bxi_k \ra \ge 0) \cdot \mathbb{I}(y_k = -j) \|\bxi_k\|_2^2\\
    &= \underline{\rho}_{j,r,k}^{(\tilde T-1)}\\
    &\ge -2\sqrt{\log(8mn/\delta)}\cdot \sigma_0 \sigma_p\sqrt{d} - 8 \sqrt{\frac{\log(4n^2/\delta)}{d}} n\alpha.
\end{align*}
The last inequity is from induction hypothesis. On the other hand, when $\underline{\rho}_{j,r,k}^{(t-1)} > -2\sqrt{\log(8mn/\delta)}\cdot \sigma_0 \sigma_p\sqrt{d} - 4 \sqrt{\frac{\log(4n^2/\delta)}{d}} n\alpha$, we first prove that $v_{j,r,2}^{(\tilde T-1)}$ is upper bounded. From Proposition \ref{prop: stage1_s} we know that $v^{(t)}_{j,r,2} /\la\wb_{j,r}^{(t)},\bxi_k\ra = \Theta(\sqrt{n} / \sigma_p \sqrt{d})$ for $t \ge T_1$. Assume that $\frac{k_1 \sqrt{n}}{\sigma_p \sqrt{d}} \le  \frac{v^{(t)}_{j,r,2}}{\la\wb_{j,r}^{(t)},\bxi_k\ra} \le \frac{k_2 \sqrt{n}}{\sigma_p \sqrt{d}}$ for all $j \in \{\pm 1\}$ and $k \in [n]$, and thus
\begin{align}
    v_{j,r,2}^{(\tilde T-1)} &\le \frac{k_2 \sqrt{n}}{\sigma_p \sqrt{d}} \cdot \la\wb_{j,r}^{(\tilde T-1)},\bxi_k\ra\nonumber\\
    &\le \frac{k_2 \sqrt{n}}{\sigma_p \sqrt{d}} \cdot \bigg[\la\wb_{j,r}^{(0)},\bxi_k\ra + \rho_{j,r,k}^{(\tilde T-1)} + 4 \sqrt{\frac{\log(4n^2/\delta)}{d}} n\alpha \bigg]\nonumber\\
    &\le \frac{2\alpha k_2 \sqrt{n}}{\sigma_p \sqrt{d}}. \label{eq: v_bound_s}
\end{align}

So we have
\begin{align*}
     \underline{\rho}_{j,r,k}^{(\tilde T)} &= \underline{\rho}_{j,r,k}^{(\tilde T-1)} + \frac{\eta}{nm} \cdot \ell'^{(\tilde T-1)}_k v_{j,r,2}^{(\tilde T-1)} \cdot \mathbb{I}(\la \wb_{j,r}^{(\tilde T-1)}, \bxi_k \ra \ge 0) \cdot \mathbb{I}(y_k = j) \|\bxi_k\|_2^2\\
     &\ge -2\sqrt{\log(8mn/\delta)}\cdot \sigma_0 \sigma_p\sqrt{d} - 4 \sqrt{\frac{\log(4n^2/\delta)}{d}} n\alpha - \frac{3\eta \alpha k_2 \sigma_p \sqrt{d}}{\sqrt{n} m}\\
     &\ge -2\sqrt{\log(8mn/\delta)}\cdot \sigma_0 \sigma_p\sqrt{d} - 8 \sqrt{\frac{\log(4n^2/\delta)}{d}} n\alpha,
\end{align*}
where the first inequity is from Lemma \ref{lemma: noise bound} and the upper bound of $v_{j,r,2}^{(\tilde T-1)}$. The last inequity is due to the definition of $\eta$ in Condition \ref{condition: main condition}.

Next we prove (\ref{eq: pos_noise_bound_c2}) holds for $t = \tilde T$. Consider the loss derivative
\begin{align}
    |\ell'^{(t)}_k| &= \frac{1}{1 + \exp\{y_k \cdot [(F_{+1}(\bW^{(t)}_{+1}, \bV^{(t)}_{+1}; \bx_k) - F_{-1}(\bW^{(t)}_{-1}, \bV^{(t)}_{-1}; \bx_k)]\}}\nonumber\\
    &\le \exp\{-y_k \cdot [(F_{+1}(\bW^{(t)}_{+1}, \bV^{(t)}_{+1}; \bx_k) - F_{-1}(\bW^{(t)}_{-1}, \bV^{(t)}_{-1}; \bx_k)]\}\nonumber\\
    &\le \exp\{-F_{y_k}(\bW^{(t)}_{y_k}, \bV^{(t)}_{y_k}; \bx_k) + 0.5\}.\label{eq: loss_bound}
\end{align}
The last inequity is from Lemma \ref{lemma: output_leading_term_c2}. Then recall the update equation of $\overline{\rho}_{j,r,k}^{(t)}$:
\begin{align*}
    \overline{\rho}_{j,r,k}^{(t+1)} = \overline{\rho}_{j,r,k}^{(t)} - \frac{\eta}{nm} \cdot \ell'^{(t)}_k v_{j,r,2}^{(t)} \cdot \mathbb{I}(\la \wb_{j,r}^{(t)}, \bxi_k \ra \ge 0) \cdot \mathbb{I}(y_k = j) \|\bxi_k\|_2^2.
\end{align*}
Now assume $t_{j,r,k}$ is the last time $t \le T^*$ when $\overline{\rho}_{j,r,k}^{(t)} \le 0.5\alpha$. Then we have
\begin{align*}
    \overline{\rho}_{j,r,k}^{(\tilde T)} =& \overline{\rho}_{j,r,k}^{(t_{j,r,k})} - \underbrace{\frac{\eta}{nm} \cdot \ell'^{(t_{j,r,k})}_k v_{j,r,2}^{(t_{j,r,k})} \cdot \mathbb{I}(\la \wb_{j,r}^{(t_{j,r,k})}, \bxi_k \ra \ge 0) \cdot \mathbb{I}(y_k = j) \|\bxi_k\|_2^2}_{I_3}\\
    &-\underbrace{ \sum\limits_{t_{j,r,k} < t <\tilde T} \frac{\eta}{nm} \cdot \ell'^{(t)}_k v_{j,r,2}^{(t)} \cdot \mathbb{I}(\la \wb_{j,r}^{(t)}, \bxi_k \ra \ge 0) \cdot \mathbb{I}(y_k = j) \|\bxi_k\|_2^2}_{I_4}.
\end{align*}
The idea is that we will upper bound $|I_3| \le 0.25\alpha$ and $|I_4| \le 0.25\alpha$. So $\overline{\rho}_{j,r,k}^{(\tilde T)} \le \alpha$ holds. We first bound $I_3$:
\begin{align*}
    |I_3| &\le \frac{\eta}{nm} \cdot v_{j,r,2}^{(t_{j,r,k})} \|\bxi_k\|_2^2\\
    &\le \frac{\eta}{nm} \cdot \frac{2\alpha k_2 \sqrt{n}}{\sigma_p \sqrt{d}} \cdot \frac{3\sigma_p^2d}{2}\\
    &\le 0.25\alpha,
\end{align*}
where the second inequity is from Lemma \ref{lemma: noise bound} and (\ref{eq: v_bound_s}). The last inequity is by the definition of $\eta$ in Condition \ref{condition: main condition}.

Then we bound $I_4$. For $t_{j,r,k} < t <\tilde T$ and $y_k = j$, we can lower bound $\la \wb_{j,r}^{(t)}, \bxi_k \ra$ as follows:
\begin{align*}
    \la \wb_{j,r}^{(t)}, \bxi_k \ra &\ge \la \wb_{j,r}^{(0)}, \bxi_k \ra + \overline{\rho}_{j,r,k}^{(t)} - 4 \sqrt{\frac{\log(4n^2/\delta)}{d}} n\alpha\\
    &\ge -2\sqrt{\log(8mn/\delta)}\cdot \sigma_0 \sigma_p\sqrt{d} + 0.5\alpha - 4 \sqrt{\frac{\log(4n^2/\delta)}{d}} n\alpha\\
    &\ge 0.25\alpha,
\end{align*}
where the second inequity is due to $\overline{\rho}_{j,r,k}^{(t)} > 0.5\alpha$ for $t > t_{j,r,k}$. Remember that from Proposition \ref{prop: stage1_s} we have $\frac{k_1 \sqrt{n}}{\sigma_p \sqrt{d}} \le  \frac{v^{(t)}_{j,r,2}}{\la\wb_{j,r}^{(t)},\bxi_k\ra}$ for $k_1 = \Theta(1)$. So we could also lower bound $v_{j,r,2}^{(t)}$ as
\begin{align}
    v_{j,r,2}^{(t)} \ge \frac{k_1 \sqrt{n}}{\sigma_p \sqrt{d}} \cdot \la\wb_{j,r}^{(t)},\bxi_k\ra \ge \frac{k_1 \sqrt{n} \alpha}{4\sigma_p \sqrt{d}}.\label{eq: v_lower_bound}
\end{align}

Therefore, we have
\begin{align*}
    |I_4| &\le \sum\limits_{t_{j,r,k} < t <\tilde T} \frac{\eta}{nm} \cdot \exp\bigg(-\sum_{r=1}^m v_{j,r,2}^{(t)} \sigma(\la \wb_{j,r}^{(t)}, \bxi_k \ra ) + 0.5\bigg) \cdot v_{j,r,2}^{(t)} \|\bxi_k\|_2^2\\
    &\le \frac{2 \eta (\tilde T - t_{j,r,k} + 1)}{nm} \cdot \exp\bigg(-\frac{k_1 m \sqrt{n} \alpha^2}{16 \sigma_p \sqrt{d}} \bigg) \cdot \frac{k_1 \sqrt{n} \alpha}{4\sigma_p \sqrt{d}} \cdot \frac{3\sigma_p^2 d}{2}\\
    &\le \frac{2\eta \tilde T}{nm} \cdot \exp(-\log(T^*)) \cdot \frac{k_1 \sqrt{n} \alpha}{4\sigma_p \sqrt{d}} \cdot \frac{3\sigma_p^2 d}{2}\\
    &= \frac{2\eta}{nm} \cdot \frac{k_1 \sqrt{n} \alpha}{4\sigma_p \sqrt{d}} \cdot \frac{3\sigma_p^2 d}{2}\\
    &= \frac{3\eta k_1 \sigma_p \sqrt{d} \alpha}{4\sqrt{n}m}\\
    &\le 0.25\alpha,
\end{align*}
where the first inequity is from (\ref{eq: loss_bound}); the second inequity is from the lower bound of $\la \wb_{j,r}^{(t)}, \bxi_k \ra$ and $v_{j,r,2}^{(t)}$; the third inequity is by $\alpha = 4\sqrt{\frac{\sigma_p \sqrt{d} \log(T^*)}{k_1 m \sqrt{n}}}$ and the last inequity is from the definition of $\eta$ in Condition \ref{condition: main condition}.

Combining the bound of $I_3$ and $I_4$, we have $\overline{\rho}_{j,r,k}^{(t)} \le \alpha$ at $t = \tilde T$.

Then we prove (\ref{eq: signal_output_bound_c2}) holds for $t = \tilde T$. Recall the update equation of $\gamma_{j,r}^{(t)}$
\begin{align*}
    \gamma_{j,r}^{(t+1)} = \gamma_{j,r}^{(t)} - \frac{\eta}{nm} \sum_{i=1}^n \ell'^{(t)}_i v_{j,r,1}^{(t)} \|\bmu\|_2^2 \mathbb{I}(\la \wb_{j,r}^{(t)}, y_i \bmu \ra).
\end{align*}
We first prove that $\gamma_{j,r}^{(\tilde T)} \ge \gamma_{j,r}^{(\tilde T-1)}$, and hence $\gamma_{j,r}^{(\tilde T)} \ge \gamma_{j,r}^{(0)} = 0$ for any $j \in \{\pm 1\}, r \in [m]$. From Proposition \ref{prop: stage1_s}, we know that $v_{j,r,1}^{(t)}$ will remain at its initial order throughout stage 1. And recall the update equation of $v_{j,r,1}^{(t)}$:
\begin{align*}
    v_{j,r,1}^{(t+1)} & = v_{j,r,1}^{(t)} - j \frac{\eta}{n}\sum_{i=1}^{n}\ell'^{(t)}_i \la \wb_{j,r}^{(t)},\bmu \ra \mathbb{I}(\la\wb_{j,r}^{(t)},y_i \cdot \bmu\ra >0).
\end{align*}
So $v_{j,r,1}^{(t)}$ will keep increasing when $\gamma_{j,r}^{(t-1)} = \la \wb_{j,r}^{(t-1)}, j \cdot \bmu \ra \ge 0$. This holds due to the inductive hypothesis. Therefore, we have $v_{j,r,1}^{(t)} \ge 0$, so it follows that $\gamma_{j,r}^{(\tilde T)} \ge \gamma_{j,r}^{(\tilde T-1)}$.

For the other hand of (\ref{eq: signal_output_bound_c2}), we prove a strengthened hypothesis that there exists a $i^* \in [n]$ with $y_{i^*} = j$ such that for $1 \le t \le T^*$ we have that
\begin{align*}
    \frac{\gamma_{j,r}^{(t)}}{\overline{\rho}_{j,r,i^*}^{(t)}} \le \frac{C_8\sqrt{n}\|\bmu\|_2^2 v_0}{\alpha \sigma_p \sqrt{d}}.
\end{align*}
And $C_8 = \Theta(1)$, $i^*$ can be taken as any sample in the training set.

Recall the update equation of $\gamma_{j,r}^{(t)}$ and $\overline{\rho}_{j,r,i}^{(t)}$:
\begin{align*}
    \gamma_{j,r}^{(\tilde T)} &= \gamma_{j,r}^{(\tilde T-1)} - \frac{\eta}{nm} \sum_{i=1}^n \ell'^{(\tilde T-1)}_i v_{j,r,1}^{(\tilde T-1)} \|\bmu\|_2^2 \mathbb{I}(\la \wb_{j,r}^{(\tilde T-1)}, y_i \bmu \ra \ge 0),\\
    \overline{\rho}_{j,r,k}^{(\tilde T)} &= \overline{\rho}_{j,r,k}^{(\tilde T-1)} - \frac{\eta}{nm} \cdot \ell'^{(\tilde T-1)}_k v_{j,r,2}^{(\tilde T-1)} \cdot \mathbb{I}(\la \wb_{j,r}^{(\tilde T-1)}, \bxi_k \ra \ge 0) \cdot \mathbb{I}(y_k = j) \|\bxi_k\|_2^2.
\end{align*}
From Lemma \ref{lemma: key_lemma_stage2_c2}, for any $i^* \in S_{j,r}^{(0)}$, it holds that $i^* \in S_{j,r}^{(t)}$ for any $1 \le t \le \tilde T-1$. So we have
\begin{align*}
    \overline{\rho}_{j,r,i^*}^{(\tilde T)} &= \overline{\rho}_{j,r,i^*}^{(\tilde T-1)} - \frac{\eta}{nm} \cdot \ell'^{(\tilde T-1)}_{i^*} \cdot v_{j,r,2}^{(\tilde T-1)} \cdot \|\bxi_{i^*}\|^2  
    \ge \overline{\rho}_{j,r,i^*}^{(\tilde T-1)} - \frac{\eta}{nm} \cdot \ell'^{(\tilde T-1)}_{i^*} \cdot v_{j,r,2}^{(\tilde T-1)} \cdot \sigma_p^2 d/2.
\end{align*}
For $\gamma_{j,r}^{(\tilde T)}$ we have $y_{i^*} = j$ that
\begin{align*}
    \gamma_{j,r}^{(\tilde T)} &\le \gamma_{j,r}^{(\tilde T-1)} - \frac{\eta}{nm} \sum_{i=1}^n |\ell'^{(\tilde T-1)}_i| v_{j,r,1}^{(\tilde T-1)} \|\bmu\|_2^2 \cdot \mathbb{I}(y_i = j) \le \gamma_{j,r}^{(\tilde T-1)} - \frac{\eta}{m} C_6 \ell'^{(\tilde T-1)}_{i^*} v_{j,r,1}^{(\tilde T-1)} \|\bmu\|_2^2
\end{align*}
where the first inequality is from triangle inequality and the second is from Lemma \ref{lemma: key_lemma_stage2_c2}. 
Then from the third conclusion in Lemma \ref{lemma: key_lemma_stage2_c2}, there exist $k_3, k_4 = \Theta(1)$ that $\frac{k_3}{\|\bmu\|_2} \le \frac{v_{j, r, 1}^{(t)}}{\la \wb_{j, r}^{(t)}, j\bmu \ra} \le \frac{k_4 \sigma_p \sqrt{d}}{\sqrt{n} \|\bmu\|_2^2}$ for all $j \in \{\pm 1\}, r \in [m]$. So we have
\begin{align*}
    \frac{\gamma_{j,r}^{(\tilde T)}}{\overline{\rho}_{j,r,i^*}^{(\tilde T)}} &\le \max\Bigg\{\frac{\gamma_{j,r}^{(\tilde T-1)}}{\overline{\rho}_{j,r,i^*}^{(\tilde T-1)}}, \frac{ C_6 n v_{j,r,1}^{(\tilde T-1)} \|\bmu\|_2^2}{v_{j,r,2}^{(\tilde T-1)} \sigma_p^2 d/2} \Bigg\}\\ 
    &\le \max\Bigg\{\frac{\gamma_{j,r}^{(\tilde T-1)}}{\overline{\rho}_{j,r,i^*}^{(\tilde T-1)}}, \frac{ C_6 n  \frac{k_4 \sigma_p \sqrt{d}}{\sqrt{n}  \|\bmu\|_2^2} |\la \wb_{j, r}^{(t)}, j\bmu \ra| \|\bmu\|_2^2}{\frac{k_1 \sqrt{n}}{\sigma_p \sqrt{d}} \overline{\rho}_{j,r,i^*}^{(\tilde T-1)} \sigma_p^2 d/2} \Bigg\}\\
    &= \max\Bigg\{\frac{\gamma_{j,r}^{(\tilde T-1)}}{\overline{\rho}_{j,r,i^*}^{(\tilde T-1)}}, \frac{C_6 k_4 \gamma_{j,r}^{(\tilde T-1)}}{\frac{k_1}{2} \overline{\rho}_{j,r,i^*}^{(\tilde T-1)}} \Bigg\}\\
    &= \frac{\gamma_{j,r}^{(\tilde T-1)}}{\overline{\rho}_{j,r,i^*}^{(\tilde T-1)}}\\
    &\le \frac{ C_8 \sqrt{n}  v_0 \|\bmu\|_2^2}{\sigma_p \sqrt{d}}.
\end{align*}
The second inequality is due to (\ref{eq: v_lower_bound}) and the last inequality is from the induction hypothesis.

Last we prove (\ref{eq: v1_bound_c2}). From the third conclusion of Lemma \ref{lemma: key_lemma_stage2_c2}, we have $\Theta \Big(\frac{1}{\|\bmu\|_2}\Big) \le \frac{v_{j, r, 1}^{(t)}}{\la \wb_{j, r}^{(t)}, j\bmu \ra} \le \Theta\Big(\frac{\sigma_p \sqrt{d}}{\sqrt{n} \|\bmu\|_2^2}\Big)$ for all $j \in \{\pm 1\}, r \in [m]$. Meanwhile, we can upper bound $\la \wb_{j, r}^{(t)}, j\bmu \ra$ as
\begin{align*}
    \la \wb_{j, r}^{(t)}, j\bmu \ra &= \la \wb_{j, r}^{(0)}, j\bmu \ra + \gamma_{j, r}^{(t)}\\
    &\le \sqrt{2\log(8m/\delta)} \cdot \sigma_0\|\bmu\|_2 + \gamma_{j, r}^{(t)}\\
    &\le \frac{ C_5 \sqrt{n}  v_0 \|\bmu\|_2^2 \alpha}{\sigma_p \sqrt{d}}.
\end{align*}
The first inequality is from Lemma \ref{lemma: noise bound} and the second inequality is from the upper bound of $\gamma_{j,r}^{(t)}$ we just derived. So we could upper bound $v_{j,r,1}^{(t)}$ as 
\begin{align*}
    v_{j,r,1}^{(t)} &\le \la \wb_{j, r}^{(t)}, j\bmu \ra \cdot \Theta\Big(\frac{\sigma_p \sqrt{d}}{\sqrt{n} \|\bmu\|_2^2}\Big)\\
    &\le \frac{ C_5 \sqrt{n}  v_0 \|\bmu\|_2^2 \alpha}{\sigma_p \sqrt{d}}\cdot \Theta\Big(\frac{\sigma_p \sqrt{d}}{\sqrt{n} \|\bmu\|_2^2}\Big)\\
    &\le C_4 v_0 \alpha.
\end{align*}
Therefore, we complete the proof of all four conclusions at time t.
\end{proof}

Then we prove that the training loss can be arbitrarily small in a long time. Same as above, we have the following lemma to lower bound the output of training sample:
\begin{lemma}
\label{lemma: output_lower_c2}
    Under Condition \ref{condition: main condition}, for $T^{*, 1} \le t \le T^*$, we have
    \begin{align*}
        \sum_{r=1}^m v_{y_k, r, 2}^{(\tilde t)} \sigma(\la \wb_{y_k, r}^{(\tilde t)}, \bxi_i \ra) \ge \log(M t)
    \end{align*}
    for all $k \in [n]$. Here $M = \frac{1}{2} e^{-0.25} \eta (\frac{k_1 \|\bxi_k\|^2}{\sqrt{n}\sigma_p \sqrt{d}} + \frac{\eta \sigma_p \sqrt{d}}{C_6 k_2 \sqrt{n}}\cdot (|S_{y_k, r}^{(\tilde t-1)}|/n)) = \Theta(\frac{\eta \sigma_p \sqrt{d}}{\sqrt{n}})$.
\end{lemma}

\begin{proof}[Proof of Lemma \ref{lemma: output_lower_c2}]
    We use induction to prove this lemma. First from the analysis in stage 1, the conclusion holds at time $T^{*, 1}$. Then assume that the conclusion holds at time $\tilde t-1$, we consider the time $\tilde t$.

    We can lower bound the loss derivative as
\begin{align}
    |\ell'^{(\tilde t-1)}_k| &= \frac{1}{1 + \exp\{y_k \cdot [(F_{+1}(\bW^{(\tilde t-1)}_{+1}, \bV^{(\tilde t-1)}_{+1}; \bx_k) - F_{-1}(\bW^{(t)}_{-1}, \bV^{(t)}_{-1}; \bx_k)]\}}\nonumber\\
    &\ge \frac{1}{2} \exp(-F_{y_k}(\bW^{(\tilde t-1)}_{y_k}, \bV^{(\tilde t-1)}_{y_k}; \bx_k))\nonumber\\
    &\ge \frac{1}{2} e^{-0.25} \exp(-\sum_{r=1}^m v_{y_k, r, 2}^{(\tilde t-1)} \sigma(\la \wb_{y_k, r}^{(\tilde t-1)}, \bxi_k \ra)) .\label{eq: loss_lower_bound_c1_s}
\end{align}
The first inequality is from the fact that $F_{y_k}(\bW^{(t)}_{y_k}, \bV^{(t)}_{y_k}; \bx_k) \ge F_{-y_k}(\bW^{(t)}_{-y_k}, \bV^{(t)}_{-y_k}; \bx_k)$ and the second inequality is from Lemma \ref{lemma: output_leading_term_c2}.

Recall the update equation of $v_{y_k, r, 2}^{(t)} \sigma(\la \wb_{y_k, r}^{(t)}, \bxi_k \ra)$:
    \begin{align*}
        v_{y_k, r, 2}^{(\tilde t)} \sigma(\la \wb_{y_k, r}^{(\tilde t)}, \bxi_k \ra) =& v_{y_k, r, 2}^{(\tilde t-1)} \sigma(\la \wb_{y_k, r}^{(\tilde t-1)}, \bxi_k \ra) - \frac{\eta}{n} v_{y_k, r, 2}^{(\tilde t-1)2} \|\bxi_k\|_2^2 \ell'^{(\tilde t-1)}_k\\
        &- \frac{y_k \eta}{n} \la \wb_{y_k, r}^{(\tilde t-1)}, \bxi_k \ra \sum_{i=1}^n \ell'^{(\tilde t-1)}_i y_k \la \wb_{y_k, r}^{(\tilde t-1)}, \bxi_i \ra \mathbb{I}(\la \wb_{y_k, r}^{(\tilde t-1)}, \bxi_i \ra > 0)\\
        &+ \frac{\eta^2}{n^2} v_{y_k, r, 2}^{(\tilde t-1)} \sum_{i=1, i\neq k}^n y_k \ell'^{(\tilde t-1)}_i \la \bxi_k, \bxi_i \ra \mathbb{I}(\la \wb_{y_k, r}^{(\tilde t-1)}, \bxi_i \ra \ge 0) \sum_{i=1}^n \ell'^{(\tilde t-1)}_i y_k \la \wb_{y_k, r}^{(\tilde t-1)}, \bxi_i \ra \mathbb{I}(\la \wb_{y_k, r}^{(\tilde t-1)}, \bxi_i \ra \ge 0).
    \end{align*}
    
From the analysis in Lemma \ref{lemma: key_lemma_stage2_c1}, We could assume that $\frac{k_1 \sqrt{n}}{\sigma_p \sqrt{d}} \le  \frac{v^{(t)}_{j,r,2}}{\la\wb_{j,r}^{(t)},\bxi_k\ra} \le \frac{k_2 \sqrt{n}}{\sigma_p \sqrt{d}}$ for all $j = y_k$, $k \in [n]$ and lower bound $v_{y_k, r, 2}^{(\tilde t)} \sigma(\la \wb_{y_k, r}^{(\tilde t)}, \bxi_i \ra)$ that
    \begin{align*}
        v_{y_k, r, 2}^{(\tilde t)} &\sigma(\la \wb_{y_k, r}^{(\tilde t)}, \bxi_i \ra) \\
        \ge& v_{y_k, r, 2}^{(\tilde t-1)} \sigma(\la \wb_{y_k, r}^{(\tilde t-1)}, \bxi_i \ra) - 
\frac{k_1 \eta }{\sqrt{n} \sigma_p \sqrt{d}} v_{y_k, r, 2}^{(\tilde t-1)} \la \wb_{y_k, r}^{(\tilde t -1)}, \bxi_i \ra \|\bxi_i\|_2^2 \ell'^{(\tilde t-1)}_i\\
&- \frac{\eta \sigma_p \sqrt{d}}{C_6 k_2 \sqrt{n}} v_{y_k, r, 2}^{(\tilde t-1)} \la \wb_{y_k, r}^{(\tilde t -1)}, \bxi_i \ra \ell'^{(\tilde t-1)}_k \cdot (|S_{y_k, r}^{(\tilde t-1)}|/n))\\
=& v_{y_k, r, 2}^{(\tilde t-1)} \sigma(\la \wb_{y_k, r}^{(\tilde t-1)}, \bxi_i \ra) - v_{y_k, r, 2}^{(\tilde t-1)} \sigma(\la \wb_{y_k, r}^{(\tilde t-1)}, \bxi_i \ra) \cdot \eta \ell'^{(\tilde t-1)}_k (\frac{k_1 \|\bxi_k\|^2}{\sqrt{n}\sigma_p \sqrt{d}} +  \frac{\eta \sigma_p \sqrt{d}}{C_6 k_2 \sqrt{n}} \cdot (|S_{y_k, r}^{(\tilde t-1)}|/n)).
    \end{align*}
    The inequality is from the definition of $k_1, k_2$ and $\ell'^{(t)}_k \le 0$. Sum up from $r=1$ to m, we have
    \begin{align*}
        \sum_{r=1}^m v_{y_k, r, 2}^{(\tilde t)}& \sigma(\la \wb_{y_k, r}^{(\tilde t)}, \bxi_i \ra)\\
        &\ge \sum_{r=1}^m v_{y_k, r, 2}^{(\tilde t-1)} \sigma(\la \wb_{y_k, r}^{(\tilde t-1)}, \bxi_i \ra) - v_{y_k, r, 2}^{(\tilde t-1)} \sigma(\la \wb_{y_k, r}^{(\tilde t-1)}, \bxi_i \ra) \cdot \eta \ell'^{(\tilde t-1)}_k (\frac{k_1 \|\bxi_k\|^2}{\sqrt{n}\sigma_p \sqrt{d}} + \frac{\eta \sigma_p \sqrt{d}}{C_6 k_2 \sqrt{n}} \cdot (|S_{y_k, r}^{(\tilde t-1)}|/n))\\
        &\ge  \sum_{r=1}^m v_{y_k, r, 2}^{(\tilde t-1)} \sigma(\la \wb_{y_k, r}^{(\tilde t-1)}, \bxi_i \ra) + M v_{y_k, r, 2}^{(\tilde t-1)} \sigma(\la \wb_{y_k, r}^{(\tilde t-1)}, \bxi_i \ra) \cdot \exp(-\sum_{r=1}^m v_{y_k, r, 2}^{(\tilde t-1)} \sigma(\la \wb_{y_k, r}^{(\tilde t-1)}, \bxi_k \ra))\\ 
        &\ge \log(M(\tilde t-1)) + \frac{M \log(\tilde t - 1)}{\tilde t -1}\\
        &\ge \log(M \tilde t).
    \end{align*}
    In the second inequality we define $M = \frac{1}{2} e^{-0.25} \eta (\frac{k_1 \|\bxi_k\|^2}{\sqrt{n}\sigma_p \sqrt{d}} + \frac{\eta \sigma_p \sqrt{d}}{C_6 k_2 \sqrt{n}} \cdot (|S_{y_k, r}^{(\tilde t-1)}|/n))$; The third inequality is from our induction hypothesis $\sum_{r=1}^m v_{y_k, r, 2}^{(\tilde t-1)} \sigma(\la \wb_{y_k, r}^{(\tilde t-1)}, \bxi_i \ra) \ge \log(M(\tilde t-1))$ and $f(x) = \frac{x}{e^x}$ is monotonically decreasing when $x \ge 0$; The last inequality is because $f(x) = \frac{\log(x)}{x}$ is monotonically decreasing when $x \ge e$.
    
\end{proof}

With the lower bound of the output, we can prove that the training loss can be arbitrarily small as time grows.
\begin{lemma}
\label{lemma: small_train_loss_c2}
Under Condition \ref{condition: main condition}, there exists a time $t \le O(\frac{1}{M \epsilon})$ that 
\begin{align*}
    L_D(\Wb^{(t)}, \vb^{(t)}) \le \epsilon.
\end{align*}
Here $M$ is the same as the definition in Lemma \ref{lemma: output_lower_c2}.
\end{lemma}

\begin{proof}[Proof of Lemma \ref{lemma: small_train_loss_c2}]
    Combining the results of Lemma \ref{lemma: output_leading_term_c2} and Lemma \ref{lemma: output_lower_c2} we have
    \begin{align*}
        y_i f(\bW, \btheta; \bx_i) &\ge -0.25 + \sum_{r=1}^m v_{y_k, r, 2}^{(t)} \sigma(\la \wb_{y_k, r}^{(t)}, \bxi_k \ra)\\
        &\ge -0.25 + \log(M t).
    \end{align*}
    Therefore, we have
    \begin{align*}
        L_D(\Wb^{(t)}, \vb^{(t)}) &\le \log(1 + e^{0.25}/(M t)) \le \frac{e^{0.25}}{Mt}.
    \end{align*}
    The last inequality is from $\log(1 + x) \le x$ for $x \ge 0$. So when $t \ge \Omega(\frac{1}{M \epsilon})$, we have $L_D(\Wb^{(t)}, \vb^{(t)}) \le \epsilon$.
\end{proof}

\subsection{Test error analysis}
 Now we estimate the test error and derive the last two conclusions in Theorem \ref{thm: double_phase}. Assume a new data $(\bx, y), \bx = (\bx^{(1)}, \bx^{(2)})^{\top}$ where $\bx^{(1)} = y \cdot \bmu$. This is equivalent to estimating $\mathbb{P}(y \cdot f(\bW, \bV; \bx) > 0)$. The output of our trained model is
    \begin{align*}
        y \cdot f(\bW^{(t)}, \bV^{(t)}; \bx) = & \sum_{r=1}^{m} [v_{y,r,1}^{(t)} \sigma(\la \wb_{y,r}^{(t)}, y\bmu \ra) + v_{y,r,2}^{(t)}\sigma(\la \wb_{y,r}^{(t)}, \bxi \ra)]\\
        &- \sum_{r=1}^{m} [v_{-y,r,1}^{(t)} \sigma(\la \wb_{-y,r}^{(t)}, y\bmu \ra) + v_{-y,r,2}^{(t)}\sigma(\la \wb_{-y,r}^{(t)}, \bxi \ra)].
    \end{align*}

From previous analysis, when $t = \Omega(\frac{\sqrt{n}}{\eta \sigma_p \sqrt{d}})$, the output of signal part is lower bounded,
\begin{align}
\label{eq: signal_output_lower_c2}
    \sum_{r=1}^m v_{j,r,1}^{(t)} \la\wb_{j,r}^{(t)},y_k \bmu\ra \ge \tilde \Theta\Bigg(\frac{\sqrt{n}m \|\bmu\|_2^2 v_0^2}{\sigma_p \sqrt{d}}\Bigg).
\end{align}
    Similar to previous case, $\la \wb_{-y, r}^{(t)}, y\bmu \ra$ will be non-increasing and its order is neglectable compared with $\la \wb_{y,r}^{(t)}, y\bmu \ra$ which is non-decreasing.
    
So we have
\begin{align}
\label{eq: output_lower_c2}
    y \cdot f(\bW^{(t)}, \bV^{(t)}; \bx) &\ge \sum_{r=1}^m v_{j,r,1}^{(t)} \la\wb_{j,r}^{(t)},y_k \bmu\ra - \sum_{r=1}^m v_{-y,r,2}^{(t)} \sigma(\la \wb_{-y,r}^{(t)}, \bxi \ra).
\end{align}

Then we denote $g(\bxi) = \sum_{r=1}^{m} v_{-y,r,2}^{(t)} \sigma(\la \wb_{-y,r}^{(t)}, \bxi \ra)$. We can follow the proof of previous case and compute the expectation and Lipschitz norm of $g(\bxi)$:
\begin{align*}
    &\mathbb{E}(\sum_{r=1}^m v_{y,r,2}^{(t)} \sigma(\la \wb_{y,r}^{(t)}, \bxi \ra))
    = \sum_{r=1}^{m} v_{-y,r,2}^{(t)} \frac{\|\wb_{-y,r}^{(t)}\|_2 \sigma_p}{\sqrt{2\pi}}\\
    &\|g\|_{Lip} \le \sum_{r=1}^m v_{-y,r,2}^{(t)} \|\wb^{(t)}_{-y,r}\|_2.
\end{align*}

So we could upper bound $\|\wb_{j,r}^{(t)}\|$ as
    \begin{align*}
        \|\wb_{j,r}^{(t)}\| &\le  \|\wb_{j,r}^{(0)}\| + |\gamma_{j,r}^{(t)}\cdot \|\bmu\|_2^{-2} \cdot \bmu| + \bigg|\sum_{i=1}^n \rho_{j,r,i}^{(t)} \cdot \|\bxi_i\|_2^{-2} \cdot \bxi_i \bigg|\\ 
        &\le \|\wb_{j,r}^{(0)}\| + |\gamma_{j,r}^{(t)}\cdot \|\bmu\|_2^{-2} \cdot \bmu| + \frac{1}{\sqrt{n} \sigma_p \sqrt{d}} \cdot \bigg|\sum_{k=1}^n \rho_{j,r,k}^{(t)}\bigg|\\        
        &\le \frac{3}{2} \sigma_0^2 d + \tilde \Theta \bigg(\frac{n^{\frac{1}{4}}  v_0 \|\bmu\|_2}{m^{\frac{1}{2}} \sigma_p^{\frac{1}{2}} d^{\frac{1}{4}}}\bigg) + \tilde \Theta \bigg(\sqrt{\frac{\sqrt{n}}{m \sigma_p \sqrt{d}}} \bigg)\\
        &= \tilde \Theta \bigg(\sqrt{\frac{\sqrt{n}}{m \sigma_p \sqrt{d}}} \bigg),
    \end{align*}
    where the first inequality is from triangle inequality and the next two inequalities are from our estimate of parameters $\gamma$ and $\rho$ in Proposition \ref{prop: stage2_analysis_c2}.

    To estimate the final expectation of the noise output we also need to calculate $v_{-y, r, 2}^{(t)}$. As the two layers are balanced in this case, we have
    \begin{align*}
        v_{j,r,2}^{(t)} = \tilde \Theta\bigg(\sqrt{\frac{\sqrt{n}}{m\sigma_p \sqrt{d}}}\bigg).
    \end{align*}
So we have
    \begin{align}
    \label{eq: expect_upper_c2}
        \mathbb{E}(\sum_{r=1}^m v_{y,r,2}^{(t)} \sigma(\la \wb_{y,r}^{(t)}, \bxi \ra)) &= \tilde \Theta\bigg(\sqrt{\frac{n}{d}}\bigg).
    \end{align}

Now with results above, we can obtain
    \begin{align*}
        \mathbb{P}_{(\bx, y) \sim \textit{D}}& (y \cdot f(\bW^{(t)}, \bV^{(t)}; \bx) < 0) \\
        \le& \mathbb{P}_{(\bx, y) \sim \textit{D}}\bigg(\sum_{r=1}^m  v_{-y,r,2}^{(t)}\sigma(\la \wb_{-y,r}^{(t)}, \bxi \ra) > \sum_{r=1}^{m} v_{y,r,1}^{(t)} \sigma(\la \wb_{y,r}^{(t)}, y\bmu \ra)\bigg)\\
        =& \mathbb{P}_{(\bx, y) \sim \textit{D}}\bigg(g(\bxi) - \mathbb{E}[g(\bxi)] > \sum_{r=1}^{m} v_{y,r,1}^{(t)} \sigma(\la \wb_{y,r}^{(t)}, y\bmu \ra)\\
        & - \sum_{r=1}^{m} v_{-y,r,2}^{(t)} \frac{\|\wb_{-y,r}^{(t)}\|_2 \sigma_p}{\sqrt{2\pi}}\bigg)\\
        \le& \exp{\Bigg[-\frac{c (\sum_{r=1}^{m} v_{y,r,1}^{(t)} \sigma(\la \wb_{y,r}^{(t)}, y\bmu \ra - \sum_{r=1}^{m} v_{-y,r,2}^{(t)} \frac{\|\wb_{-y,r}^{(t)}\|_2 \sigma_p}{\sqrt{2\pi}})^2}{\sigma_p^2 (c_1 \sum_{r=1}^m v_{-y,r,2}^{(t)} \|\wb^{(t)}_{-y,r}\|_2)^2}\bigg]}\\
        =& \exp{\Bigg[-c\bigg(\frac{\sum_{r=1}^{m} v_{y,r,1}^{(t)} \sigma(\la \wb_{y,r}^{(t)}, y\bmu \ra)}{c_1 \sigma_p \sum_{r=1}^m v_{-y,r,2}^{(t)} \|\wb^{(t)}_{-y,r}\|_2}-\frac{1}{\sqrt{2\pi}c_1}\bigg)^2\Bigg]}\\
        \le& \exp{(1/c_1^2)}\exp{\Bigg(-0.5c\bigg(\frac{\sum_{r=1}^{m} v_{y,r,1}^{(t)} \sigma(\la \wb_{y,r}^{(t)}, y\bmu \ra)}{c_1 \sigma_p \sum_{r=1}^m v_{-y,r,2}^{(t)} \|\wb^{(t)}_{-y,r}\|_2}\bigg)^2\Bigg)}\\
        \le&\exp{\bigg(\frac{1}{c_1^2} - \frac{\tilde \Theta\Bigg(\frac{\sqrt{n}m \|\bmu\|_2^2 v_0^2}{\sigma_p \sqrt{d}}\Bigg)^2}{\tilde \Theta(\sqrt{\frac{n}{d}})^2} \bigg)}\\
        =& \exp(- c_2 m^2 \|\bmu\|_2^4 v_0^4 \sigma_p^{-2}))\\
        =& o(1).
    \end{align*}
    Here $c_1, c_2 = \tilde \Theta(1)$. The first inequality is from (\ref{eq: output_lower_c2}); The second inequality is from (\ref{eq: test_error_p}); The third inequality is from the fact that $(s-t)^2 \ge s^2/2 - t^2$ for all $s, t \ge 0$; The last inequality is from (\ref{eq: signal_output_lower_c2}) and (\ref{eq: expect_upper_c2}). This completes the proof for the second conclusion in Theorem \ref{thm: double_phase}.

Then we compute the test error lower bound. To estimate this, we again introduce an important lemma in this case:
    
    \begin{lemma}
\label{lemma: large_noise_change_c1_s}
    When $t = T_1$, denote $g(\bxi) = \sum_{j,r}jv_{j,r,2}^{(t)} \sigma(\la \wb_{j,r}^{(t)}, \bxi \ra )$. There exists a fixed vector $\btheta$ with $\|\btheta\|_2 \le 0.04 \sigma_p$ such that,
    \begin{align*}
        \sum_{j'\in \{\pm 1\}} [g(j'\bxi + \btheta) - g(j'\bxi)] \ge 4 c_3 \max \limits_{j'\in \{\pm 1\}} \Big\{\sum_r v_{j,r,1}^{(t)} \sigma(\la \wb_{j,r}^{(t)}, j\bmu \ra)\Big\},
    \end{align*}
    for all $\bxi \in \mathbb{R}^d$.
\end{lemma}

\begin{proof}[Proof of Lemma \ref{lemma: large_noise_change_c1_s}]
    Without loss of generality, let $\max \limits_{j'\in \{\pm 1\}} \{\sum_r v_{j,r,1}^{(t)} \sigma(\la \wb_{j,r}^{(t)}, j\bmu \ra)\} = \sum_r v_{1,r,1}^{(t)} \sigma(\la \wb_{1,r}^{(t)}, \bmu \ra)$.  We construct $\btheta$ as
    \begin{align*}
        \btheta = \lambda \cdot \sum_i \mathbb{I}(y_i = 1) \bxi_i,
    \end{align*}
    where $\lambda = \frac{c_4 m \|\bmu\|_2^2 v_0^2}{\sqrt{n} \sigma_p \sqrt{d}}$ and $c_4$ is a sufficiently large constant. Then we have
    \begin{align*}
        &\sigma(\la \wb_{1,r}^{(t)}, \bxi+\btheta \ra) - \sigma(\la \wb_{1,r}^{(t)}, \bxi \ra) + \sigma(\la \wb_{1,r}^{(t)}, -\bxi+\btheta \ra) - \sigma(\la \wb_{1,r}^{(t)}, -\bxi \ra)\\
        &\ge \sigma'(\la \wb_{1,r}^{(t)}, \bxi \ra) \la \wb_{1,r}^{(t)}, \btheta \ra + \sigma'(\la \wb_{1,r}^{(t)}, -\bxi \ra) \la \wb_{1,r}^{(t)}, \btheta \ra\\
        &= \la \wb_{1,r}^{(t)}, \btheta \ra\\
        &\ge \lambda \cdot \sum_{y_i=1} \la \wb_{1,r}^{(t)}, \bxi_i \ra
    \end{align*}
    The first inequity is because ReLU is a Liptchitz.
    Sum up all neurons, we have
    \begin{align*}
        &g(\bxi+\btheta) - g(\bxi) + g(-\bxi + \btheta) - g(-\bxi)\\
        &\ge \lambda \bigg[\sum_{r=1}^m v_{j,r,2}^{(t)} \sum_{y_i=1} \sigma(\la \wb_{j,r,2}^{(t)}, \bxi_i \ra )\bigg]\\
        &\ge \lambda \sum_{r=1}^m \frac{\sum_{y_i=1} v_{j,r,2}^{(t)} \sigma(\la \wb_{j,r,2}^{(t)}, \bxi_i \ra)}{v_{j,r,1}^{(t)} \sigma(\la \wb_{j,r}^{(t)}, \bmu \ra)} \cdot (v_{j,r,1}^{(t)} \sigma(\la \wb_{j,r}^{(t)}, \bmu \ra))\\
        &\ge \lambda \cdot \sum_{r=1}^m \tilde \Theta(\frac{n/m}{\sqrt{n}m \|\bmu\|_2^2 v_0^2 \sigma_p^{-1} d^{-\frac{1}{2}}}) \cdot (v_{j,r,1}^{(t)} \sigma(\la \wb_{j,r}^{(t)}, \bmu \ra))\\
         &\ge \lambda \cdot \tilde \Theta\Bigg(\frac{\sqrt{n} \sigma_p \sqrt{d}}{m \|\bmu\|_2^2 v_0^2} \Bigg) \cdot \sum_{r=1}^m v_{j,r,1}^{(t)} \sigma(\la \wb_{j,r}^{(t)}, \bmu \ra)\\
        &\ge 4 c_3 \cdot \sum_{r=1}^m v_{j,r,1}^{(t)} \sigma(\la \wb_{j,r}^{(t)}, \bmu \ra).
    \end{align*}
    The third inequity is from Proposition \ref{prop: stage2_analysis_c1} which estimates the outputs of signal and noise, and the last inequity is due to our definition of $\lambda$ and from Condition \ref{condition: main condition} that $\sigma_p \sqrt{d} \ge O(n/m)$. 
    
    Finally, the norm of $\btheta$ is small that
    \begin{align*}
        \|\btheta\|_2 = \bigg\|\lambda \cdot \sum_{y_i=1}\bxi_i \bigg\|_2 = \tilde \Theta(m \|\bmu\|_2^2 v_0^2) \le 0.04 \sigma_p,
    \end{align*}
    where the last inequity is by condition $m \|\bmu\|_2^2 v_0^2 = O(\sigma_p)$.
\end{proof}

With this Lemma, we can follow the same process as in case 1 to prove that 
\begin{align*}
         \mathbb{P}_{(\bx, y) \sim \textit{D}}(&y \cdot f(\bW^{(t)}, \bV^{(t)}; \bx) < 0) \ge 0.1.
     \end{align*}
And this completes the proof of Theorem \ref{thm: double_phase}.
\end{document}